\documentclass[11pt]{article}
\usepackage[margin=1.25in]{geometry}

\usepackage[margin=1.25in]{geometry}
\RequirePackage[OT1]{fontenc}
\RequirePackage{amsthm,amsmath,amsmath,amsfonts,amssymb}
\RequirePackage{natbib}
\RequirePackage[colorlinks,citecolor=blue,urlcolor=blue]{hyperref}

\RequirePackage{tablefootnote}

\RequirePackage{graphicx}

\allowdisplaybreaks[4]

\usepackage{times}

\usepackage{algorithm2e}
\SetKwRepeat{Do}{do}{while}%
\SetKwInput{KwInput}{Input}
\SetKwInput{KwOutput}{Output}
\SetKwInput{KwFunction}{Function}

\usepackage{titling}

\usepackage{authblk}
\usepackage{multibib}
\newcites{supp}{References}

\usepackage{graphicx}

\usepackage{enumitem}
\usepackage{subfigure}

\newtheorem{definition}{Definition}[section]

\newtheorem{theorem}{Theorem}[section]
\newtheorem{lemma}{Lemma}[section]
\newtheorem{corollary}{Corollary}[section]

\newtheorem{proposition}{Proposition}[section]

\theoremstyle{remark}
\newtheorem{remark}{Remark}[section]

\newcommand{\mat}{}
\newcommand{\vct}{}

\newcommand{\ud}{\mathrm d}

\newcommand{\kl}{\mathrm{KL}}

\newcommand{\argmin}{\mathrm{argmin}}

\newcommand{\vol}{\mathrm{vol}}

\newcommand{\var}{\mathrm{Var}}

\begin{document}

\title{Non-stationary Stochastic Optimization under $L_{p,q}$-Variation Measures}
\author[1]{Xi Chen}
\author[2]{Yining Wang}
\author[2,3]{Yu-Xiang Wang}
\affil[1]{Stern School of Business, New York University}
\affil[2]{Machine Learning Department, Carnegie Mellon University}
\affil[3]{Amazon, Inc.}

\maketitle

\begin{abstract}
We consider a non-stationary sequential stochastic optimization problem, in which the underlying cost functions change over time under a variation budget constraint. { We propose an $L_{p,q}$-variation functional to quantify the change,
 which yields less variation for dynamic function sequences whose changes are constrained to short time periods or small subsets of input domain.}
%which captures local spatial and temporal variations of the sequence of functions.
 Under the $L_{p,q}$-variation constraint, we derive both upper and matching lower regret bounds for smooth and strongly convex function sequences, which generalize previous results in \cite{besbes2015nonstationary}. {Furthermore, we provide an upper bound for general convex function sequences with noisy gradient feedback, which matches the optimal rate as $p \rightarrow \infty$.}
Our results reveal some surprising phenomena under this general variation functional, such as the curse of dimensionality of the function domain. The key technical novelties in our analysis include affinity lemmas that characterize the distance of the minimizers of two convex functions with bounded $L_p$ difference, and a cubic spline based construction that attains matching lower bounds.

\end{abstract}

\section{Introduction}

%\xnote{Comments on the revision:
%\begin{enumerate}
%\item Highlight the generalization to convex case in abstract, introduction, and conclusion.
%\item In the introduction, we always talk about he strongly convex. Now we need to also discuss the weakly convex case.
%\item We need to trim the paper to a technical note. We need to re-organize the proof (only leave the upper bound part in Section 4 (with the proof of affinity lemma)
%\end{enumerate}
%}

Non-stationary stochastic optimization studies the problem of optimizing a non-stationary sequence of convex functions on the fly, with either noisy gradient or function value feedback. %\xnote{consistent on function value feedback vs function feedback.}
This problem has important applications in operations research and machine learning, such as dynamic pricing, online recommendation services, and simulation optimization \citep{Gur:14:phdthesis,denBoer:15,denBoer:15:survey,Keskin:17}.
For example, in the case of dynamic pricing, an analyst is given the task of pricing a specific item over a long period of time, with feedback in the form of sales volumes in each time period.
As the demand changes constantly over time, the problem can be naturally formulated as non-stationary sequential stochastic optimization, where the analyst adjusts his/her pricing over time based on noisy temporal feedback data.

Formally, consider a sequence of $T$ convex functions $f_1,\cdots,f_T:\mathcal X\to\mathbb R$ over $T$ epochs,
where $\mathcal X\subseteq\mathbb R^d$ is a  convex, compact domain in the $d$-dimensional Euclidean space $\mathbb R^d$.
At each epoch $t\in\{1,\cdots, T\}$,
a policy $\pi$ selects an action $x_t\in\mathcal X$, based on \emph{stochastic or noisy feedback} (defined in Sec.~\ref{sec:formulation}) of previous epochs $1,\cdots,t-1$, and suffers loss $f_t(x_t)$.
The objective is to compete with the \emph{dynamic} optimal sequence of actions in hindsight; that is, to minimize regret
$$
\mathbb E^\pi\left[\sum_{t=1}^T{f_t(x_t)}\right] - \sum_{t=1}^T{\inf_{x\in\mathcal X}f_t(x)}.
$$
To ensure existence of policy with sub-linear regret {(i.e., the non-trivial regret of $o(T)$)}, constraints are imposed upon function sequences $f_1,\cdots,f_T$ such that any pair of consecutive functions $f_t$ and $f_{t+1}$ are sufficiently close, and therefore feedback through previous epochs are informative for later ones. { These constraints usually carry strong practical implications. For example, in dynamic pricing problems, an action $x$ represents the price and $f_t(x)$ is the (negative) revenue function at time $t$ in terms of price. Since the demand functions cannot change too rapidly, it is natural to impose a constraint on adjacent pairs of revenue functions (see, e.g., \cite{Keskin:17}).
%In simulation optimization, a simulator (represented by $f_t$) usually changes
%The work by \cite{Gur:14:phdthesis} also discussed applications to online content recommendation, where $f_t$ could represent the click  through rate of a recommended article.
}

The question of optimizing regret for non-stationary convex functions with stochastic feedback has received much attention in recent years.
One particular interesting instance of non-stationary stochastic convex optimization was considered in \cite{besbes2015nonstationary},
where sub-linear regret policies were derived when the average $L_{\infty}$ difference $\frac{1}{T}\sum_{t=1}^{T-1}{\|f_{t+1}-f_t\|_{\infty}}$ is assumed to go to zero as $T\to\infty$.
Optimal upper and lower regret bounds were derived for both noisy gradient and noisy function value feedback settings.

In this work, we generalize the results of \cite{besbes2015nonstationary} so that local spatial and temporal changes of functions are taken into consideration. %\xnote{We could move the examples from Section 2.1  and add better technical motivations here}
For any measurable function $f:\mathcal X\to\mathbb R$, define
\begin{equation}
\|f\|_p := \left\{\begin{array}{ll}
\left(\frac{1}{\vol(\mathcal X)}\int_{\mathcal X}{|f(x)|^p\ud x}\right)^{1/p}& 1\leq p<\infty;\\
\sup_{x\in\mathcal X}|f(x)|&p=\infty.\end{array}\right.
\label{eq:fp}
\end{equation}
Here,  $\vol(\mathcal X)=\int_{\mathcal X}{1\ud x}$ is the Lebesgue measure of the domain $\mathcal X$ and is finite because of the compactness of $\mathcal X$.
%Note that $\|\cdot\|_p$ satisfies the triangle inequality: $\|f+g\|_p\leq \|f\|_p+\|g\|_p$ for all $f,g:\mathcal X\to\mathbb R$ that have finite $\|f\|_p$ and $\|g\|_p$,
%thanks to the Minkowski's inequality.
We shall refer to $\|f\|_p$ as the $L_p$-norm
of $f$ in the rest of this paper.
(Conventionally in functional analysis the $L_p$ norm of a function is defined as the \emph{unnormalized} integration $\left(\int_{\mathcal X}|f(x)|^p\ud x\right)^{1/p}$.)
Nevertheless, we adopt the volume normalized definition  for the convenience of presentation.
{ It is worth noting  that this normalization will not affect our results.
In particular, because $\mathcal X\subseteq\mathbb R^d$ is a compact domain and $\vol(X)$ is a constant,
the regrets using the two definitions of function $L_p$ norm only differ by a multiplicative constant.}
Moreover, the Minkowski's inequality $\|f+g\|_p\leq\|f\|_p+\|g\|_p$, as well as other basic properties of $L_p$ norm, remains valid.
Also, for a sequence of convex functions $f_1,\cdots,f_T:\mathcal X\to\mathbb R$, define the \emph{$L_{p,q}$-variation functional} of $\vct f=(f_1,\cdots,f_T)$ as
\begin{equation}
\var_{p,q}(\vct f) := \left\{\begin{array}{ll}\left(\frac{1}{T}\sum_{t=1}^{T-1}{\|f_{t+1}-f_t\|_p^q}\right)^{1/q} &1\leq p\leq \infty, 1\leq q<\infty;\\
 \sup_{1\leq t\leq T-1}\|f_{t+1}-f_t\|_p& 1\leq p\leq\infty, q=\infty.\end{array}\right.
\label{eq:varpq}
\end{equation}
Note that in both Eqs.~(\ref{eq:fp}) and (\ref{eq:varpq}) we restrain ourselves to convex norms $p\geq 1$ and $q\geq 1$.
We can then define function classes
\begin{equation}\label{eq:function-class}
\mathcal F_{p,q}(V_T) := \left\{\vct f: \var_{p,q}(\vct f)\leq V_T\right\},
\end{equation}
{which serves as the budget constraint for a function sequence $\vct f$. The definition of $\mathcal F_{p,q}$ is more general than $\mathcal F_{\infty,1}$ introduced in \cite{besbes2015nonstationary}  since it better reflects the spatial and temporal locality of $\vct f$ in the subscripts $p$ and $q$.}

{

\subsection{A motivating example of dynamic pricing}\label{sec:motivation}

{ To motivate the $L_{p,q}$-variation constraint, we use dynamic pricing as a motivating example and illustrate the advantages of the $L_{p,q}$-variation measure for loss functions with ``local'' spatial or temporal changes.
%We give a dynamic pricing example to motivate our considerations of $L_{p,q}$ norms defined above, and the advantages the $L_{p,q}$ framework brings on objectives with ``local'' spatial or temporal changes.
We also provide guidelines on how $p,q$ values should be set qualitatively.}

We consider a stylized dynamic pricing problem of a single item  under changing revenue functions.
Let $\mathcal T=\{1,2,\cdots,T\}$ be a collection of $T$ time periods, at each of which the item receives a pricing $x_t$, $t\in\mathcal T$.
We normalize the prices so that their range is the unit interval $\mathcal X=[0,1]$.
At time period $t\in\mathcal T$, an unknown function $f_t:\mathcal X\to\mathbb R$ characterizes the negative expected revenue $f_t(x_t)$ a retailer collects
by setting the price at $x_t\in\mathcal X$.
The revenue function $f_t$ is assumed to be non-stationary over the time periods $t\in\mathcal{T}$.
The objective of the retailer is to design a pricing policy $\{x_t\}_{t=1}^T$ such that the aggregated (negative) expected revenue $\sum_{t=1}^Tf_t(x_t)$
is minimized.

\subsubsection{Spatial (pricing) locality of revenue changes}\label{subsec:example-lp}

We first fix $q=1$ in the $L_{p,q}$ variation framework and show how different values of $p$ reflect degrees of spatial (pricing) locality of the revenue functions $f_t$.
Suppose for all $t\in\{1,2,\cdots,T-1\}$, there exists a short interval $I_t\subseteq\mathcal X$ with its length $|I_t|\leq w$ such that $|f_t(x)-f_{t+1}(x)|\leq \delta$ for all $x\in I_t$,
and $f_t(x)=f_{t+1}(x)$ for all $x\in\mathcal X\backslash I_t$.
Intuitively, the assumption implies that the changes of the revenue functions $f_t, f_{t+1}$ between consecutive time periods $t,t+1\in\mathcal T$
are ``spatially local'', and the revenues are different only at prices in a small range $I_t$.
{This is a reasonable assumption in practice since the revenue $f_t(x_t)$ will not be sensitive to all possible prices in $\mathcal X$ (e.g., a pair of adjacent revenue function values remain the same when  price is very high or very low).}

Under the existing $L_{\infty,1}$ variation measure ($p=\infty$), simple calculation shows that $\var_{\infty,1}(\vct f) \leq \delta$.
On the other hand, for $p\in[1,\infty)$, the $L_{p,1}$ variation measure satisfies $\var_{p,1}(\vct f) \leq w^{1/p}\delta$.
When the ``locality'' level $w=|I_t|$ is much smaller than 1, $\var_{p,1}(\vct f)\ll \var_{\infty,1}(\vct f)$.
Furthermore, in cases where $\delta=1$ and $w=o(1)$, we have $\var_{\infty,1}(\vct f)=\Theta(1)$ and therefore the existing algorithm/analysis in \cite{besbes2015nonstationary}
cannot achieve sub-linear regret on $\vct f$;
on the other hand, by considering the $L_{p,1}$ measure,
one has $\var_{p,1}(\vct f)=o(1)$ for all $p\in[1,\infty)$, and therefore by applying algorithm/analysis in this paper we can achieve sub-linear regret on $\vct f$.

\subsubsection{Temporal locality of revenue changes}\label{subsec:example-lq}
We next consider $p=\infty$ in the $L_{p,q}$ variation framework and show how different values of $q$ reflect degrees of temporal locality of the revenue function $f_t$.
Suppose there exists a subset if time periods $\mathcal S\subseteq\mathcal T$, $|\mathcal S|= s\ll T=|\mathcal T|$ such that $\|f_{t+1}-f_t\|_\infty = \delta$ for all $t\in\mathcal S$,
and $f_{t+1}\equiv f_t$ for all $t\in\mathcal T\backslash\mathcal S$.
Intuitively, this assumption implies that the revenue function $f_t$ has \emph{local temporal changes},
meaning that the $f_t$ changes only in short time intervals $\mathcal S$  and remains the same for most of the other times.
This is a relevant assumption when demands of the item have clear temporal correlations, such as seasonal food and clothes.

Simple calculations show that, for $p=\infty$ and $q\in[1,\infty]$, the $L_{\infty,q}$ variation measure of the above described function sequence
is $(s/T)^{1/q}\delta$.
This demonstrates the effect of the parameter $q$ in $L_{\infty, q}$-variation for $\vct f$ with local temporal changes, i.e.,
a smaller $q$ leads to a smaller variation measure $L_{\infty,q}$ of $\vct f$ when $s\ll T$.

\subsubsection{Guidelines on the selection of $p,q$ values}
Though the underlying sequence of expected revenue functions $\vct f=(f_1,\cdots,f_T)$ is assumed to be unknown, in practice it is common that certain background knowledge or prior information
is available regarding $\vct f$.
In this section we discuss how such prior information, especially regarding the magnitude changes of $f_{t+1}$ and $f_t$ in $\vct f$, can qualitatively help us select the parameters $p,q$ in
the variation measure.

We first discuss the selection of $p$ and fix the choice $q=1$ for the moment.
Suppose we have the prior knowledge that each pairs of $f_{t+1}$ and $f_t$ differ significantly on $w\ll 1$ portion of the domain $\mathcal X$ by a difference of $\delta\leq 1$,
as exemplified in Sec.~\ref{subsec:example-lp}.
{ Then the $L_{p,1}$ variation of such function sequence is approximately $V_T=\var_{p,q}(\vct f) \approx w^{1/p}\delta$.
 According to our results in Theorems \ref{thm:main-upper}--\ref{thm:general-convex}, the worst-case regret is $T\cdot V_T^{\varrho(p,d)}$
where $\varrho(p,d)\in\{2p/(4p+d), 2p/(6p+d), p/(3p+d)\}$ depending on feedback types (e.g., noisy gradient or function value feedback) and (strong) convexity of $\vct f$.}
The regret can be further re-parameterized as $T\cdot w^{\varphi(p,d)}\delta^{\varrho(p,d)}$ where $\varphi(p,d)\in\{2/(4p+d), 2/(6p+d), 1/(3p+d)\}$.

%Because $\varphi(p,d)$ and $\varrho(p,d)$ are strictly decreasing/increasing functions of $p$, and $w,\delta\leq 1$,
{ The above analysis leads to the following insights providing qualitative suggestions of $p$ choices:}
\begin{enumerate}
\item The $w^{\varphi(p,d)}$ term is smaller for smaller $p$ values, because $w\ll 1$ and $\varphi(p,d)$ is a strictly decreasing function in $p$.
{ This suggests that for function sequences with stronger spatial locality (e.g.,  revenue functions that only change on a small range of prices), one should use a smaller $p$ value in $L_{p,1}$-variation measure;}
\item The $\delta^{\varrho(p,d)}$ term is smaller for larger $p$ values, because $\delta\leq 1$ and $\varrho(p,d)$ is a strictly increasing function in $p$.
This suggests that for function sequences with smaller absolute amount of perturbation, one should use a larger $p$ in $L_{p,1}$-variation measure.
\end{enumerate}

{
We next discuss the selection of $q$ and fix the choice of $p\in[1,\infty]$.
Unlike the spatial locality parameter $p$, our Theorems \ref{thm:main-upper} and \ref{thm:main-lower} suggest that
the optimal worst-case regret is insensitive to the choice of $q\in[1,\infty]$.
This might sound surprising, but is the characteristic of the adopted \emph{worst-case} analytical framework. To see this, we note that the worst-case function sequence
is the one that evenly distributes the function changes $\|f_{t+1}-f_t\|_p$ across all $t\in\mathcal T$ (see also the detailed construction in the online supplement),
{ in which case the $L_{p,q}$-variation measure is the same for all $q\in[1,\infty]$.
It should also be noted that the choice of $q$ does not affect our optimization algorithm or its re-starting procedure.
Therefore,  we simply recommend the selection of $q=1$ but we choose to include $q$ in our theorem statements for mathematical generality.
}
}
}

\subsection{Results and techniques}

The main result of this paper is to characterize the optimal regret over function classes $\mathcal F_{p,q}(V_T)$, which includes
explicit algorithms that are computationally efficient and attain the regret, and a lower bound argument based on Fano's inequality \citep{Ibragimov:Hasminskii:81book,yu1997assouad,Cover:06,tsybakov2009introduction} that shows the regret attained is
optimal and cannot be further improved.
Below is an informal statement of our main result (a formal description is given in Theorems \ref{thm:main-upper} and \ref{thm:main-lower}):

\paragraph{Main result (informal).}  \emph{For smooth and strongly convex function sequences under certain regularity conditions, the optimal regret over $\mathcal F_{p,q}(V_T)$ is $T\cdot V_T^{2p/(4p+d)}$ with noisy gradient feedback,
and $T\cdot V_T^{2p/(6p+d)}$ with noisy function value feedback, {provided that $V_T$ is not too small}.
{ In addition, for general convex function sequences satisfying only Lipschitz continuity on function values, we obtain a regret upper bound of $T\cdot V_T^{p/(3p+d)}$
with noisy gradient feedback, provided that $V_T$ is not too small.}
Here $d$ is the dimension of the domain $\mathcal X$.}

{ We clarify that our results also cover the case of small $V_T$, i.e., $V_T$ converges to 0 as $T\to\infty$ at a very fast rate. However, the case of ``not too small $V_T$'' is of more interest. This is because  if $V_T$ is very small, meaning that the underlying function sequence $\{f_1,\cdots,f_T\}$ is close to a stationary one (i.e., $f_1=f_2=\ldots=f_T=f$),
then one could re-produce the standard $O(\sqrt{T})$ and/or $O(\log T)$ regrets ($O(\sqrt{T})$ for strongly convex and smooth functions with noisy function feedback,
$O(\log T)$ for strongly convex and smooth functions with noisy gradient feedback, and $O(\sqrt{T})$ for general convex functions with noisy gradient feedback;
% for stationary online optimization with general convex and strong convex functions
 see also, e.g., \cite{jamieson2012query,agarwal2010optimal,hazan2007logarithmic}.)
 These rates are also  known to be optimal \citep{jamieson2012query,hazan2014beyond}.
Technical details of this point are given in the statements of Theorems \ref{thm:main-upper}, \ref{thm:main-lower}, \ref{thm:general-convex}.
}

% { It is worthy noting that the case that $V_T$ is not too small is of more interest since when $V_T \rightarrow \infty$
More importantly, our result reveals several interesting facts about the regret over function sequences with local spatial and temporal changes.
Most surprisingly, the optimal regret suffers from \emph{curse of dimensionality}, as the regret depends \emph{exponentially}
on the domain dimension $d$.
Such phenomenon does not occur in previous works on stationary and non-stationary stochastic optimization problems.
For example,  for the case of $\vct f$ being strongly convex and smooth, as spatial locality in $\vct f$ becomes less significant (i.e., $p\to\infty$), the optimal regrets approach $T\cdot V_T^{1/2}$ (for noisy gradient feedback) and $T\cdot V_T^{1/3}$ (for noisy function value feedback),
which recovers the dimension-independent regret bounds in \cite{besbes2015nonstationary} derived for the special case of $p=\infty$ and $q=1$. { Similar phenomenon of curse of dimensionality also appears in the general convex case.
We also note that, when $V_T$ is not too small,  the obtained regret bound $T\cdot V_T^{p/(3p+d)}$  matches the {optimal} $O(T\cdot V_T^{1/3})$ rate for $p=\infty$ in \cite{besbes2015nonstationary} as $p\to\infty$.
}

To obtain results for general $L_{p,q}$-variation and the optimal regrets for strongly convex case, we make several important technical contributions in this paper, which are highlighted as follows.
\begin{enumerate}%[itemindent=0cm]
\item { For noisy function value feedback, instead of using the online gradient descent (OGD) from \cite{besbes2015nonstationary}, we adopt a regularized ellipsoidal (RE) algorithm from \cite{hazan2014bandit} and extend it from exact function value evaluation to the \emph{noisy} version.}
    Our analysis relaxes an important assumption in \cite{besbes2015nonstationary} that requires the optimal solution to lie far away from the boundary of $\mathcal X$. Our policy based on the RE algorithm allows the optimal solution to be closer to the boundary of $\mathcal X$ as $T$ increases.
\item On the upper bound side, we prove an interesting affinity result (Lemma \ref{lem:key2}) which shows that the optimal solutions $x_t^*,x_\tau^*$ of $f_t,f_\tau$
cannot be too far apart provided that both $f_t,f_\tau$ are smooth and strongly convex functions, and $\|f_t-f_\tau\|_p$ is upper bounded.
{ The affinity result is also generalizable to non-strongly convex functions $f_t,f_\tau$ (Lemma \ref{lem:key1}), by
directly integrating function differences in a close neighborhood of $x_t^*$ (or $x_\tau^*$) without resorting to $\|x_t^*-x_\tau^*\|$
(that could be unbounded without strong convexity).
}
Both affinity results are key in deriving upper bounds for our problem, and have not been discovered in previous literatures.
They might also be potentially useful for other non-stationary stochastic optimization problems (e.g., adaptivity to unknown parameters \citep{besbes2015nonstationary,karnin2016multi}).

%\xnote{We need to discuss the generalization to weakly convex case.}
\item On the lower bound side, we present a systematic framework to prove lower bounds by first reducing the non-stationary stochastic optimization problem to an estimation problem with active queries,
and then applying the Fano's inequality with a ``sup-argument'' similar in spirit to \cite{castro2008minimax} that handles the active querying component. { To adapt Fano's inequality, we also design a new construction of adversarial function sets, which is quite different from the one in \cite{besbes2015nonstationary}. More specifically, to prove that the regret exhibits ``curse of dimensionality'', one needs to construct functions $f_1,f_2$ that not only have different minima but also ``localized'' difference (meaning that $f_1(x)=f_2(x)$ for most $x\in\mathcal X$) such that $\|f_1-f_2\|_p$ is small.
  %one needs to construct functions $f_1,f_2$ with different minima but very ``localized'' differences (meaning that $f_1(x)=f_2(x)$ for most $x\in\mathcal X$)
  To construct such adversarial functions, we use the idea of ``smoothing splines'' from nonparametric statistics that connects two pieces of quadratic functions   using a \emph{cubic} function to ensure the smoothness and strong convexity of the constructed functions.
}
%We also construct a cubic spline example to facilitate the lower bound construction and satisfy all desired smoothness and convexity properties at the same time.
Our analytical framework and spline-based lower bound construction could inspire new lower bounds for other online and non-stationary optimization problems.
\end{enumerate}

%The primary focus of this paper is on smooth and strongly convex function sequences (see Sec.~\ref{sec:formulation} for detailed technical definitions). While the more general case of merely convex functions are certainly of interest, optimal regret bounds are very challenging to obtain under the general $L_{p,q}$-variation functional constraint. In particular, for noisy function value feedback the conventional gradient estimating procedures are sub-optimal, and considerably more complicated techniques are required to even solve
%the stationary benchmark version of bandit convex optimization \citep{bubeck2016kernel}. { In fact, even for the function class $\mathcal{F}_{\infty,1}$ as in \cite{besbes2015nonstationary}, the optimal regret bounds are still open for noisy function value feedback.}
%For noisy gradient feedback, \cite{besbes2015nonstationary} derived optimal regret bounds for the $p=\infty$ case.
%However, extension of their arguments to the $1\leq p<\infty$ case is highly non-trivial for general convex functions, as we remark in Sec.~\ref{sec:conclusion}.

\subsection{Related work}
\label{sec:related}

In addition to the literature discussed in the introduction, we briefly review a few additional recent works from machine learning and optimization communities.

\paragraph{Stationary stochastic optimization.}
The stationary stochastic optimization problem considers a stationary function sequence $f_1=f_2=\cdots=f_T=f$,
and aims at finding a near-optimal solution $x\in\mathcal X$ such that $f(x)$ is close to $f^*=\inf_{x\in\mathcal X}f(x)$.
When only noisy function evaluations are available at each epoch, the problem is also known as \emph{zeroth-order optimization}
and has received much attention in the optimization and machine learning  community.
Classical approaches include confidence-band methods \citep{agarwal2013stochastic} and pairwise comparison based methods \citep{jamieson2012query},
both of which achieve $\widetilde O(\sqrt{T})$ regret with polynomial dependency on domain dimension $d$.
{ Here in $\widetilde O(\cdot)$ notation we drop poly-logarithmic dependency on $T$.}
%\xnote{A bit strange of talking regret and convergence in the same sentence.}
The tight dependency on $d$, however, remains open.
In the more restrictive \emph{statistical optimization} setting $f(x)=\mathbb E_{z\sim P}[F(z,x)]$, optimal dependency on $d$ can be attained
by the so-called ``two-point query'' model \citep{shamir2015optimal}.

\paragraph{Online convex optimization.}
In online convex optimization, an \emph{arbitrary} convex function sequence $f_1,\cdots,f_T$ is allowed, and the regret of a policy $\pi$
is compared against the optimal \emph{stationary} benchmark $\inf_{x\in\mathcal X}\{\sum_{t=1}^T{f_t(x)}\}$ in hindsight.
Unlike the stochastic optimization setting, in online convex optimization the full information of $f_t$ is revealed to the optimizing algorithm after epoch $t$,
which allows for exact gradient methods.
It is known that for unconstrained online convex optimization, the simplest gradient descent method attains $O(\sqrt{T})$ regret for convex functions,
and $O(\log T)$ regret for strongly convex and smooth functions, both of which are optimal in the worst-case sense \citep{hazan2016introduction}.
For constrained optimization problems, projection-free methods exist following mirror descent or follow-the-regularized-leader (FTRL) methods \citep{hazan2014bandit}.
{
\cite{Zinkevich2003,Hall2015} considered the question of online convex optimization by competing against the optimal dynamic \emph{solution sequence} $x_1^*,\cdots,x_T^*$
subject to certain smoothness constraints like $\sum_t \|x_{t+1}^*-x_t^*\|\leq C$.
\cite{Jadbabaie2015,Mokhtari2016} further imposed the constraint on both solution sequences and function sequences in terms of $L_{\infty, 1}$-variation {and showed that adaptivity to the unknown smoothness parameter $V_T$ is possible with noiseless gradient and the information of $\|f_t-f_{t-1}\|_\infty$.}
\cite{Daniely2015,Zhang2017} also designed algorithms that \emph{adapt} to the unknown smoothness parameter,
under the model that the entire function $f_t$ is revealed after time  $t$.
However, the adaptation still remains an open problem  in the ``bandit'' feedback setting considered in our paper,
in which only noisy evaluations of $f_t(x_t)$ or $\nabla f_t(x_t)$ are revealed. {Under the bandit feedback setting, the function perturbations (e.g., $\|f_{t+1}-f_t\|_\infty$) cannot be easily estimated,
 making it unclear whether adaptation to $V_T$ is possible.
}
}

\paragraph{Bandit convex optimization.} Bandit convex optimization is a combination of stochastic optimization and online convex optimization,
where the stationary benchmark in hindsight of a sequence of arbitrary convex functions $\inf_{x\in\mathcal X}\{\sum_{t=1}^T{f_t(x)}\}$ is used
to evaluate regrets. %\xnote{Should be $t$ or $T$?}
At each time  $t$, only the function evaluation at the queried point $f_t(x_t)$ (or its noisy version) is revealed to the learning algorithm.
Despite its similarity to stochastic and/or online convex optimization, convex bandits are considerably harder due to its lack of first-order information and the arbitrary change of functions.
\cite{flaxman2005online} proposed a novel finite-difference gradient estimator,
which was adapted by \cite{hazan2014bandit} to an ellipsoidal gradient estimator
that achieves $\widetilde O(\sqrt{T})$ regret for constrained smooth and strongly convex bandits problems.
For the non-smooth and non-strongly convex bandits problem, the recent work of \cite{bubeck2016kernel} attains $\widetilde O(\sqrt{T})$ regret
with an explicit algorithm whose regret and running time both depend polynomially on dimension $d$.

\subsection{Notations and basic properties of $\var_{p,q}$}

For a $d$-dimensional vector we write $\|x\|_p=(\sum_{i=1}^d{|x_i|^p})^{1/p}$ to denote the $\ell_p$ norm of $x$, for $0< p<\infty$,
and $\|x\|_\infty = \max_{1\leq i\leq d}|x_i|$ to denote the $\ell_\infty$ norm of $x$.
Define $\mathbb B_d(r) := \{x\in\mathbb R^d: \|x\|_2\leq r\}$ and $\mathbb S_d(r):= \{x\in\mathbb R^d: \|x\|_2=r\}$
as the $d$-dimensional ball and sphere of radius $r$, respectively.
We also abbreviate $\mathbb B_d=\mathbb B_d(1)$ and $\mathbb S_d=\mathbb S_d(1)$.
For a $d$-dimensional subset $\mathcal X\subseteq\mathbb R^d$,
denote $\mathcal X^o=\{x\in\mathcal X: \exists r>0, \forall z\in\mathbb B_d(r), x+r\in\mathcal X\}$
as the \emph{interior} of $\mathcal X$, $\bar{\mathcal X} = \{\lim_{n\to\infty} x_n: \{x_n\}_{n=1}^{\infty}\subseteq\mathcal X\}$ as the \emph{closure} of $\mathcal X$,
and $\partial\mathcal X = \bar{\mathcal X}\backslash\mathcal X^o$ as the \emph{boundary} of $\mathcal X$.
{For any $r>0$, we also define $\mathcal X^o_r=\{x\in\mathcal X^o: \forall z\in\mathbb B_d(r), x+z\in\mathcal X\}$ as the ``strict interior'' of $\mathcal X$,
where every point in $\mathcal X^o_r$ is guaranteed to be at least $r$ away from the boundary of $\mathcal X$.}

We note that the $\var_{p,q}$ defined in \eqref{eq:varpq} is monotonic in $p$ and $q$, as shown below:
\begin{proposition}
For any $1\leq p\leq p'\leq \infty$ and $1\leq q\leq q'\leq\infty$ it holds that $\var_{p,q}(\vct f)\leq \var_{p',q'}(\vct f)$.
In addition, for any $1\leq p\leq\infty$ we have $\lim_{q\to\infty}\var_{p,q}(\vct f) = \var_{p,\infty}(\vct f)$,
and similarly for any $1\leq q\leq\infty$ we have $\lim_{p\to\infty}\var_{p,q}(\vct f)=\var_{\infty,q}(\vct f)$,
assuming all functions in $\vct f$ are continuous.
\label{prop:monotonicity}
\end{proposition}
The proof of Proposition \ref{prop:monotonicity} is deferred to Section \ref{sec:proof_mono} in the online supplement.

The rest of the paper is organized as follows. In Section \ref{sec:formulation}, we introduce the problem formulation. Section \ref{sec:main} contains the main results and describes the policies. Section \ref{sec:proof} presents the proof of our main positive result. The concluding remarks and future works are discussed in Section \ref{sec:conclusion}. Additional proofs can be found in the online supplement.

%meaning that for any $1\leq p\leq p'\leq\infty$ and $1\leq q\leq q'\leq\infty$,
%it holds that $\var_{p,q}(\vct f)\leq\var_{p',q',}(\vct f)$.
%In addition, $p=\infty$ and $q=\infty$ are the limits of $p,q\to\infty$: $\lim_{p\to\infty}\var_{p,q}(\vct f)=\var_{\infty,q}(\vct f)$ for all $q$ and
%$\lim_{q\to\infty}\var_{p,q}(\vct f)=\var_{p,\infty}(\vct f)$ for all $p$.  Due to space constraints we omit the verification of the above-mentioned properties of $\var_{p,q}$, which is routine in functional analysis.

\section{Problem formulation}\label{sec:formulation}

Suppose $f_1,\cdots,f_T$ are a sequence of unknown convex differentiable functions supported on a bounded convex set $\mathcal X\subseteq\mathbb R^d$.
At epoch $t\in\{1,\cdots,T\}$, a policy selects a point $x_t\in\mathcal X$ (i.e., makes an action) and suffers loss $f_t(x_t)$.
Certain feedback $\phi_t(x_t,f_t)$ is then observed which can guide the decision of actions in future epochs.
Two types of feedback structures are considered in this work:
\begin{itemize}
\item[-] \emph{Noisy gradient feedback}: $\phi_{t}^{\mathcal G}(x_t,f_t) = \nabla f_t(x_t) + \varepsilon_t$,
where $\nabla f_t(x_t)$ is the gradient of $f_t$ evaluated at $x_t$, and $\varepsilon_t$ are independent $d$-dimensional random vectors such that each component $\varepsilon_{ti}$ is a random variable with $\mathbb E[\varepsilon_{ti}|x_t]=0$; %and $\mathbb E[\varepsilon_{ti}^2|x_t]\leq\sigma^2<\infty$;
furthermore, $\varepsilon_{ti}$ conditioned on $x_t$ is a sub-Gaussian random variable with parameter $\sigma^2$, meaning that
$\mathbb E[\exp(a\varepsilon_{ti})|x_t] \leq \exp(a^2\sigma^2/2)$ for all $a\in\mathbb R$;
%\yxwcomment{[Question] Do we need the noise $\varepsilon_t$ to be independent (conditioning on $x_t$) in every dimension?}

\item[-] \emph{Noisy function value feedback}: $\phi_t^{\mathcal F}(x_t,f_t) = f_t(x_t) + \varepsilon_t$,
where $\varepsilon_t$ are independent univariate random variables that satisfy $\mathbb E[\varepsilon_t|x_t]=0$; % and $\mathbb E[\varepsilon_t^2|x_t]\leq\sigma^2<\infty$.
furthermore, $\varepsilon_t$ conditioned on $x_t$ is a sub-Gaussian random variable with parameter $\sigma^2$, meaning that
$\mathbb E[\exp(a\varepsilon_t)|x_t]\leq \exp(a^2\sigma^2/2)$ for all $a\in\mathbb R$.
%\yxwcomment{[Question] I believe subgaussianity is used in the proof (Section EC 3.1). Don't we need to add that here?}

\end{itemize}
%\xnote{random vectors ro variables? Do we make independence assumption? }
Both feedback structures are popular in the optimization literature and were considered in previous work on online convex optimization and stochastic bandits (e.g., \citet{hazan2016introduction} and references therein).
For notational convenience, we shall use $\phi_t(x_t,f_t)$ or simply $\phi$ to refer to a general feedback structure without specifying its type, which can be either $\phi_t^{\mathcal G}(x_t,f_t)$
or $\phi_t^{\mathcal F}(x_t,f_t)$.

Apart from $\mathcal X$ being closed convex and $f_1,\cdots,f_T$ being convex and differentiable, we also make the following additional assumptions on the domain $\mathcal X$
and functions $f_1,\cdots,f_T$:
\begin{enumerate}
\item[(A1)] (\emph{Bounded domain}): there exists constant $D>0$ such that $\sup_{x,x'\in\mathcal X}\|x-x'\|_2 \leq D$;
\item[(A2)] (\emph{Bounded function and gradient}): there exists constant $H>0$ such that $\sup_{x\in\mathcal X}|f_t(x)|\leq H$ and $\sup_{x\in\mathcal X}\|\nabla f_t(x)\|_2\leq H$;
%\item[(A3)] {(\emph{Unique interior minimizer}): there exists unique $x_t^*\in\mathcal X$ such that $f_t(x_t^*)=\inf_{x\in\mathcal X}f_t(x)$;
%Furthermore, there exists $\nu>0$ such that $\mathcal X_{\nu/T}^o:=\{x\in\mathcal X^o: x+z\in\mathcal X,\forall z\in\mathbb R^d,\|z\|_2\leq\nu/T\}\neq\emptyset$;}
%furthermore, there exists $\nu>0$ such that $x_t^*+z\in\mathcal X$ for all $z\in\mathbb B_d(\nu/T)$; or equivalently, $x_t^*\in\mathcal X^o_{\nu/T}$;}

\item[(A3)] (\emph{Unique interior optimizer}): there exists unique $x_t^*\in\mathcal X$ such that $f_t(x_t^*)=\inf_{x\in\mathcal X}f_t(x)$. Furthermore,  the interior of $\mathcal X$ is a non-empty set (i.e., $\mathcal X^o\neq\emptyset$) and there exists $\nu>0$ such that $\{z\in\mathbb R^d: \|z-x_t^*\|_2\leq\nu/T\}\subseteq\mathcal X$.

%\yxwcomment{[Comment] As I explained in the email (and added in the proof of Lemma 1), I think we can drop the assumption of $x_t^*$ being in the $\nu/T$ interior of $\cX$.  Then the non-emptiness of the $\tilde{X}$ (the set obtained by removing a $\nu/T$ ``shell'' of $\cX$) becomes without loss of generality. Please check through this and the discussion of the assumptions in the subsequent paragraphs should be updated accordingly.}
\item[(A4)] (\emph{Smoothness}): there exists constant $L>0$ such that $f_t(x')\leq f_t(x)+\nabla f_t(x)^\top(x'-x) + \frac{L}{2}\|x'-x\|_2^2$ for all $x,x'\in\mathcal X$.
\item[(A5)] (\emph{Strong convexity}): there exists constant $M>0$ such that $f_t(x')\geq f_t(x)+\nabla f_t(x)^\top(x'-x) + \frac{M}{2}\|x'-x\|_2^2$ for all $x,x'\in\mathcal X$.
\end{enumerate}
The assumptions (A1), (A2) are standard assumptions that were imposed in previous works on both stationary and non-stationary stochastic optimization \citep{flaxman2005online, agarwal2013stochastic,shamir2015optimal,besbes2015nonstationary}.
%{\color{red}%The condition (A3) assumes that the optimal solution $x_t^*$ is not too close to the boundary of the domain $\mathcal X$.
%The condition (A3) assumed the existence of a ``strict'' interior subset $\mathcal X_{\nu/T}^o$ of the domain $\mathcal X$ such that any point in the strict interior $\mathcal X_{\nu/T}^o$
%is at least $\nu/T$ apart from the boundary of $\mathcal X$.
The condition (A3) assumes that the optimal solution $x_t^*$ is not too close to the boundary of the domain $\mathcal X$.
Compared to similar assumptions in existing work \citep{flaxman2005online,besbes2015nonstationary}, our assumption is considerably weaker
since $x_t^*$ can be within $\Omega(1/T)$ distance to the boundary;
while in \cite{flaxman2005online,besbes2015nonstationary}, $x_t^*$ must be $\Omega(1)$ distance away from the boundary (i.e., away from the boundary by at least a constant).
%\xnote{Could you add more on why this relaxation might be important and useful?}
Finally, the conditions (A4) and (A5) concern second-order properties of $f_t$ and enable smaller regret rates for gradient descent algorithms.
We note that the condition $M \mat I_d\preceq \nabla^2 f_t(x)\preceq L\mat I_d, \forall x\in\mathcal X$ in \cite{besbes2015nonstationary}  (see Eq.~(10) in \cite{besbes2015nonstationary})
is stronger and implies our (A4) and (A5) since we do not assume  that $f_t$ is twice differentiable.
We also consider parameters  $D,\nu,H,L,M$ in (A1)--(A5) and domain dimensionality $d$ as constants throughout the paper and omit their (polynomial) {multiplicative} dependency in regret bounds. { In Section \ref{subsec:general-convex}, we further relax the assumptions (A3)--(A5) and provide upper bound results for general convex function sequences.
}

%\xnote{Should we add this word multiplicative since $d$ appears in the exponent of $V_T$.}
%\yxwcomment{[Comment] Should we add that we do not have super-polynomial dependence in any of these constants?}

%We remark that \cite{besbes2015nonstationary} also considered the case of noisy gradient feedbacks when functions are only assumed to be convex and Lipschitz instead of being strongly convex.
%We, however, choose to focus solely on the smooth and strongly convex setting because deriving optimal regrets over convex functions that are merely Lipschitz is particularly challenging for our generalized $\mathcal F_{p,q}$ function class.
%We will provide an in-depth discussion on the convex setting in Sec.~\ref{sec:conclusion} and shed lights on the technical difficulty of deriving tight regret bounds.

%\xnote{We should be careful on this. \citep{besbes2015nonstationary} also considered the non-strongly convex case. We should highlight that why we only consider strongly-convex case and difficulty for convex case?}

Let $U$ be a random quantity defined over a probability space.
A policy $\pi$ that outputs a sequence of $x_1,\cdots,x_T$ is \emph{admissible} if it is a measurable function that can be written in the following form:
$$
x_t = \left\{\begin{array}{ll}
\pi_1(U),& t=1;\\
\pi_t(U, x_1, \phi_1(x_1,f_1), \cdots, x_{t-1}, \phi_{t-1}(x_{t-1}, f_{t-1})),& t=2,\cdots,T,\end{array}\right.
$$
%where $U_1,\cdots,U_T$ are randomized bits utilized by the policy and the range of $\pi_t$ is contained in $\mathcal X$. %\xnote{OR people do not use ``randomized bits''. Could you change to the filtration type argument as in  \citep{besbes2015nonstationary} and indicate $\pi_i$ is measurable function w.r.t. this filtration?}
Let $\mathcal P_T^\pi$ denote the class of all admissible policies for $T$ epochs. { A widely used metric for evaluating the performance of an admissible policy $\pi$ is the regret against dynamic oracle $\{x_t^*\}_{t=1}^T$:%which is defined as
\begin{equation}
\sup_{\vct f=(f_1,\cdots,f_T)\in\mathcal F_{p,q}(V_T)} R_{\phi}^\pi(\vct f) := \mathbb E^\pi\left[\sum_{t=1}^T{f_t(x_t)}\right] - \sum_{t=1}^T{f_t(x_t^*)}, \;\;\;\;\;
x_t^* = \arg\min_{x\in\mathcal X}f_t(x).
\label{eq:regret_0}
\end{equation}
Here $\phi$ is either the noisy gradient feedback $\phi_t^{\mathcal G}(x_t,f_t)$ or the noisy function feedback $\phi_t^{\mathcal F}(x_t, f_t)$.
Note that a unique minimizer $x_t^*\in\mathcal X$ exists due to the strong convexity of $f_t$ (condition A5).
The goal of this paper is to  characterize the optimal regret:
\begin{equation}
  \inf_{\pi\in\mathcal P_T^\pi} \sup_{\vct f=(f_1,\cdots,f_T)\in\mathcal F_{p,q}(V_T)} R_{\phi}^\pi(\vct f), \label{eq:regret}
\end{equation}
and find policies that achieve the rate-optimal regret, i.e., attain the optimal regret up to a polynomial of $\log(T)$ factor. The optimal regret in \eqref{eq:regret} is also known as the \emph{minimax regret} in the literature, because it minimizes over all admissible policies
and maximizes over all convex function sequences $\vct f\in\mathcal F_{p,q}(V_T)$.}

\section{Main results}
\label{sec:main}

We establish theorems giving both upper and lower bounds on worst-case regret for both noisy gradient feedback $\phi_t^{\mathcal G}(x_t,f_t)$
and noisy function feedback $\phi_t^{\mathcal F}(x_t,f_t)$ over $\mathcal F_{p,q}(V_T)$. The policies for achieving the following upper bound result will be introduced in the next section.

\begin{theorem}[Upper bound for strongly-convex function sequences]
Fix arbitrary $1\leq p<\infty$ and $1\leq q<\infty$.
Suppose (A1) through (A5) hold, and { $0\leq V_T\leq 1$}.
%$1/T\leq V_T\leq 1$.
Then there exists a computationally efficient policy $\pi$ and
{ $C_1=g_1(\log T,\log V_T,d,D,\nu,L,H,M)>0$
for some function $g_1$ that is a polynomial function in $\log T$ and $\log V_T$, such that}
%as a polynomial function of $\log T$ such that
$$
{
 \sup_{\vct f\in\mathcal F_{p,q}(V_T)} R_\phi^\pi(\vct f) \leq C_1\cdot \max\left\{T\cdot V_T^{2p/(4p+d)}, \log T\right\} \;\;\;\;\;\text{for noisy gradient feedback $\phi=\phi_t^{\mathcal G}(x_t,f_t)$.}
 }
$$
For the noisy function value feedback, there exists another computationally efficient policy $\pi'$ and
{ $C_2=g_2(\log T,\log V_T,d,D,\nu,L,H,M)>0$ for some function $g_2$ that is a polynomial function in $\log T$ and $\log V_T$, such that}
%$ as a polynomial function of $\log T$ such that
$$
{
\sup_{\vct f\in\mathcal F_{p,q}(V_T)} R_\phi^{\pi'}(\vct f) \leq C_2\cdot \max\left\{T\cdot V_T^{2p/(6p+d)}, \sqrt{T}\right\} \;\;\;\;\;\text{for noisy function value feedback $\phi=\phi_t^{\mathcal F}(x_t,f_t)$.}
}
$$
\label{thm:main-upper}
\end{theorem}

\begin{theorem}[Minimax lower bound  for strongly-convex function sequences]
Suppose the same conditions hold as in Theorem \ref{thm:main-upper}. Then there exists a constant
{ $C_3=g_3(d,D,\nu,L,H,M)>0$
independent of $T$ and $V_T$} such that
$$
\inf_{\pi} \sup_{\vct f\in\mathcal F_{p,q}(V_T)} R_\phi^\pi(\vct f) \geq C_3\cdot \left\{\begin{array}{ll}
{ \max\{T\cdot V_T^{2p/(4p+d)}, \log T\}}& \text{  for  }\phi=\phi_t^{\mathcal G}(x_t,f_t);\\
{ \max\{T\cdot V_T^{2p/(6p+d)}, \sqrt{T}\}}& \text{  for  }\phi=\phi_t^{\mathcal F}(x_t,f_t).\end{array}\right.
$$
\label{thm:main-lower}
\end{theorem}

In Theorem \ref{thm:main-upper}, the quantities $C_1$ and $C_2$ {depend on $T$ and $V_T$ only via  poly-logarithmic factors and these poly-log factors are usually not the focus of studying the regret.}
 %should be interpreted as ``constants'', as they are independent of $T$ and $V_T$ (up to poly-logarithmic factors)
%and are functions of other problem   parameters ($d,D,\nu,L,H,M$) that are treated as constants. %{\color{red} These constants are usually not the focus of studying the regret.}
In Theorem \ref{thm:main-lower} the quantity $C_3$ is independent of $T$ and $V_T$.  The other problem dependent parameters are treated as constants throughout the paper. The proof of Theorem \ref{thm:main-upper} is given in Sec.~\ref{sec:proof},
while the proofs of Theorem \ref{thm:main-lower} is relegated to the online supplement.
% because Theorem \ref{thm:main-lower} is a negative result
%and we only need to construct a specific difficult example.

%\yxwcomment{[Comment] This ``average change'' interpretation is only true for $q=1$. However, since the minimax regret does not depend on $q$, and $q=1$ gives the largest function class, I think we should ask the readers to ignore cases where $q\neq 1$  until they get to the lower bound proof. }

{
The condition $V_T\leq 1$ in both Theorems \ref{thm:main-upper} and \ref{thm:main-lower} is necessary for obtaining a non-trivial sub-linear regret. %is only a technical condition.
In particular, the lower bound results in Theorem \ref{thm:main-lower} show that for $V_T=\Omega(1)$, no algorithm can achieve sub-linear regret in either feedback models.
On the other hand, a trivial algorithm that outputs $x_1=\cdots=x_T=x_0$ for an arbitrary $x_0\in\mathcal X$ leads to a linear regret.
%As a result, it is not interesting to consider function sequences with $V_T>1$.

Both upper and lower regret bounds in Theorems \ref{thm:main-upper} and \ref{thm:main-lower} consist of two terms.
The $\log T$ term for $\phi_t^{\mathcal G}(x_t,f_t)$ and $\sqrt{T}$ term for $\phi_t^{\mathcal F}(x_t,f_t)$
arise from regret bounds for \emph{stationary} stochastic optimization problems (i.e., $V_T=0$),
which were proved in \cite{jamieson2012query,hazan2014beyond}.
The other terms involving polynomial dependency on $V_T$ are the main regret terms for typical dynamic function sequences
whose perturbation $V_T$ is not too small.
}

{
We also remark that the $q$ parameter does \emph{not} affect the optimal rate of convergence in Theorem \ref{thm:main-lower} (provided that $q\geq 1$ is assumed for convexity of the norms).
While this appears counter-intuitive,
{this is a property of our worst-case analytical framework, as the function sequence that leads to the worst-case regret
is the one that distributes function changes $\|f_{t+1}-f_t\|_p$ evenly across all $t\in\mathcal T$ (see for example our detailed construction of adversarial function sequences in the online supplement),
in which case the $L_{p,q}$-variation measure is the same for all $q\in[1,\infty]$.
}
%similar behaviors have been well-known in the literature of nonparametric regression \citep{donoho1998minimax},
%where the $q$ parameter in the parameterization of Besov spaces do not affect the corresponding minimax estimation rate.
%\xnote{Need to explain in more details on the connection to nonparametric regression and why we still keep $q$.}
}

\begin{remark}[Comparing with \cite{besbes2015nonstationary}]
\cite{besbes2015nonstationary} considered the special case of $p=\infty$ and $q=1$, and established the following result:
\begin{equation}
\inf_{\pi\in\mathcal P_T^\pi} \sup_{\vct f\in\mathcal F_{p,q}(V_T)} R_\phi^\pi(\vct f) \asymp \left\{\begin{array}{ll}
T\cdot V_T^{1/2},& \phi=\phi_t^{\mathcal G}(x_t,f_t)\\
T\cdot V_T^{1/3},& \phi=\phi_t^{\mathcal F}(x_t,f_t)\end{array}\right.\;\;\;\;\;\;\text{for}\;\;\;\;p=\infty, q=1.
\label{eq:old-regret}
\end{equation}
Note that in Eq.~(\ref{eq:old-regret}) we adopt a slightly different notation from \cite{besbes2015nonstationary}. In particular, the parameter $V_T$ in our paper
is $1/T$ times the parameter $V_T$ in \citep{besbes2015nonstationary}.
{ Such normalization is for presentation clarity only (to single out the $T$ term in the regret bounds).
}

It is clear that our results reduce to Eq.~(\ref{eq:old-regret}) as $p\to\infty$ for both $\phi_t^{\mathcal G}(x_t,f_t)$ and $\phi_t^{\mathcal F}(x_t,f_t)$.
In particular, for fixed domain dimension $d$ we have that $\lim_{p\to\infty}2p/(4p+d)=1/2$ and $\lim_{p\to\infty}2p/(6p+d)=1/3$,
matching regrets in Eq.~(\ref{eq:old-regret}). Therefore, the result from \cite{besbes2015nonstationary} (for strongly convex function sequences) is a special case of our results.
\end{remark}

\begin{remark}[Curse of dimensionality]
A significant difference between $p=\infty$ and $p<\infty$ settings is the \emph{curse of dimensionality}.
In particular, when $p<\infty$ the (optimal) regret depends \emph{exponentially} on dimension $d$,
while for $p=\infty$ the dependency on $V_T$ is independent of $d$ {on the exponent}.
The curse of dimensionality is a well-known phenomenon in non-parametric statistical estimation \citep{tsybakov2009introduction}.
%and also $\chi$-armed bandits \citep{bubeck2011x}, which shares a similar non-parametric function structure.
\end{remark}

Below we first introduce the policies, which is based on a ``meta-policy'' in \citet{besbes2015nonstationary}. 

%In the remainder of this section we first review policies from \cite{besbes2015nonstationary}, which are later showed to be optimal under $1\leq p<\infty$ and $1\leq q<\infty$ settings. { We also show how the upper bound in Theorem \ref{thm:main-upper} for the noisy gradient feedback $\phi_t^{\mathcal G}(x_t,f_t)$ can be generalized to the general convex case, without assuming smoothness (A4) and strong convexity (A5).

%Proof sketches for both upper and lower bounds are given in Secs.~\ref{subsec:proof-sketch-upper}, \ref{subsec:proof-sketch-lower} and \ref{subsec:proof-sketch-convex}. \xnote{The description of the proof sketch needs to modify accordingly.

\subsection{Policies}\label{subsec:algorithms}

We first describe a ``meta-policy'' proposed in \citet{besbes2015nonstationary} based on a re-starting procedure:

\vskip 0.2in
\noindent
\fbox{\begin{minipage}{0.95\textwidth}
\textsc{Meta-policy (restarting procedure)}: input parameters $T$ and $\Delta_T$; sub-policy $\pi_s$.
\begin{enumerate}
\item Divide epochs $\{1,\cdots,T\}$ into $J=\lceil T/\Delta_T\rceil$ batches $B_1,\cdots,B_J$ such that
$B_1=\{\underline b_1,\cdots,\overline b_1\}$, $B_2=\{\underline b_2,\cdots,\overline b_2\}$, etc.,
with $\underline b_1=1$, $\overline b_J=T$ and $\underline b_{\ell+1}=\overline b_{\ell}+1$ for $\ell=1,\cdots,J-1$.
The epochs are divided as evenly as possible, so that $|B_\ell|\in\{\Delta_T,\Delta_T+1\}$ for all $\ell=1,\cdots,J$.
\item For each batch $B_\ell$, $\ell=1,\cdots,J$, do the following:
\begin{enumerate}[topsep=0pt,itemsep=0ex]
\item Run sub-policy $\pi_s$ with $\underline b_\ell$ and $\overline b_\ell$, corresponding to $f_{\underline b_\ell}, f_{\underline b_\ell+1},\cdots,f_{\overline b_\ell}$.
\end{enumerate}
\end{enumerate}
\end{minipage}}
\vskip 0.2in

\noindent The key idea behind the meta-policy is to ``restart'' certain sub-policy $\pi_s$ after $\Delta_T$ epochs.
This strategy ensures that the sub-policy $\pi_s$ has sufficient number of epochs to exploit feedback information,
while at the same time avoids usage of outdated feedback information.
{ For the noisy gradient feedback $\phi_t^{\mathcal G}(x_t,f_t)$,
we set $\Delta_T=T$ if $V_T=O(T^{-(4p+d)/2p})$ and $\Delta_T\asymp V_T^{-2p/(4p+d)}$ otherwise;
for the noisy function value feedback $\phi_t^{\mathcal F}(x_t,f_t)$, we set $\Delta_T=T$ if $V_T=O(T^{-(6p+d)/4p})$
and $\Delta_T\asymp V_T^{-4p/(6p+d)}$ otherwise.
Motivations of our scalings are given in Sec.~\ref{sec:proof}
in which we prove Theorem \ref{thm:main-upper}. %\xnote{Sec 4.2 will remove the supplement}
}

The sub-policy $\pi_s$ is carefully designed to exploit information provided from different types of feedback structures.
For noisy gradient feedback $\phi_t^{\mathcal G}(x_t,f_t)$, a simple online gradient descent (OGD, see, e.g., \cite{besbes2015nonstationary,hazan2016introduction}) policy is used: %\xnote{Do we need also to cite \citep{besbes2015nonstationary}?}:

\vskip 0.2in
\noindent
\fbox{\begin{minipage}{0.95\textwidth}
\textsc{Sub-policy $\pi_s^{\mathcal G}$ (OGD)}: input parameters $\underline b_\ell,\overline b_\ell$; step sizes $\{\eta_t\}_{t=1}^T$.
\begin{enumerate}
\item Select arbitrary $x_0\in\mathcal X$.
\item For $t=0$ to $\overline b_\ell-\underline b_\ell$ do the following:
\begin{enumerate}[topsep=0pt,itemsep=0ex]
\item Suffer loss $f_{\underline b_\ell+t}(x_t)$ and obtain feedback $\hat g_t=\phi_{\underline b_\ell+t}^{\mathcal G}(x_t, f_{\underline b_\ell+t})$.
\item Compute $x_{t+1}=P_{\mathcal X}(x_t-\eta_t \hat g_t)$, where $P_{\mathcal X}(x)=\arg\min_{z\in\mathcal X}{\|z-x\|_2}$. %is the projection operator onto the convex domain $\mathcal X$.
\end{enumerate}
\end{enumerate}
\end{minipage}}
\vskip 0.2in

For noisy function value feedback $\phi_t^{\mathcal F}(x_t,f_t)$,
the classical approach is to first obtain an estimator of the gradient $\nabla f_t(x_t)$
by perturbing $x_t$ along a random coordinate $e_j=(0,\cdots,1,\cdots,0)\in\mathbb R^d$.
This idea originates from the seminal work of \cite{yudin1983problem} and was applied to convex bandits problems (e.g., \citet{flaxman2005online,besbes2015nonstationary}).
Such an approach, however, fails to deliver the optimal rate of regret when the optimal solution $x_t^*$ lies particularly close to the boundary of the domain $\mathcal X$.
Here we describe a regularized ellipsoidal (RE) algorithm from \cite{hazan2014bandit}, which attains the optimal rate of regret even when $x_t^*$ is very close to $\partial\mathcal X$.

The RE algorithm in \cite{hazan2014bandit} is based on the idea of \emph{self-concordant barriers}:
\begin{definition}[self-concordant barrier]
Suppose $\mathcal X\subseteq\mathbb R^d$ is convex and $\mathcal X^o\neq \emptyset$.
A convex function $\varphi:\mathcal X^o\to\mathbb R$ is a $\kappa$-self-concordant barrier of $\mathcal X$ if it is three times continuously differentiable on $\mathcal X^o$
and has the following properties:
\begin{enumerate}
\item For any $\{x_n\}_{n=1}^{\infty}\subseteq\mathcal X^o$, if $\lim_{n\to\infty}x_n\in\partial\mathcal X$ then $\lim_{n\to\infty}{\varphi(x_n)} = +\infty$.
\item For any $z\in\mathbb R^d$ and $x\in\mathcal X^o$ it holds that
%the following holds:
$
|\nabla^3\varphi(x)[z,z,z]|\leq 2|z^\top\nabla^2\varphi(x) z|^{3/2}$ and $
|z^\top\nabla\varphi(x)| \leq \kappa^{1/2}|z^\top\nabla^2\varphi(x) z|^{1/2},
$
where $\nabla^3\varphi(x)[z,z,z] = \frac{\partial^3}{\partial t_1\partial t_2\partial t_3}\varphi(x+t_1z+t_2z+t_3z)\big|_{t_1=t_2=t_3=0}$.
\end{enumerate}
\end{definition}

It is well-known  that for any convex set $\mathcal X\subseteq\mathbb R^d$ with non-empty interior $\mathcal X^o$,
there exists a $\kappa$-self-concordant barrier function $\varphi$ with $\kappa=O(d)$,
and furthermore for bounded $\mathcal X$ the barrier $\varphi$ can be selected such that it is strictly convex; i.e., $\nabla^2\varphi(x)\succ 0$ for all $x\in\mathcal X^o$ \citep{nesterov1994interior,boyd2004convex}.
{For example, for linear constraints $\mathcal X=\{x: Ax\leq b\}$ with $A\in\mathbb R^{m\times d}$, a logarithmic barrier function $\varphi(x)=\sum_{i=1}^m{-\log(b_i-a_i x)}$ can be used
to satisfy all the above properties (note that $a_i$ denotes the $i$-th row of $A$).} %\xnote{I changed simplex to linear?}

%\xnote{Examples on $\varphi(x)$}
We are now ready to describe the RE sub-policy that handles noisy function value feedback. %\xnote{Put a note that the original RE algorithm only deals with exact function value feedbacks. But we extend to \emph{noisy} function value feedbacks }
The policy is similar to the algorithm proposed in \cite{hazan2014bandit}, except that \emph{noisy} function value feedback is allowed in our policy, while \cite{hazan2014bandit} considered only exact function evaluations. The analysis of our policy is also more involved for dealing with noise.

\vskip 0.2in
\noindent
\fbox{\begin{minipage}{0.95\textwidth}
\textsc{Sub-policy $\pi_s^{\mathcal F}$ (RE)}: input parameters $\underline b_\ell,\overline b_\ell$; constant step size $\eta$; self-concordant barrier $\varphi$;
%Recall that ;
\begin{enumerate}
\item Select $y_0=\argmin_{y\in\mathcal X}\varphi(y)$; %where $\mathcal X^o_{\nu/T}=\{x\in\mathcal X^o: \forall z\in\mathbb B_d(\nu/T), x+z\in\mathcal X\}$;
\item For $t=0$ to $\overline b_\ell-\underline b_\ell$ do the following:
\begin{enumerate}[topsep=0pt,itemsep=0ex]
\item Compute $A_t = (\nabla^2\varphi(y_t)+\eta M (t+1) I_d)^{-1/2}$, where $I_d$ is the identity matrix in $\mathbb R^{d\times d}$. %\yxwcomment{[Comment] This matrix $B_t$ is in namespace collison with the Batches in the meta-policy.}
\item Sample $u_t$ from the uniform distribution on the unit $d$-dimensional sphere $\mathbb S_d$.
\item Select $x_{\underline b_\ell+t}=y_t+A_tu_t$; suffer loss $f_{\underline b_\ell+t}(x_{\underline b_\ell+t})$ and obtain feedback $\phi_{\underline b_\ell+t}^{\mathcal F}(x_{\underline b_\ell+t},f_{\underline b_{\ell}+t})$.
\item Compute gradient estimate $\hat g_t=d\cdot \phi_{\underline b_\ell+t}^{\mathcal F}(x_{\underline b_{\ell}+t},f_{\underline b_\ell+t})\cdot A_t^{-1}u_t$.
\item FTRL update: $y_{t+1} = \argmin_{y\in\mathcal X}\sum_{\tau=0}^t{\left\{{\hat g_\tau^\top} y + \frac{M}{2}\|y-y_\tau\|_2^2\right\}} + \eta^{-1}\varphi(y)$.
\end{enumerate}
\end{enumerate}
\end{minipage}}
\vskip 0.2in

%\xnote{We need to better explain this sub-policy in the following perspective: (1) maybe change $x_t$ to something else and let $x_{t+1}=x_t+B_tu_t$ since we claim that $f_t$ suffer the loos $f_t(x_t)$? (2) some explanations for steps $(d)$ and $(e)$? (3) Highlight the novelty as compared to \cite{hazan2014bandit} (also highlight in the introduction). }
In step 2(d), the gradient estimate $\hat g_t=d\cdot\phi_{\underline b_{\ell+t}}^{\mathcal F}(x_{\underline b_{\ell}+t},f_{\underline b_\ell+t})\cdot A_t^{-1}u_t$
satisfies $\mathbb E[\hat g_t]\approx \nabla f_{\underline b_\ell+t}(y_t)$ by the change-of-variable formula and the smoothness of $f_{\underline b_\ell+t}$. In step 2(e), instead of the projected gradient step,
a Follow-The-Regularized-Leader (FTRL) step is executed to prevent $y_{t+1}$ from being too close to the boundary of $\mathcal X$. %\xnote{maybe add some citations and background for FTRL?}
The FTRL step is essentially a mirror descent, which uses a regularization term ($\varphi(\cdot)$ in our policy) and its associated Bregman divergence to improve the convergence rates of optimization algorithms
measured in non-standard metric. {It was shown in \cite{mcmahan2014survey} (Sec.~6) that the FTRL step is equivalent to mirror descent under minimal regularity conditions.}
%We refer the readers to \cite{mcmahan2014survey} for an excellent survey on FTRL/mirror decent and their applications in online convex optimization, {which also shows that FTRL and MD are equivalent under minimal regularity conditions}.
Finally,
{ step 2(c) is a random perturbation step originally considered in \citep{hazan2014bandit}.
An important aspect of step 2(c) is the clever choice of the matrix $A_t$, which ensures the optimal regret bound even if the optimal solution $x_t^*$
is very close to the boundary of $\mathcal X$.
More specifically,
the following proposition shows that $x_{\underline b_\ell+t}=y_t+A_tu_t$ always belongs to the domain $\mathcal X$, justifying the correctness of policy $\pi_s^{\mathcal F}$.}
Its proof is given in the online supplement.
%The sub-policy is sound thanks to the following proposition, which we prove in the appendix. \xnote{Word ``sounds'' is a bit strange. Maybe should add one more sentence on explaining this: this guarantees that $x_t + B_t u_t$ belongs to $\mathcal{X}$?}
\begin{proposition}
Suppose $\varphi$ is strictly convex on $\mathcal X^o$.
Then for any $x\in\mathcal X^o$, $\delta\geq 0$ and $u\in\mathbb S_d$, $x+(\nabla^2\varphi(x)+\delta I_d)^{-1/2}u\in\mathcal X.$
\label{prop:re-sound}
\end{proposition}

{
\subsection{Extension to general convex function sequences}\label{subsec:general-convex}
In this section we show that for the noisy gradient feedback case $\phi_t=\phi_t^{\mathcal G}$,
our upper bound can be extended to \emph{general} convex functions that do not necessarily satisfy smoothness (A4) or strong convexity (A5).
The assumption (A3) that requires unique interior minimizer can also be removed.
%\xnote{ In general convex case, provide the policy before introducing the theorem (including the comparison of stepsizes).}
%\xnote{ How about the uniqueness assumption in (A3)?}
Our result is summarized in the following theorem:
\begin{theorem}[Upper bound for general convex function sequences]
Fix arbitrary $1\leq p<\infty$ and $1\leq q<\infty$.
Suppose (A1) through (A2) hold, and { $0\leq V_T\leq 1$}.
%$1/T\leq V_T\leq 1$.
%Then there exists a computationally efficient policy $\pi$ and
Also suppose that the meta-policy is carried out with the OGD sub-policy $\pi_s^{\mathcal G}$ and step sizes $\eta_t=1/\sqrt{t}$.
Then there exists
$C_4=g_4(\log T,\log V_T,d,D,\nu,H)>0$
for some function $g_4$ that is also a polynomial function in $\log T$ and $\log V_T$, such that
$$
\sup_{f\in\mathcal F_{p,q}(V_T)} R_\phi^\pi(\vct f)\leq C_4\cdot\max\left\{T\cdot V_T^{p/(3p+d)}, \sqrt{T}\right\}.
$$
\label{thm:general-convex}
\end{theorem}

%\xnote{Discuss the optimality for general convex (1) $p\rightarrow \infty$; (2) the optimality of $\sqrt{T}$}

We remark that as $p\to\infty$, the regret upper bound derived in Theorem \ref{thm:general-convex} approaches $T\cdot V_T^{1/3}$,
which matches the result in \cite{besbes2015nonstationary} for the $p=\infty,q=1$ case.
Since $T\cdot V_T^{1/3}$ is proved to be optimal for the $p=\infty,q=1$ case in \cite{besbes2015nonstationary},
this implies the optimality of our Theorem \ref{thm:general-convex} for the $p=\infty,q=1$ case as well.
However, for $1\leq p<\infty$, it is still an open question on establishing a tight minimax lower bound.
%One simple lower bound is to directly apply Theorem \ref{thm:main-lower}, which implies an $\Omega(T\cdot V_T^{2p/(4p+d)})$ regret lower bound.
%There is thus an $O(V_T^{-\frac{p(2p+d)}{(4p+d)(3p+d)}})$ multiplicative-gap between our derived upper and lower bounds for general convex functions that are not smooth or strongly convex.

%\xnote{How this is derived $O(V_T^{-\frac{p(2p+d)}{(4p+d)(3p+d)}})$?  Do you think the upper bound is not tight or lower bound is not tight?}

The structure of the proof of Theorem \ref{thm:general-convex} is similar to the one for Theorem \ref{thm:main-upper}.
It is important to note that since strong convexity is no longer assumed, the important ``affinity lemma'' cannot be proved by analyzing $\|x_t^*-x_\tau^*\|_2$. Instead, we prove another version of ``affinity lemma'' (see Lemma \ref{lem:key1} in Sec.~\ref{subsec:proof-sketch-convex}) by developing new strategies that directly bound perturbation of function values.
The proof of Theorem \ref{thm:general-convex} is provided in Sec.~\ref{subsec:proof-sketch-convex} in the supplement.

Also note that for the general convex setting with noisy function value feedback, even the case of $p=\infty$ remains a challenging open problem (see \cite{besbes2015nonstationary}); which is left as future work.}

\section{Proof of Theorem \ref{thm:main-upper}}
\label{sec:proof}

%In this section, we provide the proof sketch and illustrations of the main ideas behind the proofs for  Theorem \ref{thm:main-upper} and \ref{thm:main-lower}. The complete proofs of key lemmas are deferred to the online supplement.

{
In this section we provide the complete proof of our main positive result (upper bound) in Theorem \ref{thm:main-upper} for strongly smooth and convex function sequences $f_1,\cdots,f_T$.
Due to space constraints, the proofs of Theorems \ref{thm:main-lower} and \ref{thm:general-convex} as well as Lemma \ref{lem:stationary} are presented in the online supplement.
}

%\subsection{Proof sketch of Theorem \ref{thm:main-upper}}\label{subsec:proof-sketch-upper}

%In this section we sketch the proof of Theorem \ref{thm:main-upper}.
Our proof of Theorem \ref{thm:main-upper} is roughly divided into three steps.
In the first step, we review existing results for the OGD and the RE algorithms on upper bounding the weak regret against \emph{stationary} benchmarks. %\xnote{EGS has not been appeared, should it be changed to RE?}
In the second step, we present a novel local integration analysis that upper bounds the gap between regret against stationary and dynamic benchmarks
using the $L_p$-norm difference between two smooth and strongly convex functions.
Finally, we use a sequence of H\"{o}lder's inequality to analyze the restarting procedure in the meta-policy described in the previous section.
%and eventually prove the upper bounds in Theorem \ref{thm:main}.

\subsection{Regret against stationary benchmarks.}
For a sequence of convex functions $\vct f=(f_1,\cdots,f_{T'})$, an admissible policy $\pi$ and a feedback structure $\phi$,
the \emph{weak regret} against  any stationary point $x^* \in \mathcal{X}$ is defined as %\xnote{Should we change the word ``stationary solution'' since it is clear it is the solution of what?}
\begin{equation}
S_\phi^\pi(\vct f;x^*) := \mathbb E^\pi\left[\sum_{t=1}^{T'}{f_t(x_t)}\right] - \sum_{t=1}^{T'}{f_t(x^*)}.
\label{eq:stationary-regret}
\end{equation}
Compared to the regret against dynamic solution sequence $R_\phi^\pi$ defined in Eq.~(\ref{eq:regret}), in $S_\phi^\pi$ the benchmark solution $x^*$ is forced to be \emph{stationary}
among all $T'$ epochs, resulting in smaller regret. %\xnote{I donot quite follow ``more flexibility in the choice''? }
In fact, it always holds that $S_\phi^\pi(\vct f;x^*)\leq R_\phi^\pi(\vct f)$ for any $\vct f$ and $x^*\in\mathcal X$.
In the remainder of this section, we shall refer to $S_\phi^\pi$ as the ``weak regret'' and $R_\phi^\pi$ as the ``strong regret''.

The next lemma states existing results on upper bounding the weak regret of both OGD and RE policies for \emph{adversarial} function sequences $\vct f$. %\xnote{EGS has not appeared in our paper}
The result for OGD is folklore and documented in \cite{hazan2016introduction,besbes2015nonstationary}.
For the RE algorithm, we extend the weak regret bound in \cite{hazan2014bandit} from the exact function value feedbacks to noisy feedbacks and establish  the following lemma. The proof of Lemma \ref{lem:stationary} is deferred to Section \ref{sec:proof_lem_stationary} in the online supplement.
%the idea of using stochastic gradient estimation (steps 2(c) and 2(d)) { in online learning}
%\xnote{We should be careful here, the stochastic gradient estimation has been appeared in old literature. We should add some quantification}
%was initiated by \cite{flaxman2005online} and the application of a regularized ellipsoidal estimate appeared in \citep{hazan2014bandit}.
%In this work we further borrow the proof idea in \citep{besbes2015nonstationary} and extend the approach to noisy function evaluations,
%Due to space constraints we omit the proof of Lemma \ref{lem:stationary}. %\xnote{(1) Comment on RE in Lemma 1 and highlight we extend to the noisy function value feedbacks?}

%For completeness, we include the proof of Lemma \ref{lem:stationary} in the appendix.

\begin{lemma}
Fix $1\leq T'\leq T$.
Let $\vct f=(f_1,\cdots,f_{T'})$ be an \emph{arbitrary} sequence of smooth and strongly convex functions satisfying (A1) through (A5).
For noisy gradient feedback and the OGD policy, the following holds with $\eta_t=1/Mt$:
\begin{equation}
\label{eq:re-stationary_0}
S_\phi^\pi(\vct f;x^*) = O(\log T'), \;\;\;\;\;\;\text{for}\;\;\phi=\phi_t^{\mathcal G}(x_t,f_t),\;\; \pi=\pi_s^\mathcal G\;\;\text{and all}\;\; x^*\in\mathcal X.
\end{equation}
In addition, for noisy function value feedback and the RE policy, suppose $\varphi$ is a strictly convex $\kappa$-self-concordant barrier of $\mathcal X$,
with $\kappa=O(d)$, and $\eta=d(H+10\sigma\sqrt{\log T})/\sqrt{2T'}$. Then
\begin{equation}
S_\phi^\pi(\vct f;x^*) = O(\sqrt{T'\log T}), \;\;\;\;\;\;\text{for}\;\;\phi=\phi_t^{\mathcal F}(x_t,f_t),\;\; \pi=\pi_s^{\mathcal F}\;\;\text{and all } x^*\in\mathcal X^o_{\nu/T}.
\label{eq:re-stationary}
\end{equation}
Recall the definition that $\mathcal X^o_{\nu/T}:=\{x\in\mathcal X^o: \forall z\in\mathbb B_d(\nu/T), x+z\in\mathcal X\}$
is the strict interior of $\mathcal X$ that is at least $\nu/T$ apart from $\partial\mathcal X$. %as defined in Assumption (A3).
%\yxwcomment{[Question] Just to make sure that this is not a typo. It is $\log T$ instead of $\log T'$ because of the union bound of the subgaussian noises in Section EC3.1?}
%Recall that $x_t^*\in\mathcal X^o$ is the unique minimizer of $f_t$ and satisfies $x_t^*+u\in\mathcal X$ for all $\|u\|_2\leq\nu/T$, thanks to assumption (A2).
Also, in both results we omit dependency on $\sigma, d, D, \nu, H, L$ and $M$.
%\yxwcomment{[TODO] Edit the second part of the claim accordingly w.r.t. to the changes in the proof, once someone checked the argument through. }
\label{lem:stationary}
\end{lemma}

{
We note that when using this Lemma \ref{eq:re-stationary} in our later proofs, we will replace $x^*$ in \eqref{eq:re-stationary_0} and \eqref{eq:re-stationary} by $x_t^*$, which the is the minimizer of $f_t$. For \eqref{eq:re-stationary_0}, it is easy to see that $x^*_t \in \mathcal{X}$; and for \eqref{eq:re-stationary}, by Assumption (A3) and the definition of  $\mathcal X^o_{\nu/T}$, we have $x^*_t \in \mathcal X^o_{\nu/T}$.
%Finally, we remark that any minimizer $x_t^*$ of $f_t$ belongs to $\mathcal X^o_{\nu/T}$, thanks to Assumption (A3).
%we first show that the assumption (A3) implies $x_t^*\in\mathcal X_{\nu/T}^o$.
%This essentially reduces to checking $x_t^*$ satisfies the definition of $\mathcal X_{\nu/T}^o$, because (A3) assumes that $\{z\in\mathbb R^d:\|z-x_t^*\|_2\leq\nu/T\}\subseteq\mathcal X$, and by definition all points with such properties are included in $\mathcal X_{\nu/T}^o$. One can then apply Eq.~(8) and the fact that $x_t^*\in\mathcal X_{\nu/T}^o$  to obtain $S_\phi^\pi(x_t^*,\vct f) =O(\sqrt{T'\log T})$ for the RE policy with noisy function value feedback.
}

\subsection{Gap between weak and strong regret.}
\begin{figure}[t]
\centering
%\subfigure[]{\label{fig:fg-gap} \includegraphics[width=0.45\textwidth]{fig1.pdf}}
%\subfigure[]{\label{fig:stability} \includegraphics[width=0.45\textwidth]{fig2.pdf}}
\includegraphics[width=0.45\textwidth]{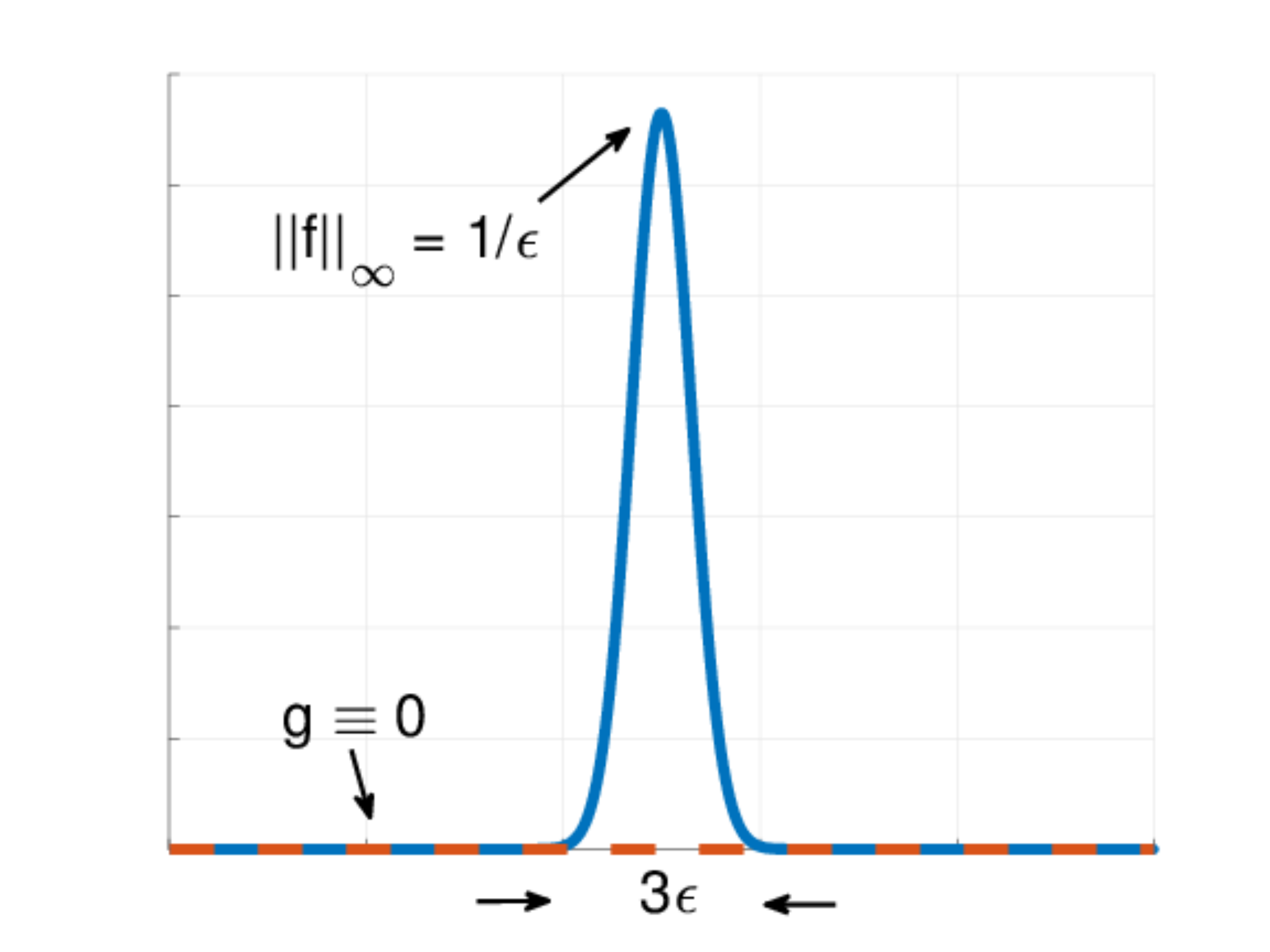}
\includegraphics[width=0.45\textwidth]{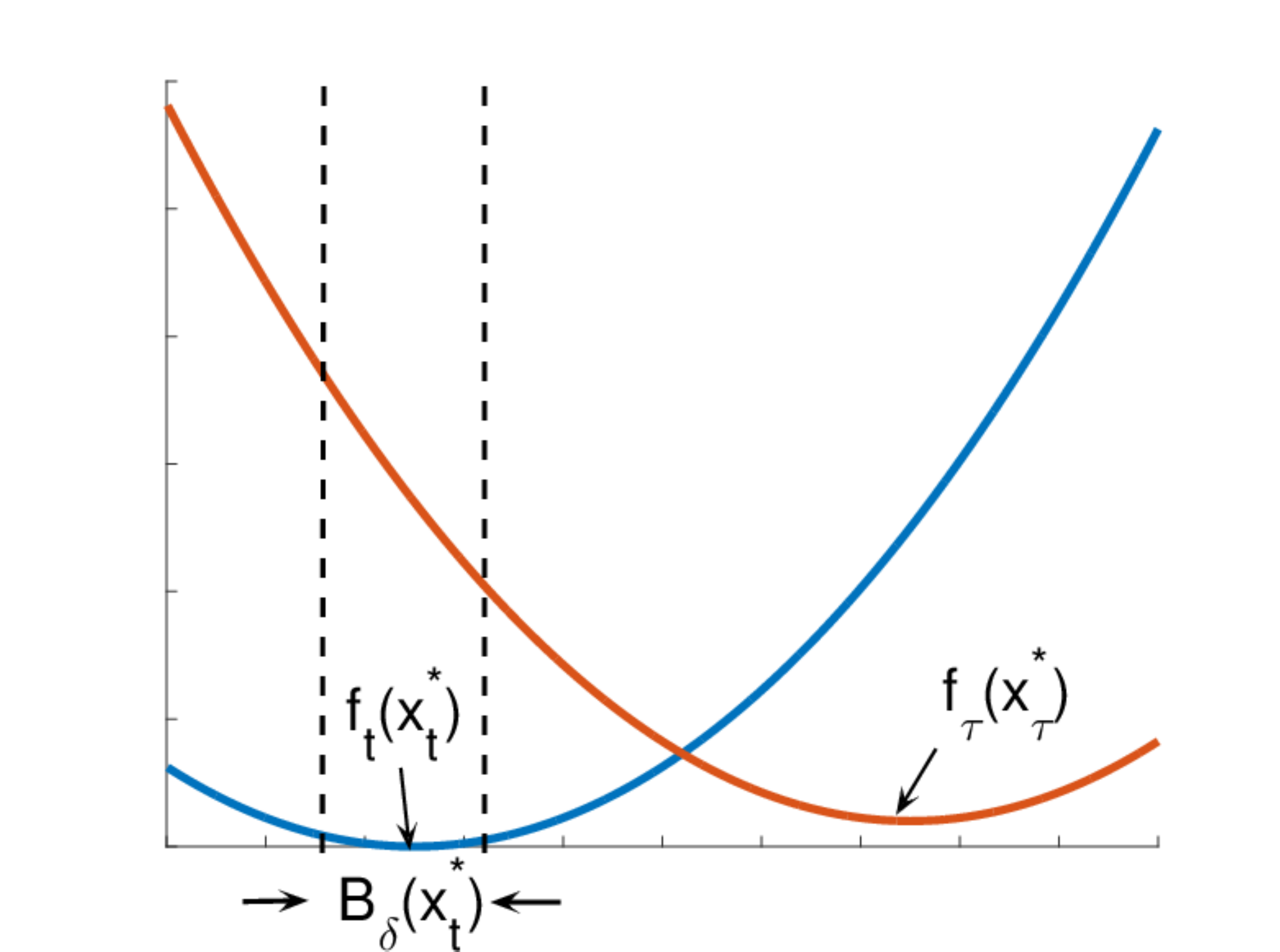}
%\begin{subfigure}[b]{\textwidth}
%\centering
%\includegraphics[width=\linewidth]{fig1.pdf}
%\caption{}
%\label{fig:fg-gap}
%\end{subfigure}
%\begin{subfigure}[b]{0.49\textwidth}
%\centering
%\includegraphics[width=\linewidth]{fig2.pdf}
%\caption{}
%\label{fig:stability}
%\end{subfigure}
\caption{The left figure illustrates how two functions $f$ and $g$ can have very different $L_p$ and $L_{\infty}$ differences ($1\leq p<\infty$).
Both functions are defined on $\mathcal X=[0,1]$, with $f(x)= \frac{1}{\sqrt{2\pi}\epsilon}\exp\left\{-\frac{(x-0.5)^2}{2\epsilon^2}\right\}$ and $g\equiv 0$.
Because $f$ is the pdf of a univariate Normal distribution with zero mean and $\epsilon^2$ variance,
$f$ and $g$ are essentially the same outside of $[0.5-3\epsilon,0.5+3\epsilon]$, leading to $\|f-g\|_{p}\leq O(\epsilon^{1/p})\cdot \|f-g\|_{\infty} = O(\epsilon^{(1-p)/p})$,
which can be  arbitrarily smaller than $\|f-g\|_{\infty}=\Omega(\epsilon^{-1})$
for $1\leq p<\infty$ and $\epsilon$ sufficiently small.
The right figure provides a graphical explanation of the key argument  in the proof of Lemma \ref{lem:key2}.
It shows that when $x_\tau^*$ is far away from $x_t^*$,
 $f_t$ and $f_\tau$ would have a large difference in a neighborhood around $x_t^*$, because of the strong convexity of $f_\tau$
and the smoothness of $f_t$.
Since such difference is upper bounded by $\|f_t-f_\tau\|_p$ on the entire domain $\mathcal X$,
one can conclude that $x_t^*$ and $x_\tau^*$ cannot be too far apart.}
\label{fig:stability}
\end{figure}

By definition, the gap between $S_\phi^\pi$ and $R_\phi^\pi$ is independent of policy $\pi$:
\begin{equation}
R_\phi^\pi(\vct f)-S_\phi^\pi(\vct f;x_\tau^*) =
 \sum_{t=1}^{T'}{f_t(x_\tau^*)-f_t(x_t^*)}, \;\;\;\;\forall\tau\in\{1,\cdots,T'\}.
\label{eq:gap}
\end{equation}
Eq.~\eqref{eq:gap} shows that it is possible to upper bound the regret gap by the two-point difference of each function $f_t$ evaluated at
the optimal solution $x_t^*$ of $f_t$ and the optimal solution $x_\tau^*$ of $f_\tau$, for arbitrary $\tau\in\{1,\cdots,T'\}$.
Such differences, however, can be large as $x_t^*$ could be far away from $x_\tau^*$ as the functions drift.
In the special case of $p=\infty$, \cite{besbes2015nonstationary} observes %the following decomposition
\begin{align}\label{eq:decomp}
f_t(x_\tau^*)-f_t(x_t^*)
= f_t(x_\tau^*)-f_\tau(x_\tau^*) + f_\tau(x_\tau^*)-f_t(x_t^*)
\leq f_t(x_\tau^*) - f_\tau(x_\tau^*) + f_\tau(x_t^*)-f_t(x_t^*)
\end{align}
and further bounds both $|f_t(x_\tau^*)-f_\tau(x_\tau^*)|$ and $|f_\tau(x_t^*)-f_t(x_t^*)|$ with $\|f_t-f_\tau\|_{\infty}$.
Such arguments, however, meet significant challenges in the more general setting when $1\leq p<\infty$,
because the difference between two functions at one point can be arbitrarily larger than the $L_p$-norm of the difference of the two functions.
We give an illustrative example in Figure \ref{fig:stability}, where two functions $f$ and $g$ are presented,  with $\|f-g\|_p/|f(x)-g(x)|\to 0$ for $x=0.5$ and $p<\infty$.  %\xnote{From Figure 1, I donot see $|f(x)-g(x)|\to\infty$.}

%To overcome such difficulties, \cite{besbes2015nonstationary} suggested the following decomposition on the right-hand side of Eq.~(\ref{eq:gap}):
%\begin{align}
%\sum_{t=1}^{T'}{f_t(x_\tau^*)-f_t(x_t^*)}
%&= \sum_{t=1}^{T'}{f_t(x_\tau^*)-f_\tau(x_\tau^*)+f_\tau(x_\tau^*)-f_t(x_t^*)}\nonumber\\
%&\leq \sum_{t=1}^{T'}{f_t(x_\tau^*)-f_\tau(x_\tau^*)+f_\tau(x_t^*)-f_t(x_t^*)}\nonumber\\
%&\leq \sum_{t=1}^{T'}{\big|f_t(x_\tau^*)-f_\tau(x_\tau^*)\big|+\big|f_\tau(x_t^*)-f_t(x_t^*)}\big|\nonumber\\
%&\leq 2T'\cdot \sup_{t,\tau\in\{1,\cdots,T'\}} \big|f_t(x_t^*)-f_\tau(x_t^*)\big|.\label{eq:decomposition}
%\end{align}
%
%Here the second inequality holds because $x_\tau^*$ is the minimizer of $f_\tau$ on $\mathcal X$.
%It is then argued in \citep{besbes2015nonstationary}, that
%in the special case of $p=\infty$, $|f_t(x_t^*)-f_\tau(x_t^*)|$ is upper bounded by $\|f_t-f_\tau\|_{\infty}$.
%A regret bound could then be derived by summing over $\|f_{t+1}-f_t\|_{\infty}$ for all epochs.

%
%Despite the obviously hard case in Figure \ref{fig:fg-gap}, there are two reasons to believe that at least under additional constraints,
%it is possible to upper bound $|f_t(x_t^*)-f_\tau(x_t^*)|$ using $\|f_t-f_\tau\|_p$.
%First, Eq.~(\ref{eq:decomposition}) only asks for an upper bound of difference  at the \emph{optimal solution $x_t^*$} of one of the two convex functions (instead of at an arbitrary point).
%To see why this is relevant, imagine $\mathcal X=[0,1]$, $f_t(x)=x^2$ and $f_\tau(x)=(x-h)^2$ for some small parameter $h>0$.
%We then have $x_t^*=0$ and $|f_t(x_t^*)-f_\tau(x_t^*)|= h^2$, which is significantly smaller than, say, $|f_t(1)-f_\tau(1)|\asymp h$.
%Perhaps more importantly, conditions (A3) and (A5) assert that the gradient as well as perturbation of gradients of both $f_t$ and $f_\tau$ are uniformly bounded,
%essentially excluding ``bad'' cases in Figure \ref{fig:fg-gap}, where the gradient of $f$ explodes near $1/2$.
%
%To make the above intuition precise, we present the following key {{affinity}} lemma that upper bounds $|f_t(x_t^*)-f_\tau(x_\tau^*)|$ as a function of $\|f_t-f_\tau\|_p$:

{In this paper we give an alternative analysis that directly upper bounds the left-hand side of Eq.~(\ref{eq:decomp}), $f_t(x_\tau^*)-f_t(x_t^*)$ (i.e., the difference of the same function $f_t$ at two points) using $\|f_t-f_\tau\|_p$,}
%provided that both $f_t$ and $f_\tau$ are smooth and strongly convex.}
The following is our key affinity lemma:
\begin{lemma}
Suppose $\mathcal X\subseteq\mathbb R^d$.
Fix $1\leq p<\infty$, $t\neq \tau$ and let $x_t^*,x_\tau^*$ be the minimizers of $f_t$ and $f_\tau$, respectively. Then under (A1) through (A5) we have that
$$
\max\left\{\big|f_t(x_t^*)-f_t(x_\tau^*)\big|, \big|f_\tau(x_t^*)-f_\tau(x_\tau^*)\big|\right\} = O\left(\|f_t-f_\tau\|_p^{r}\right) \;\;\;\;\;\text{where}\;\;\;\;r=\frac{2p}{2p+d}\in(0,1).
$$
\label{lem:key2}
\end{lemma}

\proof{Proof of Lemma \ref{lem:key2}.}
%Let $x_\tau^*$ be the minimizer of $f_\tau$.
Without loss of generality we assume $f_t(x_t^*)\leq f_\tau(x_\tau^*)$ throughout this proof.
Define $\delta=\|f_t-f_\tau\|_p^{r/2}$. We first prove that
$\|x_t^*-x_\tau^*\|_2\leq 2C \delta$, where $C=\max\{\sqrt{(4D^{d/p}+2L)/M}, 1\}$.

Assume by way of contradiction that $\|x_t^*-x_\tau^*\|_2 > 2C\delta$.
For any $x\in\mathcal X^o$ and $\alpha\in(0,1)$ define $\mathcal X_\alpha(x) := \{x + \rho(y-x): 0\leq \rho\leq\alpha, y\in\partial\mathcal X\}$.
It is easy to verify that $\mathcal X_\alpha(x)\subseteq\mathcal X$ and $\sup_{x'\in\mathcal X_\alpha(x)}\|x'-x\|_2\leq \alpha D$ (recall that $D=\sup_{y,y'\in\mathcal X}\|y-y'\|_2$ is the diameter of $\mathcal X$).
In addition, $\vol(\mathcal X_\alpha(x))\geq \alpha^d\cdot \vol(\mathcal X)$,
because $\mathcal X-x\subseteq \alpha^{-1}[\mathcal X_\alpha(x)-x]$, where $\mathcal X-x=\{z-x: z\in\mathcal X\}$ is the deflation of $\mathcal X$ by a specific vector,
and similarly $\mathcal X_\alpha(x)-x=\{z-x: z\in\mathcal X_\alpha(x)\}$.
Now set $\alpha = \delta/D$, and note that $\alpha< 1/2$ because $D\geq \|x_t^*-x_\tau^*\|_2> 2C\delta\geq 2\delta$.
By strong convexity of $f_\tau$, we have $\forall x\in \mathcal X_\alpha(x_t^*)$,
\begin{align}
f_\tau(x)
&\geq f_\tau(x_\tau^*) + \frac{M}{2}\|x_\tau^*-x\|_2^2 \geq f_t(x_t^*)+\frac{M}{2}\|x_\tau^*-x\|_2^2\label{eq:key2-ineq1}\\
&\geq f_t(x_t^*)+\frac{M}{2}\left(2C\delta-\delta\right)^2 \geq f_t(x_t^*) + \frac{MC^2}{2}\delta^2.\label{eq:key2-ineq2}
\end{align}
Here Eq.~(\ref{eq:key2-ineq1}) holds because $f_\tau(x_\tau^*)\geq f_t(x_t^*)$, and Eq.~(\ref{eq:key2-ineq2}) is true
because $\|x_t^*-x_\tau^*\|_2>2C\delta$ and $\|x-x_t^*\|_2\leq\alpha D = \delta \leq C\delta$ for all $x\in\mathcal X_\alpha(x_t^*)$.
%because $f_\tau(x_\tau^*)\geq f_t(x_t^*)$ and $\|x_t^*-x_\tau^*\|_2\geq 2C\delta\geq C\delta$.
On the other hand, by smoothness of $f_t$, we have that %for all $x\in \mathcal X_\alpha(x_t^*)$,
\begin{equation}
f_t(x)\leq f_t(x_t^*)+\frac{L}{2}\|x-x_t^*\|_2^2 \leq f_t(x_t^*)+L\delta^2\;\;\;\;\;\;\forall x\in\mathcal X_\alpha(x_t^*).
\label{eq:key2-ineq3}
\end{equation}
%where $x_\xi=x_t^*+\xi(x-x_t^*)$ for some $\xi\in(0,1)$.
Combining Eqs.~(\ref{eq:key2-ineq2},\ref{eq:key2-ineq3}) we have that, for arbitrary $1\leq p<\infty$ and $x\in\mathcal X_\alpha(x_t^*)$
\begin{equation}
\big|f_\tau(x)- f_t(x)\big|^p \geq \left|\left(f_t(x_t^*)+\frac{MC^2}{2}\delta^2\right) - \left(f_t(x_t^*) + L\delta^2\right)\right|^p \geq (MC^2/2-L)^p \delta^{2p},
\label{eq:key2}
\end{equation}
provided that $L\leq MC^2/2$, which holds true because $C\geq \sqrt{2L/M}$ by definition.
Integrating both sides of Eq.~(\ref{eq:key2}) on $\mathcal X_\alpha(x_t^*)$ and recalling the definition of $\|f_t-f_\tau\|_p$, we have that
\begin{align*}
\|f_t-f_\tau\|_p^p &= \frac{1}{\vol(\mathcal X)}\int_{\mathcal X}{|f_t(x)-f_\tau(x)|^p\ud x}
\geq \frac{1}{\vol(\mathcal X)}\int_{\mathcal X_\alpha(x_t^*)}{|f_t(x)-f_\tau(x)|^p\ud x}\\
&\geq \frac{\vol(\mathcal X_\alpha(x_t^*))}{\vol(\mathcal X)}\cdot (MC^2/2-L)^p\delta^{2p}
\geq \frac{\delta^d}{D^d}\cdot (MC^2/2-L)^p\delta^{2p}\\
&\geq \frac{(MC^2/2-L)^p}{D^d}\delta^{2p+d}
= \frac{(MC^2/2-L)^p}{D^d}\|f_t-f_\tau\|_p^p,
\end{align*}
where the last equality holds because $\delta=\|f_t-f_\tau\|_p^{r/2}$ and $(2p+d)\cdot r/2=p$.
With $C\geq \sqrt{(4D^{d/p}+2L)/M}$, we have that $(MC^2/2-L)^p/D^d\geq 2^d>1$ and hence the contradiction.

%We are now ready to prove Lemma \ref{lem:key2}.
We have now established that $\|x_t^*-x_\tau^*\|_2 \leq 2C\delta \leq O(\delta)$.
By smoothness of $f_t$ and $f_\tau$, %we have that
\begin{align*}
f_t(x_t^*) \leq f_t(x_\tau^*) &\leq f_t(x_t^*) + \frac{L}{2}\|x_t^*-x_\tau^*\|_2^2 \leq f_t(x_t^*)+O(\delta^2);\\
f_\tau(x_\tau^*) \leq f_\tau(x_t^*) &\leq f_\tau(x_\tau^*) + \frac{L}{2}\|x_t^*-x_\tau^*\|_2^2 \leq f_\tau(x_\tau^*) + O(\delta^2).
\end{align*}
The proof of Lemma \ref{lem:key2} is then completed by plugging in $\delta=\|f_t-f_\tau\|_p^{r/2}$.
\endproof

\subsection{Analysis of the re-starting procedure.}
We focus on the noisy gradient feedback $\phi_t^{\mathcal G}(x_t,f_t)$ first and briefly remark at the end of this section on how to handle noisy function value feedback.
%by slight revision of the arguments presented.
Recall that the $T$ epochs are divided into $J$ batches $B_1,\cdots,B_J$ in the meta-policy, with each batch having either $\Delta_T$ or $\Delta_T+1$ epochs.
Applying Lemmas \ref{lem:stationary}, \ref{lem:key2} together with Eq.~(\ref{eq:gap}) we have
\begin{align}
R_\phi^\pi(\vct f)
&\leq \sum_{\ell=1}^J{\inf_{\tau\in B_{\ell}}\left\{S_\phi^{\vct\pi}(f_{\underline b_{\ell}},\cdots,f_{\overline b_{\ell}};x_\tau^*) + {\sum_{t=\underline b_\ell}^{\overline b_{\ell}}{f_t(x_\tau^*)-f_t(x_t^*)}}\right\}}\nonumber\\
&\leq \sum_{\ell=1}^J{O(\log|B_\ell|) + |B_\ell|\cdot \sup_{t,\tau\in B_{\ell}} \big|f_t(x_\tau^*)-f_t(x_t^*)\big|}\nonumber\\
&\leq O\left(\frac{T}{\Delta_T}\cdot \log\Delta_T\right) + O(\Delta_T)\cdot \sum_{\ell=1}^J{\sup_{t,\tau\in B_\ell}\|f_t-f_\tau\|_p^r}\nonumber\\
&\leq  O\left(\frac{T\log T}{\Delta_T}\right) + O(\Delta_T)\cdot \sum_{\ell=1}^J{\left(\sum_{t=\underline b_\ell}^{\overline b_\ell-1}{\|f_{t+1}-f_t\|_p}\right)^r}.
\label{eq:intermediate}
\end{align}
Here the last inequality holds because (assuming without loss of generality that $\underline b_\ell\leq t\leq \tau\leq \overline b_\ell$)
$\|f_t-f_\tau\|_p \leq \sum_{k=t}^{\tau-1}{\|f_{k+1}-f_k\|_p} \leq \sum_{k=\underline b_\ell}^{\overline b_\ell-1}{\|f_{k+1}-f_k\|_p}$.

We next present another key lemma that upper bounds the critical summation term in Eq.~(\ref{eq:intermediate}) using $J$, $\Delta_T$ and $\var_{p,q}(\vct f)$.
The proof is based on consecutively applying the H\"{o}lder's inequality.
%The proof of Lemma \ref{lem:holder} is technical and involves sequential applications of H\"{o}lder's inequality for vector norms,
%which is deferred to the appendix.
\begin{lemma}
Suppose $\max_{1\leq \ell\leq J}|B_\ell|\leq\Delta_T+1$, $1\leq q\leq\infty$ and $\var_{p,q}(\vct f)\leq V_T$. Then
$$
\sum_{\ell=1}^J{\left(\sum_{t=\underline b_\ell}^{\overline b_\ell-1}{\|f_{t+1}-f_t\|_p}\right)^r}
\leq \Delta_T^{r-r/q}\cdot J^{1-r/q}\cdot T^{r/q}\cdot V_T^r. %\;\;\;\;\;\;\phi=\phi_t^{\mathcal G}(x_t,f_t).
$$
\label{lem:holder}
\end{lemma}

\proof{Proof of Lemma \ref{lem:holder}.}
By H\"{o}lder's inequality, for any $d$-dimensional vector $x$ we have that
\begin{equation}
\|x\|_\alpha\leq \|x\|_\beta\leq d^{1/\beta-1/\alpha}\|x\|_\alpha\;\;\;\;\;\;\forall\; 0<\beta\leq \alpha\leq\infty.
\label{eq:holder1}
\end{equation}
Apply Eq.~(\ref{eq:holder1}) with $\alpha=q$ and $\beta=1$ on $x=(\|f_{\underline b_\ell+1}-f_{\underline b_\ell}\|_p, \cdots, \|f_{\overline b_\ell}-f_{\overline b_\ell-1}\|_p)\in\mathbb R^{|B_\ell|-1}$:
$$
\sum_{t=\underline b_\ell}^{\overline b_\ell-1}{\|f_{t+1}-f_t\|_p} = \|x\|_1 \leq |B_\ell-1|^{1-1/q}\|x\|_q \leq \Delta_T^{1-1/q}\cdot \left(\sum_{t=\underline b_\ell}^{\overline b_\ell-1}{\|f_{t+1}-f_t\|_p^q}\right)^{1/q}.
$$
Subsequently,
\begin{equation}
\sum_{\ell=1}^J{\left(\sum_{t=\underline b_\ell}^{\overline b_\ell-1}{\|f_{t+1}-f_t\|_p}\right)^{r}}
\leq \sum_{\ell=1}^J{\Delta_T^{r-r/q}\cdot \left(\sum_{t=\underline b_\ell}^{\overline b_\ell-1}{\|f_{t+1}-f_t\|_p^q}\right)^{r/q}}.
\label{eq:holder-ineq1}
\end{equation}
We next consider $\tilde x=(\tilde x_1,\cdots,\tilde x_J)\in\mathbb R^J$, where $\tilde x_\ell=\sum_{t=\underline b_\ell}^{\overline b_\ell-1}{\|f_{t+1}-f_t\|_p^{q}}$.
Apply Eq.~(\ref{eq:holder1}) with $\alpha=1$ and $\beta=r/q$ on $\tilde x$ ($\beta<1$ because $r\in(0,1)$ and $q\geq 1$):
$$
 \left[\sum_{\ell=1}^J\left(\sum_{t=\underline b_\ell}^{\overline b_\ell-1}{\|f_{t+1}-f_t\|_p^{q}}\right)^{r/q}\right]^{q/r} = \|\tilde x\|_{r/q}\leq J^{1/\beta-1/\alpha}\cdot \|\tilde x\|_1
= J^{q/r-1}\cdot \sum_{\ell=1}^J\sum_{t=\underline b_\ell}^{\overline b_\ell-1}{\|f_{t+1}-f_t\|_p^q}.
$$
Raise both sides of the inequality to the power of $r/q$ and note that $\sum_{\ell=1}^J{\sum_{t=\underline b_\ell}^{\overline b_\ell-1}{\|f_{t+1}-f_t\|_p^q}}=\sum_{t=1}^{T-1}{\|f_{t+1}-f_t\|_p^q} \leq T\cdot V_T^q$.
We then have
\begin{equation}
\sum_{\ell=1}^J{\left(\sum_{t=\underline b_\ell}^{\overline b_\ell-1}{\|f_{t+1}-f_t\|_p}\right)^{r/q}}
\leq J^{1-r/q}\cdot \left(\sum_{\ell=1}^J{\sum_{t=\underline b_\ell}^{\overline b_\ell-1}{\|f_{t+1}-f_t\|_p^q}}\right)^{r/q}
\leq J^{1-r/q}T^{r/q}V_T^r.
\label{eq:holder-ineq2}
\end{equation}
Combining Eqs.~(\ref{eq:holder-ineq1},\ref{eq:holder-ineq2}) we proved the desired lemma.
%$$
%\sum_{\ell=1}^J{\left(\sum_{t=\underline b_\ell}^{\overline b_\ell-1}{\|f_{t+1}-f_t\|_p}\right)^r}
%\leq \Delta_T^{r-r/q}J^{1-r/q}T^{r/q}V_T^r.
%$$
\endproof

{
\subsection{Completing the proof}
We now prove Theorem \ref{thm:main-upper} by combining Lemmas \ref{lem:stationary}, \ref{lem:key2} and \ref{lem:holder}
with Eq.~(\ref{eq:intermediate}) and setting $\Delta_T$ appropriately.
First consider the noisy gradient feedback case $\phi_t^{\mathcal G}(x_t,f_t) = \nabla f_t(x_t)+\varepsilon_t$.
By Eq.~(\ref{eq:intermediate}),
$$
R_\phi^{\vct\pi}(\vct f) \leq O\left(\frac{T\log T}{\Delta_T}\right)+ O(\Delta_T)\cdot \sum_{\ell=1}^J{\left(\sum_{t=\underline b_\ell}^{\overline b_{\ell}-1}{\|f_{t+1}-f_t\|_p}\right)^r}.
$$
Subsequently invoking Lemma \ref{lem:holder} we have
$$
R_\phi^{\vct\pi}(\vct f) \leq  O(J\log T) + O(\Delta_T^{1+r-r/q}J^{1-r/q} T^{r/q}V_T^r),
$$
where $J=O(T/\Delta_T)$.
If $V_T=O(T^{-(4p+d)/2p})$, then we set $\Delta_T=T$, $J=1$ and obtain regret $O(\log T) +  O(T^{1+r}V_T^r) = O(\log T)$.
Otherwise, when $V_T=\omega(T^{-(4p+d)/2p})$, one selects $\Delta_T\asymp V_T^{-r/(r+1)} = V_T^{-2p/(4p+d)}$
and notes that $\Delta_T=o(T)$.
This yields a regret of $\widetilde O(T\cdot V_T^{2p/(4p+d)})$,
where in $\tilde O(\cdot)$ we drop poly-logarithmic dependency on $T$.

Finally we describe how the above analysis can be generalized to
the noisy function value feedback case $\phi_t^{\mathcal F}(x_t,f_t) = f_t(x_t)+\varepsilon_t$.
By Lemma \ref{lem:stationary}, $S_{\phi}^{\vct\pi}(f_{\underline b_{\ell}},\cdots,f_{\overline b_{\ell}};x_\tau^*)\leq O(\sqrt{|B_{\ell}|\log  T})\leq O(\sqrt{\Delta_T\log T})$
for $\phi=\phi_t^{\mathcal F}(x_t,f_t)$. Subsequently,
%$$
%\mathbb E\left[\sum_{t=\underline b_{\ell}}^{\overline b_{\ell}}{f_t(x_t) - f_t(x_\tau^*)}\right] \leq O(\sqrt{\Delta_T\log T}).
%$$
%Subsequently,
$$
R_{\phi}^{\vct\pi}(\vct f)\leq O(J\sqrt{\Delta_T\log T}) + O(\Delta_T^{1+r-r/q}J^{1-r/q} T^{r/q}V_T^r).
$$
%The regret decomposition in Eq.~(\ref{eq:intermediate}) should be replaced by
%$$
%R_\phi^\pi(\vct f)\leq \widetilde O(T/\sqrt{\Delta_T}) + O(\Delta_T)\cdot \sum_{\ell=1}^J{\left(\sum_{t=\underline b_\ell}^{\overline b_\ell-1}{\|f_{t+1}-f_t\|_p}\right)^r}.
%$$
%we again use a case analysis on the value of $V_T$.
If $V_T=O(T^{-(6p+d)/4p})$, then we set $\Delta_T=T$, $J=1$ and obtain regret $\tilde O(\sqrt{T}) + O(T^{1+r}V_T^r) = \tilde O(\sqrt{T})$.
Otherwise, when $V_T=\omega(T^{-(6p+d)/4p})$, one selects $\Delta_T\asymp V_T^{-2r/(2r+1)}=V_T^{-4p/(6p+d)}$ and observes that
$\Delta_T=o(T)$.
This yields a regret of $\tilde O(T\cdot V_T^{2p/(6p+d)})$.
}

{

\section{Numerical results}
\label{sec:num}
\begin{figure}[p]
\centering
\includegraphics[width=0.45\textwidth]{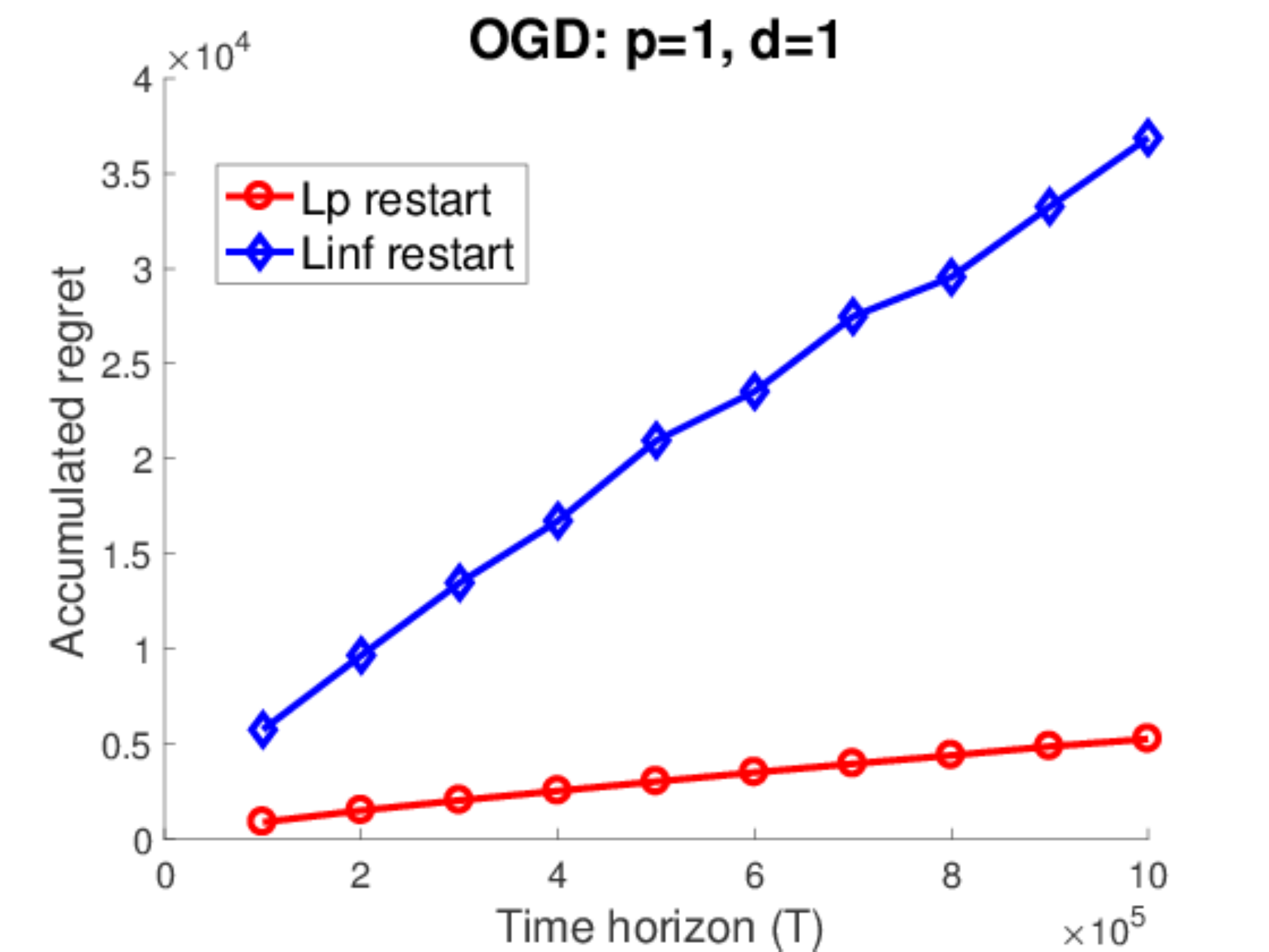}
\includegraphics[width=0.45\textwidth]{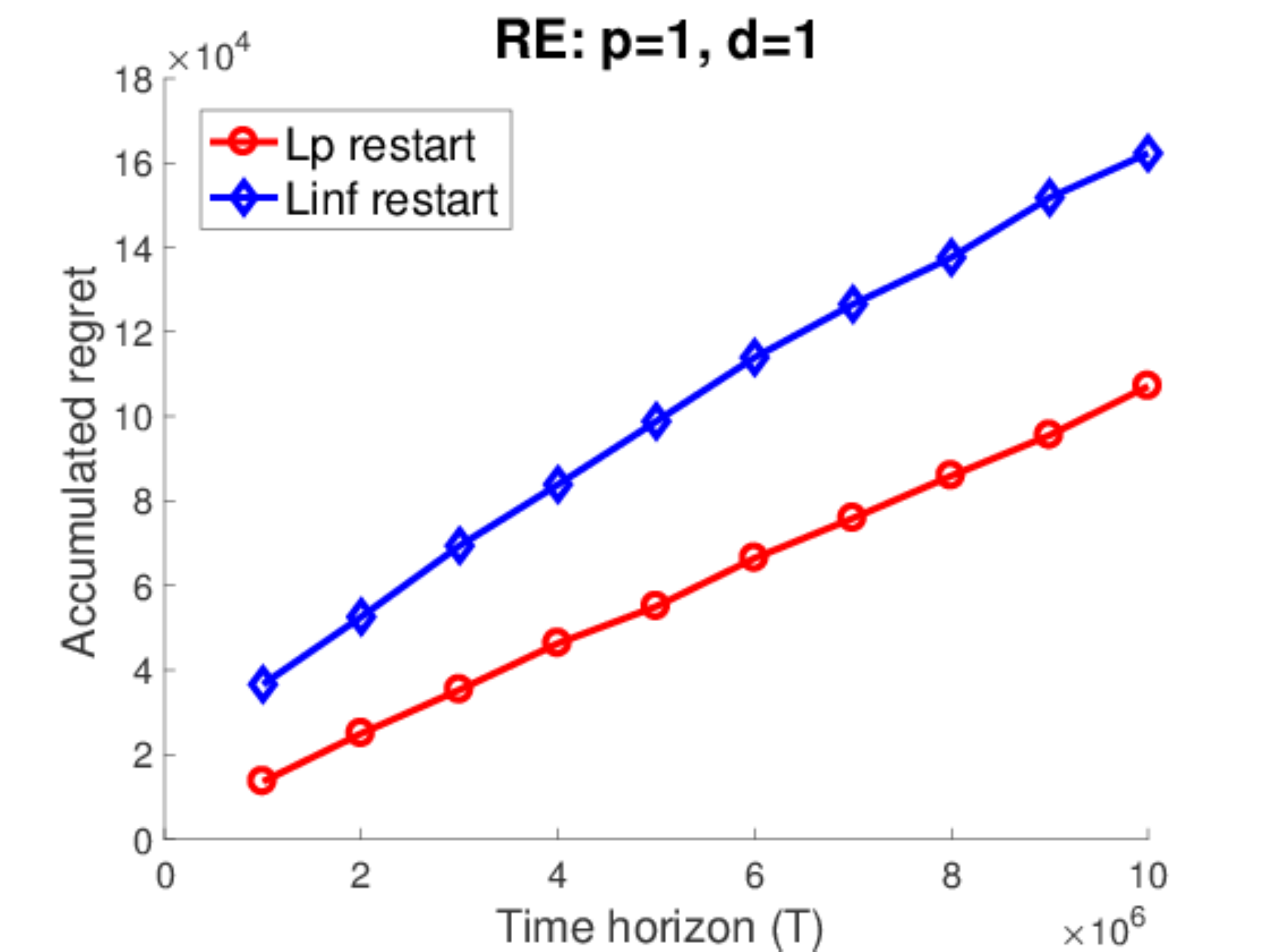}
\includegraphics[width=0.45\textwidth]{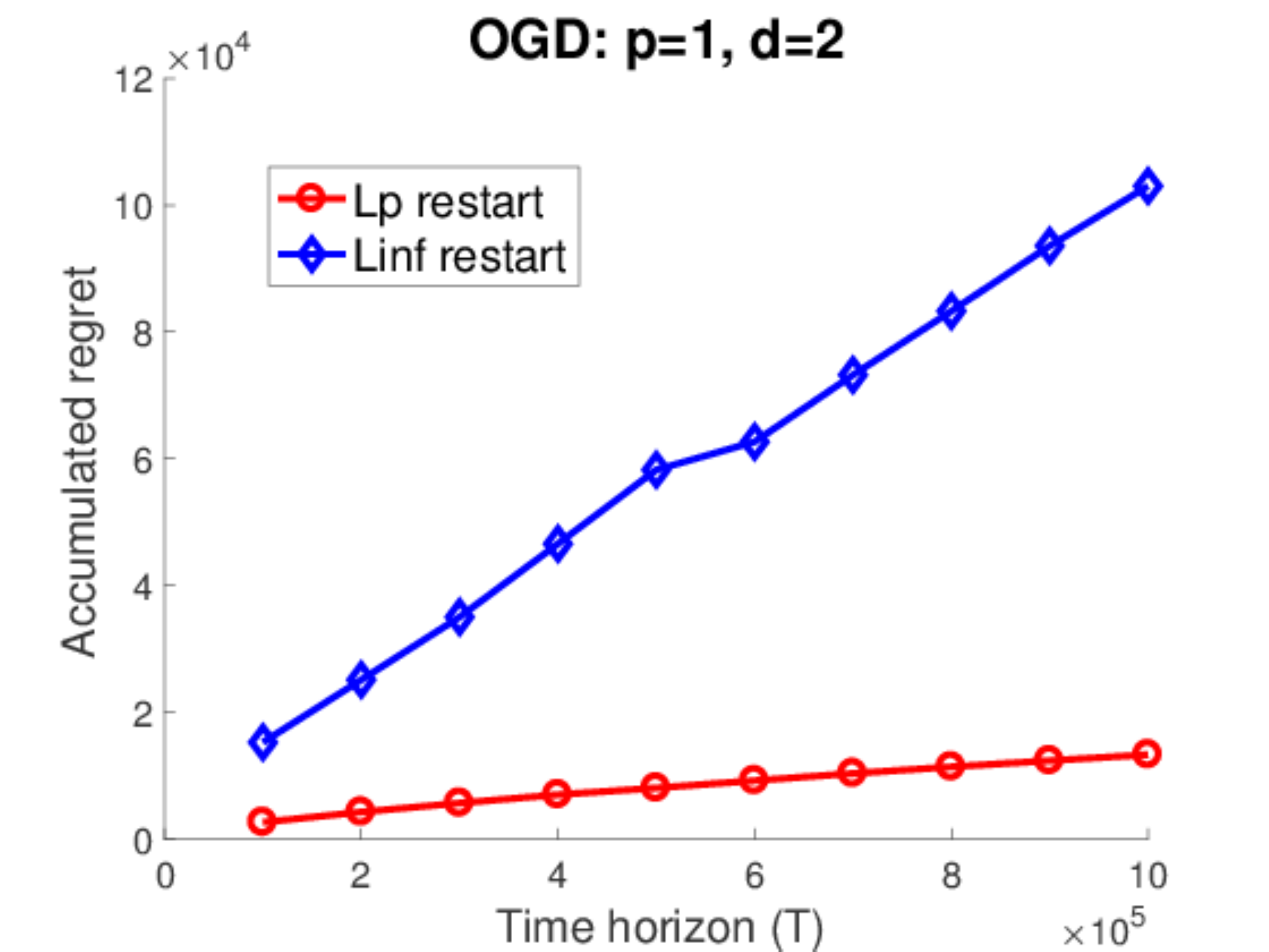}
\includegraphics[width=0.45\textwidth]{fig_egs_p1d1.pdf}
\includegraphics[width=0.45\textwidth]{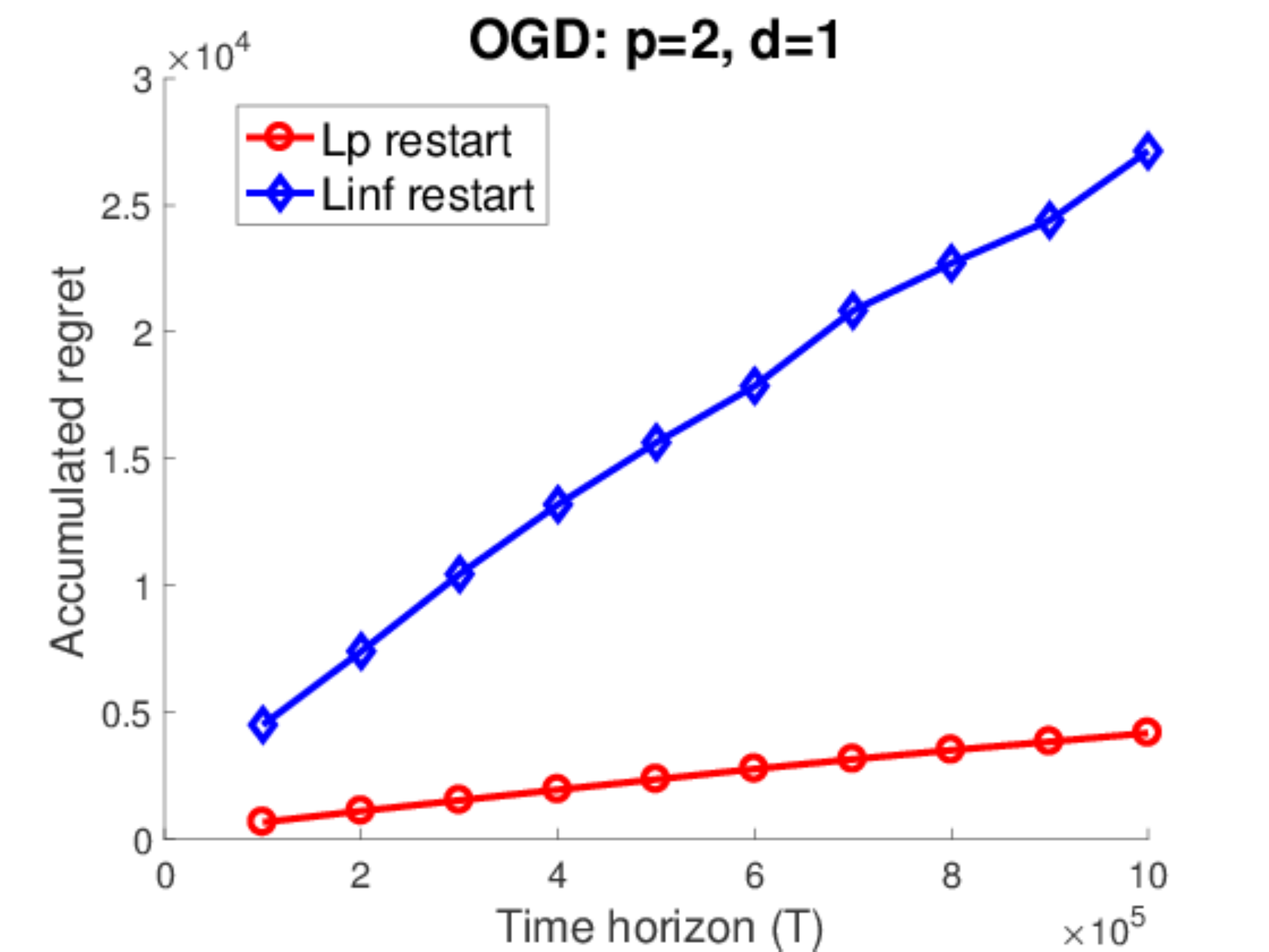}
\includegraphics[width=0.45\textwidth]{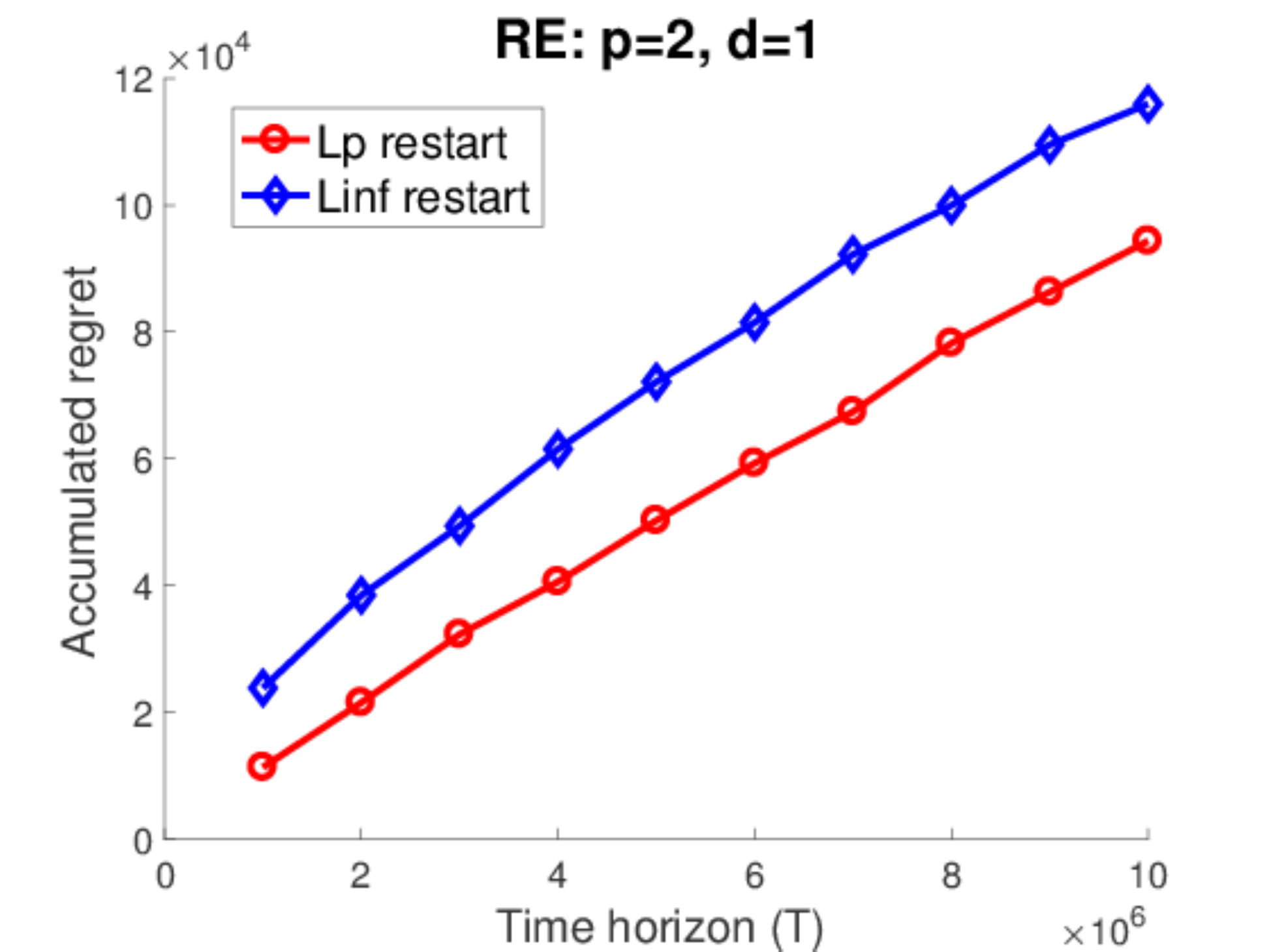}
\caption{Accumulated regret for the OGD method (left column) using noisy gradient feedback, and the RE method (right column) using noisy function value feedback.
The red curve corresponds to restarted OGD/RE with our defined $L_p$ variation measure, with $p\in\{1,2\}$ depending on how the underlying function sequence is synthesized;
the blue curve corresponds to restarted OGD/RE using only the $L_{\infty}$ variation measure in \citep{besbes2015nonstationary}.
From top to bottom three different synthesis settings of the underlying function sequence are considered: univariate functions with $L_1$ variation,
bivariate functions with $L_1$ variation, and univariate functions with $L_2$ variation.}
\label{fig:simulation}
\end{figure}

We compare our restarting procedure under $L_{p,q}$-variation measure with the algorithm in \cite{besbes2015nonstationary}.
We choose $V_T=1/T$, $q=1$, and let $p$ take values in $\{1,2\}$ to demonstrate performance in terms of the accumulated regret.
The underlying function sequence $\vct f=(f_1,\cdots,f_T)$ is constructed using the adversarial construction in Sec.~\ref{appsec:proof-lower}
which is also used to prove our lower regret bound (see Theorem \ref{thm:main-lower}).
In particular, we use the function constructions in Eqs.~(\ref{eq:F0},\ref{eq:F1},\ref{eq:construction-1d}) and its multi-variate extension in Eq.~(\ref{eq:barf}) in the online supplement.
%\xnote{Could we give either (1) more explicit pointer to the functions; or (2) repeat the formula here?}

In our simulations, both the OGD and the RE algorithm (after each restarting point) are initialized at $x_0=(1/2d,\cdots,1/2d)$,
making it at the center of the domain $\{x\in\mathbb R^d: x\geq 0, \vct 1^\top x\leq 1\}$. %\xnote{Why is $2/d$ not $1/d$?}
For OGD, the step size is set as $\eta_t=1/(t+1)$;
for the RE algorithm, we use the log-barrier function $\varphi(x) = -\sum_{i=1}^d\log(x_i) - \log(1-\vct 1^\top x)$
with step size rules $\eta=50/\sqrt{T'+1}$, where $T'$ is the number of iterations between two restarting points.
For the restarting OGD/RE algorithm with $L_p$-variation (red lines in Figure \ref{fig:simulation}),
the restarting points are selected using $V_T$ and $p$ directly (more specifically, $\Delta_T = \lceil V_T^{-2p/(4p+d)}\rceil$ or $\Delta_T = \lceil V_T^{-4p/(6p+d)}\rceil$).
For the restarting OGD/RE algorithm in \cite{besbes2015nonstationary}, we first utilize the knowledge of  the underlying function sequence to  calculate the $L_\infty$-variation measure $\tilde V_T = \frac{1}{T}\sum_{t=1}^{T-1}\|f_{t+1}-f_t\|_\infty$, and then set the restarting points using the rules $\Delta_T=\lceil \tilde V_T^{-1/2}\rceil$ or $\Delta_T=\lceil\tilde V_T^{-2/3}\rceil$
according to \cite{besbes2015nonstationary}.

%with $L_\infty$ variation (blue lines in Figure \ref{fig:simulation}),
%the restarting points are selected using the actual $L_{\infty}$ variation $\tilde V_T$ of the underlying function sequences
%and the rules $\Delta_T=\lceil\tilde V_T^{-1/2}\rceil$ or $\Delta_T=\lceil\tilde V_T^{-2/3}\rceil$ specified in \citep{besbes2015nonstationary}. \xnote{What is $\tilde{V}_T$?}

Figure \ref{fig:simulation} plots the accumulated regret of our compared algorithms for different underlying function sequences.
For the OGD algorithms with noisy gradient feedback, the time horizon ($T$) ranges from $10^5$ to $10^6$;
for the RE algorithms, we took $T$ to range from $10^6$ to $10^7$ since convergence is slower with only noisy function value feedback.
Each algorithm is given 20 independent runs and the median accumulated regret is reported.
It is observed that our algorithm (the red lines) always achieve smaller regret and outperform its competitor for constructed function sequences.

%therefore outperform the baseline methods we compare to under the specific simulation settings.

}

\section{Concluding remarks and open questions}\label{sec:conclusion}
%\xnote{Could you add this and open question might be convex case?}

We considered optimal regret of non-stationary stochastic optimization with local spatial and temporal changes.
{
An important open question is to study the optimal regret for the case of \emph{general} convex functions.
In Theorem \ref{thm:general-convex} we proved an upper regret bound of $T\cdot V_T^{p/(3p+d)}$ for general convex function sequences with access to noisy gradient oracles,
which matches the tight rate of $T\cdot V_T^{1/3}$ in \citep{besbes2015nonstationary} as $p\to\infty$.
However, for $p<\infty$, it is not clear whether our bound $T\cdot V_T^{p/(3p+d)}$ is tight even for the univariate case of $d=1$.
To further improve the learning algorithm with sharper upper bounds or to develop matching lower bound will be a future direction of research.
}
\section*{Acknowledgement}
{The authors are very grateful to three anonymous referees, the associate editor, and the area editor for their detailed and constructive comments that considerably improved the quality of this paper. We would also like to thank Prof. Assaf Zeevi for helpful discussions.
}

%%%%%%%%%%%%%%%%%

%% Here starts the e-companion (EC)
%%%%%%%%%%%%%%%%%%%%%%%%%%%%%%%%%%%%%%%%%%%%%%%%%%%%%%%%%%
%\ECSwitch

%\ECDisclaimer
%%%%%%%%%%%%%%%%%%%%%%%%%%%%%%%%%%%%%%%%%%%%%%%%%%%%%%%%%%

%%% Main head for the e-companion
\appendix

%We give complete proofs for technical lemmas whose proofs are omitted due to space constraints in the main text and the appendix.

%\section{Proofs of technical lemmas in the proof of Theorem \ref{thm:main-upper}}

%\subsection{Proof of Proposition \ref{prop:re-sound}.}\label{sec:proof_re_sound}

\section{Proof of Lemma \ref{lem:stationary}}
\label{sec:proof_lem_stationary}
To simplify notations we assume $\underline b_{\ell}=0$ throughout this proof.
We first consider the noisy gradient feedback case $\phi=\phi_t^{\mathcal G}(x_t,f_t)$.
Fix arbitrary $x^*\in\mathcal X$ and abbreviate $g_t(x_t)=\nabla f_t(x_t)$, $\hat g_t(x_t)=\phi_t^{\mathcal G}(x_t,f_t)=\nabla f_t(x_t)+\varepsilon_t$.
By strong convexity of $f_t$, we have that
\begin{equation}
2(f_t(x_t)-f_t(x^*)) \leq 2g_t(x_t)^\top (x_t-x^*) - M\|x_t-x^*\|_2^2.
\label{eq:stc-eq1}
\end{equation}
On the other hand, because $x_{t+1}=\mathcal P_{\mathcal X}(x_t-\eta_t \hat g_t(x_t))$ by definition and $\mathcal P_{\mathcal X}$ is a contraction, we have
\begin{equation*}
\|x_{t+1}-x^*\|_2^2 = \|\mathcal P_{\mathcal X}(x_t-\eta_t\hat g_t(x_t))-x^*\|_2^2 \leq \|x_t-\eta_t\hat g_t(x_t)-x^*\|_2^2.
\end{equation*}
Using Pythagorean's theorem and the fact that $\|g_t(x_t)\|_2\leq H$, $\mathbb E[\varepsilon_t|x_t]=0$ and $\mathbb E[\varepsilon_t^\top\varepsilon_t|x_t]\leq\sigma^2$, we have
\begin{equation*}
\mathbb E\left[\|x_{t+1}-x^*\|_2^2\right] \leq \mathbb E\left[\|x_t-x^*\|_2^2\right] + \eta_t^2(H^2+\sigma^2) - 2\eta_t g_t(x_t)^\top(x_t-x^*).
\end{equation*}
Subsequently,
\begin{equation}
2\mathbb E\left[g_t(x_t)^\top(x_t-x^*)\right] \leq \frac{\mathbb E[\|x_t-x^*\|_2^2] - \mathbb E[\|x_{t+1}-x_*\|_2^2]}{\eta_t} + \eta_t (H^2+\sigma^2).
\label{eq:stc-eq2}
\end{equation}
Combining Eqs.~(\ref{eq:stc-eq1},\ref{eq:stc-eq2}) and summing over $t=0,\cdots,T'$, and defining $1/\eta_{-1} := 0$, we have
\begin{align*}
2\mathbb E\left[\sum_{t=0}^{T'}{f_t(x_t)-f_t(x^*)}\right]
&\leq \sum_{t=0}^{T'}{\mathbb E\left[\|x_t-x^*\|_2^2\right]\cdot \left(\frac{1}{\eta_t}-\frac{1}{\eta_{t-1}}-M\right)} + (H^2+\sigma^2)\sum_{t=0}^{T'}{\eta_t}\\
&\leq 0 + (H^2+\sigma^2)\sum_{t=1}^{T'}{\frac{1}{M(t+1)}}\\
&= O(\log T').
\end{align*}
Because $x^*\in\mathcal X$ is arbitrary, we conclude that $S_\phi^{\pi}(\vct f;x^*)=O(\log T')$ for $\pi=\pi_S^{\mathcal G}$, $\eta_t=1/Mt$ and all $x^*\in\mathcal X$.

We next consider the noisy function value feedback case $\phi=\phi_t^{\mathcal F}(x_t,f_t)$.
Abbreviate $\hat f_t(x_t)=\phi_t^{\mathcal F}(x_t,f_t)=f_t(x_t)+\varepsilon_t$ and define, for any invertible $d\times d$ matrix $A$, that
$$
\tilde f_t^A(x) := \mathbb E_{v_t\sim\mathbb B_d}\left[f_t(x+Av)\right], \;\;\;\;\;\;\forall x\in\mathcal X,
$$
where $v_t$ is uniformly distributed on the $d$-dimensional unit ball $\mathbb B_d=\{z\in\mathbb R^d: \|z\|_2\leq 1\}$.
It is easy to verify that $\tilde f_t$ remains strongly convex with parameter $M$ and $\hat g_t=d\cdot \phi_{t}^{\mathcal F}(y_t+A_tu_t,f_{t})\cdot A_t^{-1}u_t$ is an unbiased estimator of $\nabla\tilde f_t^{A_t}(y_t)$:
\begin{align*}
\mathbb E[\hat g_t]
&= d\cdot \mathbb E\left[\phi_{t}^{\mathcal F}(y_t+A_tu_t,f_{t})\cdot A_t^{-1}u_t\right]\\
&= \mathbb E_{u_t\sim\mathbb S_d}\left[df_t(y_t+A_tu_t)A_t^{-1}u_t\right]\\
&\overset{(a)}{=} \nabla\tilde f_t^{A_t}(y_t).
\end{align*}
Here in (a) we invoke Corollary 6 of \citep{hazan2014bandit}.

Recall the definition that $x_t=y_t+A_tu_t\in\mathcal X$, the point in $\mathcal X$ at which loss is suffered and feedback is obtained.
For any $x^*\in\mathcal X^o_{\nu/T}$, decompose the regret $S_{\phi}^{\pi}(\vct f;x^*)=\mathbb E\left[\sum_{t=0}^T{f_t(x_t)-f_t(x^*)}\right]$ as
\begin{align}
S_\phi^{\pi}(\vct f;x^*)
&= \mathbb E\left[\sum_{t=0}^{T'}{f_t(x_t)-f_t(y_t)}\right] + \mathbb E\left[\sum_{t=0}^{T'}{f_t(y_t)-\tilde f_t^{A_t}(y_t)}\right]\nonumber\\
&- \mathbb E\left[\sum_{t=0}^{T'}{f_t(x^*)-\tilde f_t^{A_t}(x^*)}\right] + \mathbb E\left[\sum_{t=0}^{T'}{\tilde f_t^{A_t}(y_t)-\tilde f_t^{A_t}(x^*)}\right].
\label{eq:stc-decompose}
\end{align}

The first three terms on the right-hand side of Eq.~(\ref{eq:stc-decompose}) are easy to bound.
In particular, because $\|A_t\|_2\leq 1/\sqrt{\eta M(t+1)}$ almost surely, we have that
\begin{align*}
\mathbb E\left[f_t(x_t)-f_t(y_t)\right]
&= \mathbb E_{y_t}\mathbb E_{u_t\sim\mathbb S_d}\left[{f_t(y_t+A_tu_t)-f_t(y_t)}\bigg|y_t\right]\\
&\overset{(b)}{\leq} \frac{L}{2}\mathbb E\left[\|A_t\|_2^2\right] \leq \frac{L}{2M\eta (t+1)}.
\end{align*}
Here in (b) we use the fact that $f_t$ is smooth (A4) with parameter $G>0$.
Similarly,
\begin{align*}
-\mathbb E\left[{f_t(x^*) - \tilde f_t^{A_t}(x^*)}\right]
&= \mathbb E_{v_t\sim\mathbb B_d}\left[{f_t(x^*+A_t v_t)-f_t(x^*)}\right]
\leq \frac{L}{2M\eta (t+1)}.
\end{align*}
In addition, because $f_t$ is convex, by Jensen's inequality we have $\mathbb E[f_t(x+A_t v)]\geq f_t(x+\mathbb E[A_t v])=f_t(x)$ for all $x\in\mathcal X$ and hence
$$
\mathbb E\left[{f_t(y_t)-\tilde f_t^{A_t}(y_t)}\right]
= \mathbb E_{y_t}\mathbb E_{v_t\sim\mathbb B_d}\left[{f_t(y_t)-f_t(y_t+A_t v_t)}\bigg| y_t\right]\leq 0.
$$

We next upper bound the final term in the right-hand side of Eq.~(\ref{eq:stc-decompose}).
For $t=0,\cdots,T'$ and $a\in\mathcal X^o$ define $\|a\|_t^* := \sqrt{a^\top(\nabla^2\varphi(y_t)+\eta M(t+1)I)^{-1}a}$.
Also define $\check f_t(x) := \tilde f^{A_t}_t(y_t)+\hat g_t^\top(x-y_t)+\frac{M}{2}\|x-y_t\|_2^2$.
It is obvious that $\tilde f_t^{A_t}$ is strongly convex with parameter $M$,
and that $y_{t+1}=\argmin_{y\in\mathcal X}\sum_{\tau=0}^t{\{\check f_\tau(y)\}} + \eta^{-1}\varphi(y)$ agrees with the definition of $y_{t+1}$
in step 2(e) of sub-policy $\pi_S^{\mathcal F}$, because both $\tilde f_t^{A_t}(y_t)$ and $-g_t^\top y_t$ terms are independent of $y$ to be optimized.
We then have the following lemma, which is similar to Lemma 11 in \citep{hazan2014bandit}.
\begin{lemma}
Suppose $\eta\|\hat g_t\|_t^*\leq 1/2$ for all $t=0,\cdots,T'$. Then
\begin{equation}
\sum_{t=0}^{T'}{\check f_t(y_t)} - \sum_{t=0}^{T'}{\check f_t(x^*)} \leq 2\eta\sum_{t=0}^{T'}\left(\|\hat g_t\|_t^*\right)^2 + \frac{1}{\eta}\left(\varphi(x^*)-\varphi(y_0)\right), \;\;\;\;\;\;\forall x^*\in\mathcal X.
\label{eq:stc-key}
\end{equation}
\label{lem:stc-key}
\end{lemma}

The verification of the condition $\eta\|\hat g_t\|_t^*\leq 1/2$  and an proof of Lemma \ref{lem:stc-key} is technical and will be presented later in this section.
Taking expectations on both sides of Eq.~(\ref{eq:stc-key}) and noting that $\check f_t(y_t)=\tilde f_t^{A_t}(y_t)$ and
$\check f_t(x^*)=\tilde f_t^{A_t}(y_t)+\hat g_t^\top(x^*-y_t)+\frac{M}{2}\|x^*-y_t\|_2^2$, we have
\begin{align*}
\sum_{t=0}^{T'}{\mathbb E\left[\check f_t(y_t)-\check f_t(x^*)\right]}
&= -\sum_{t=0}^{T'}{\mathbb E\left[\hat g_t^\top(x^*-y_t) + \frac{M}{2}\|x^*-y_t\|_2^2\right]}\\
&\overset{(c)}{=} -\sum_{t=0}^{T'}{\mathbb E\left[\nabla\tilde f_t^{A_t}(y_t)^\top(x^*-y_t) + \frac{M}{2}\|x^*-y_t\|_2^2\right]}\\
&\overset{(d)}{\leq} 2\eta\sum_{t=0}^{T'}{\mathbb E\left[\|\hat g_t\|_t^{*2}\right]} + \frac{1}{\eta}\left(\varphi(x^*)-\varphi(y_0)\right).
\end{align*}
Here (c) holds because $\mathbb E[\hat g_t|y_t] = \nabla\tilde f_t^{A_t}(y_t)$ and in (d) we invoke Lemma \ref{lem:stc-key}.
%{ The condition $\eta\|\hat g_t\|_t^*\leq 1/2$ for all $t$ in Lemma \ref{lem:stc-key} is left to be verified at the end of the proof.}
%\yxwcomment{[Comment]
%	The above highlighted sentence seems redundant since it's already declared at the beginning of the paragraph.
%}

On the other hand, because $\tilde f_t^{A_t}$ is strongly convex with parameter $M$, it holds that
$\tilde f_t^{A_t}(x^*)\geq \tilde f_t^{A_t}(y_t) + \nabla\tilde f_t^{A_t}(y_t)^\top(x^*-y_t) + \frac{M}{2}\|x^*-y_t\|_2^2$ and hence
$$
-\sum_{t=0}^{T'}{\mathbb E\left[\nabla\tilde f_t^{A_t}(y_t)^\top(x^*-y_t) + \frac{M}{2}\|x^*-y_t\|_2^2\right]}
\geq \sum_{t=0}^{T'}{\mathbb E\left[\tilde f_t^{A_t}(y_t)-\tilde f_t^{A_t}(x^*)\right]}.
$$
Subsequently,
$$
\sum_{t=0}^{T'}{\mathbb E\left[\tilde f_t^{A_t}(y_t)-\tilde f_t^{A_t}(x^*)\right]} \leq 2\eta\sum_{t=0}^{T'}\mathbb E\left[\|\hat g_t\|_t^{*2}\right] + \frac{1}{\eta}\left(\varphi(x^*)-\varphi(y_0)\right).
$$
It then remains to upper bound the expectation of $\|\hat g_t\|_t^*$ and the discrepancy of self-concordant barrier functions $\varphi(x^*)-\varphi(y_0)$.
By assumption (A2), we have that $\sup_{x\in\mathcal X}|f_t(x)|\leq H$. Therefore,
$$
\mathbb E\left[\|\hat g_t\|_t^{*2}\right] = \mathbb E\left[d^2(f_t(y_t+A_tu)+\varepsilon_t)^2 u^\top A_t^{-1}(\nabla^2\varphi(y_t)+\eta M(t+1)I)^{-1}A_tu\right]
\leq d^2(H^2+\sigma^2).
$$
To bound the discrepency in self-concordance barrier terms, we cite Lemma 4 from \citep{hazan2014bandit}, which asserts that for all $x,y\in\mathcal X^o$,
$$
\varphi(y)-\varphi(x) \leq \kappa\log\frac{1}{1-\pi_x(y)},
$$
where $\kappa$ is the self-concordance parameter of $\varphi$ and $\pi_x(y)$ is defined as $\pi_x(y) := \inf\{t\geq 0: x+t^{-1}(y-x)\in\mathcal X\}$.
It is easy to verify that, for all $x\in\mathcal X^o$ and $y\in\mathcal X^o_{\nu/T}$, %such that $x+u, y+u\in\mathcal X$ for all $u\in\mathbb B_d(\nu/T)$,
$\varphi(y)-\varphi(x) = O(\kappa\log T)$ because $\sup_{x\in\mathcal X^o}\sup_{y\in\mathcal X^o_{\nu/T}}\pi_x(y) \leq 1/(1+\nu/(DT)) \leq
1-O(\nu/(DT))$. %, where we omit dependency on $\nu$ and $D=\sup_{x,y\in\mathcal X}\|x-y\|_2$.
Consequently, by Assumption (A2) and the fact that $y_0\in\mathcal X^o$ (because $\lim_{x\to\partial\mathcal X}\varphi(x)=+\infty$) we have that
\begin{equation*}
\sum_{t=0}^{T'}{\mathbb E\left[\tilde f_t^{A_t}(y_t)-\tilde f_t^{A_t}(x^*)\right]} \leq O(\eta T' + \eta^{-1}\log T'), \;\;\;\;\;\forall x^*\in\mathcal X^o_{\nu/T}.
\end{equation*}
Combining all upper bounds on the terms in Eq.~(\ref{eq:stc-decompose}) we obtain
$$
\sum_{t=0}^{T'}{\mathbb E\left[f_t(x_t)-f_t(x^*)\right]} \leq O(\eta T' + \eta^{-1}\log T'), \;\;\;\;\;\forall x^*\in\mathcal X^o_{\nu/T}.
$$
With $\eta\asymp \sqrt{\log T'/T'}$ we complete the proof of Lemma \ref{lem:stationary}.

\subsection{Verification of condition $\eta\|\hat g_t\|_t^*\leq 1/2$}

We have that $\|\hat g_t\|_t^*\leq d(H+\varepsilon_t)$.
By standard concentration results for sub-Gaussian random variables,
we know that $\sup_{0\leq t\leq T}|\varepsilon_t| \leq10\sigma\sqrt{\log T}$ with probability at least $1-O(T^{-2})$.
Let $\mathcal A$ denote the event that $\sup_{0\leq t\leq T}|\varepsilon_t|\leq 10\sigma\sqrt{\log T}$.
By the law of total expectation,
\begin{align*}
S_\phi^\pi(\vct f;x^*) &= \sum_{t=0}^{T'}{\mathbb E\left[f_t(x_t)-f_t(x^*)\big|\mathcal A\right]\Pr[\mathcal A]} + \sum_{t=0}^{T'}{\mathbb E\left[f_t(x_t)-f_t(x^*)\big|\mathcal A^c\right]\Pr[A^c]}\\
&\leq \sum_{t=0}^{T'}{\mathbb E\left[f_t(x_t)-f_t(x^*)\big|\mathcal A\right]} + 2T'H\cdot T^{-2}\\
&= \sum_{t=0}^{T'}{\mathbb E\left[f_t(x_t)-f_t(x^*)\big|\mathcal A\right]} + O(T^{-1}).
\end{align*}
Here the second inequality holds because $\Pr[\mathcal A^c]\leq T^{-2}$ and $|f_t(x_t)-f_t(x^*)|\leq 2H$ due to assumption (A3).
This term can then be upper bounded by $O(T^{-1})$ and essentially neglected in the final regret bound.
For the main term conditioned on $\mathcal A$, we have $\|\hat g_t\|_t^*\leq d(H+10\sigma\sqrt{\log T})$ for all $t=0,\cdots,T$.
Therefore, by setting $\eta = d(H+10\sigma\sqrt{\log T})/2\sqrt{T'}$ we have that $\eta\|\hat g_t\|_t^*\leq 1/2$ conditioned on $\mathcal A$.

\subsection{Proof of Lemma \ref{lem:stc-key}}

By definition of $\check f_t$, we have that $\check f_t(y_t)-\check f_t(x^*) \leq \hat g_t^\top(y_t-x^*)$.
By standard analysis of mirror descent we have that
$$
\sum_{t=0}^{T'}{\hat g_t^\top(y_t-x^*)} \leq \sum_{t=0}^{T'}{\hat g_t^\top(y_t-y_{t+1})} + \frac{1}{\eta}\left(\varphi(x^*)-\varphi(y_0)\right).
$$
It remains to upper bound $\hat g_t^\top(y_t-y_{t+1})$.
For $a\in\mathcal X^o$ define $\|a\|_t := \sqrt{a^\top(\nabla^2\varphi(y_t)+\eta M(t+1)I)a}$ and recall that
$\|a\|_t^* := \sqrt{a^\top(\nabla^2\varphi(y_t)+\eta M(t+1)I)a}$.
By definition, $\|\cdot\|_t$ and $\|\cdot\|_t^*$ are dual norms and hence by H\"{o}lder's inequality
\begin{equation}
\hat g_t^\top(y_t-y_{t+1}) \leq \|\hat g_t\|_t^*\|y_t-y_{t+1}\|_t.
\label{eq:gt-holder}
\end{equation}
Denote $\Phi_t(x) := \eta\sum_{\tau=0}^{t}{\check f_\tau(x)} + \varphi(x) = C + \eta\sum_{\tau=0}^t{\hat g_\tau^\top x} + \varphi_t(x)$,
where $C$ is a constant that does not depend on $x$ and $\varphi_t(x) := \varphi(x) + \frac{\eta M}{2}\sum_{\tau=0}^t{\|x-y_\tau\|_2^2}$.
Recall the definition of
$$
\|a\|_{x} := \sqrt{a^\top\nabla^2\Phi_t(x) a}, \;\;\|a\|_x^* := \sqrt{a^\top\{\nabla^2\Phi_t(x)\}^{-1}a} \;\;\;\;\;\;\forall x\in\mathcal X^o
$$
in classical self-concordance analysis.
It is easy to verify that $\|a\|_x = \|a\|_t$ and $\|a\|_x^* = \|a\|_t^*$,
because $\nabla^2\Phi_t(x)=\nabla^2\varphi_t(x) = \nabla^2\varphi(x) + \eta M(t+1)I$.
The following lemma is standard in analysis of Newton's method for self-concordant functions:
\begin{lemma}
Suppose $\Phi_t$ is $\kappa$-self-concordant and $\|\nabla\Phi_t(x)\|_x^*\leq 1/2$. Then
$$
\left\|x-\arg\min_{z}\Phi_t(z)\right\|_x \leq 2\|\nabla\Phi_t(x)\|_x^*.
$$
\label{lem:newton}
\end{lemma}
Note that because $\varphi$ is self-concordant, $\Phi_t$ is also self-concordant and $\nabla^2\Phi_t = \nabla^2\varphi_t$.
In addition, because $y_t$ is the minimizer of $\Phi_{t-1}$, \footnote{$\Phi_{-1}$ is defined as $\Phi_{-1}(x) = \varphi(x)$.}
$\nabla\Phi_{t-1}(y_t)=0$ and therefore
\[
\nabla\Phi_t(y_t)=\nabla\left[\Phi_{t-1}(x) + \eta\hat g_t^\top x + \frac{\eta M}{2}\|x-y_t\|_2^2\right]\big|_{x=y_t} = \hat g_t.
\]
Therefore $\|\nabla\Phi_t(x)\|_x^* = \eta\|\hat g_t\|_x^* = \eta\|\hat g_t\|_t^*$.
Invoking Lemma \ref{lem:newton} we have that, under the condition that $\eta\|\hat g_t\|_t^*\leq 1/2$,
\begin{equation}
\|y_t-y_{t+1}\|_t = \left\|y_t-\argmin_{z}\Phi_t(z)\right\|_x \leq 2\|\hat g_t\|_t^*.
\label{eq:gt-holder-2}
\end{equation}
Combining Eqs.~(\ref{eq:gt-holder},\ref{eq:gt-holder-2}) we have that $\hat g_t(y_t-y_{t+1}) \leq 2\|\hat g_t\|_t^{*2}$.
The proof is of Lemma \ref{lem:stc-key} is then complete.

\section{Proof of Theorem \ref{thm:main-lower}}\label{appsec:proof-lower}

%\subsection{Proof sketch of Theorem \ref{thm:main-lower}}\label{subsec:proof-sketch-lower}

%In this section we sketch the proof of Theorem \ref{thm:main-lower} for the lower bound results.
Let us first consider the simpler univariate case ($d=1$).
The first step is to reduce the problem of lower bounding regret to the problem of lower bounding success probability of testing
sequences of functions, for which tools from information theory such as Fano's lemma \citep{Ibragimov:Hasminskii:81book,yu1997assouad,Cover:06,tsybakov2009introduction} could be applied.
We then present a novel construction of two functions satisfying (A1) through (A5) and demonstrate
that such construction leads to matching lower bounds as presented in Theorem \ref{thm:main-lower}.
Finally, we extend the lower bound construction to multiple dimensions ($d>1$) via a change-of-variable argument
and complete the proof of general cases in Theorem \ref{thm:main-lower}.

Before introducing the proof we first give the definition of an important concept that measures the ``discrepancy'' between two functions $f,\tilde f:\mathcal X\to\mathbb R$:
$$
\chi(f,\tilde f) := \inf_{x\in\mathcal X}\max\left\{f(x)-f^*, \tilde f(x)-\tilde f^*\right\}\;\;\;\;\;\text{where}\;\;f^*=\inf_{x\in\mathcal X}f(x), \tilde f^*=\inf_{x\in\mathcal X}\tilde f(x).
$$
Intuitively, $\chi(f,\tilde f)$ characterizes the best regret $f(x)-f^*$ one could achieve \emph{without} knowing whether $f$ or $\tilde f$ is the underlying function.
This quantity plays a central role in our reduction from regret minimization to testing problems,
as well as construction of indistinguishable functions pairs.

\subsection{From regret minimization to testing.}
Consider a \emph{finite} subset $\Theta=\{\vct f_1,\cdots,\vct f_M\}\subseteq\mathcal F_{p,q}(V_T)$.
The following lemma shows that if there exists an admissible policy $\pi$ that achieves small  regret over $\mathcal F_{p,q}(V_T)$,
then it leads to a hypothesis testing procedure that identifies the true function sequence $\vct f$ in $\Theta$ with large probability:  %\xnote{Should we change ``high'' to large? since high probability usually refers to probability goes to one.}
\begin{lemma}
Fix $1\leq p<\infty$, $1\leq q\leq\infty$ and {$V_T>0$}.
Let $\phi=\phi_t^{\mathcal F}(x_t,f_t)$ or $\phi=\phi_t^{\mathcal G}(x_t,f_t)$ be either the noisy function or noisy gradient feedback.
Let $\Theta\subseteq\mathcal F_{p,q}(V_T)$ be a finite subset of sequences of convex functions.
Suppose there exists an admissible policy $\pi$ such that
\begin{equation}\label{eq:LB_R_chi}
\sup_{\vct f\in\mathcal F_{p,q}(V_T)} R_\phi^\pi(\vct f) \leq \frac{1}{9}\cdot \inf_{\vct f,\tilde{\vct f}\in\Theta}\sum_{t=1}^T{\chi(f_t,\tilde f_t)},
\end{equation}
then there exists an estimator $\{\phi_t(x_t,f_t)\}_{t=1}^T\mapsto \hat{\vct f}$ such that
\begin{equation}\label{eq:LB_R_chi_2}
\sup_{\vct f\in\Theta} \Pr_{\vct f}\left[\vct f\neq\hat{\vct f}\right] \leq 1/3,
\end{equation}
where $\Pr_{\vct f}$ denotes the probability distribution parameterized by the underlying true function sequence $\vct f\in\Theta$.
\label{lem:reduction}
\end{lemma}
%\xnote{Shall we define $\Pr_{\vct f}$? }
The proof of Lemma \ref{lem:reduction} is technical and given later.
At a higher level, when there exists an admissible policy $\pi$ that achieves small  regret over $\mathcal F_{p,q}(V_T)$ (and hence small regret over $\Theta\subseteq\mathcal F_{p,q}(V_T)$ too),
then one can correctly identify the underlying function sequence $\vct f\in\Theta$ with large probability by searching all function sequences in $\Theta$
and selecting the one that has the smallest regret. %\xnote{Should we change ``high'' to large? since high probability usually refers to probability goes to one.}
%The fact that $\sum_{t=1}^T{\chi(f_t,\tilde f_t)}$ is large for any pair $\vct f,\tilde{\vct f}\in\Theta$ ensures that such selection is unique,
%provided that the regret of $\pi$ is sufficiently small. \xnote{Why unique?}

Reduction to testing is a standard approach for proving minimax lower bounds in stochastic estimation and optimization problems \citep{besbes2015nonstationary,agarwal2010information,raskutti2011minimax}.
Motivations behind such reduction are a well-established class of tools that provide lower bounds on failure probability in testing problems \citep{yu1997assouad,Ibragimov:Hasminskii:81book,tsybakov2009introduction}.  Let
$\kl(P\|Q)=\int \log \frac{\mathrm d P}{\mathrm d Q} \mathrm d P$ denote the Kullback-Leibler divergence between two distributions $P$ and $Q$. %throughout the paper.
We introduce the following version of the Fano's inequality,%, which suffices for our application.
\begin{lemma}[Fano's inequality]
Let $\Theta=\{\theta_1,\cdots,\theta_M\}$ be a finite parameter set of size $M$.
For each $\theta\in\Theta$, let $P_{\theta}$ be the distribution of observations parameterized by $\theta$.
Suppose there exists $0<\beta<\infty$ such that $\kl(P_{\theta}\|P_{\theta'})\leq\beta$ for all $\theta,\theta'\in\Theta$.
Then
\begin{equation}
\inf_{\hat\theta}\sup_{\theta\in\Theta}\Pr_{\theta}\left[\hat\theta\neq\theta\right] \geq 1 - \frac{\beta+\log 2}{\log M}.
\label{eq:fano}
\end{equation}
\label{lem:fano}
%\xnote{KL needs to be defined here.}
\end{lemma}
%\xnote{Add a bit on the importance and difficulty of the construction.}
With Lemmas \ref{lem:reduction} and \ref{lem:fano}, the question of proving Theorem \ref{thm:main-lower}
is reduced to finding a ``hard'' subset $\Theta\subseteq\mathcal F_{p,q}(V_T)$ such that the minimum discrepancy $\inf_{\vct f,\tilde{\vct f}\in\Theta}\sum_{t=1}^T{\chi(f_t,\tilde f_t)}$ is lower bounded and the maximum KL divergence $\sup_{\vct f,\tilde{\vct f}\in\Theta}\kl(P_{\vct f}\|P_{\tilde{\vct f}})$ is upper bounded.  More precisely, the upper bound on the maximum KL divergence will provide a lower bound for right hand side of Eq. \eqref{eq:fano}, which contradicts Eq. \eqref{eq:LB_R_chi_2} in Lemma \ref{lem:reduction}. Therefore, the inequality in \eqref{eq:LB_R_chi} will not hold, which implies a lower bound on the regret. The construction of such a ``worst-case example'' $\Theta$ is highly non-trivial and involves complex design of cubic splines, as we explain in Figure \ref{fig:lowerbound} and the next paragraph.
Below we first give such a construction for the univariate ($d=1$) case and later extend the construction to higher dimensions.

\subsection{Univariate constructions.} Fix $\mathcal X=[0,1]$ and $1/8T^2\leq h\leq 1/8$.
Define $F_0,F_1:\mathcal X\to\mathbb R$ as follows:
\begin{align}
F_0(x) &:= \left\{\begin{array}{ll}
x^2,& 0\leq x<\sqrt{h};\\
\frac{4}{\sqrt{h}}x^3 - 11x^2 + 12\sqrt{h}x - 4h,& \sqrt{h}\leq x<2\sqrt{h};\\
8(x-\sqrt{h})^2,& 2\sqrt{h}\leq x\leq 1.
\end{array}\right.\label{eq:F0}\\
F_1(x) &:= \left\{\begin{array}{lll}
(x-\sqrt{h})^2,& 0\leq x<\sqrt{h};\\
8(x-\sqrt{h})^2,& \sqrt{h}\leq x\leq 1.
\end{array}\right.\label{eq:F1}
\end{align}

\begin{figure}[t]
\centering
\subfigure[]{\label{fig:lowerbound} \includegraphics[width=0.47\textwidth]{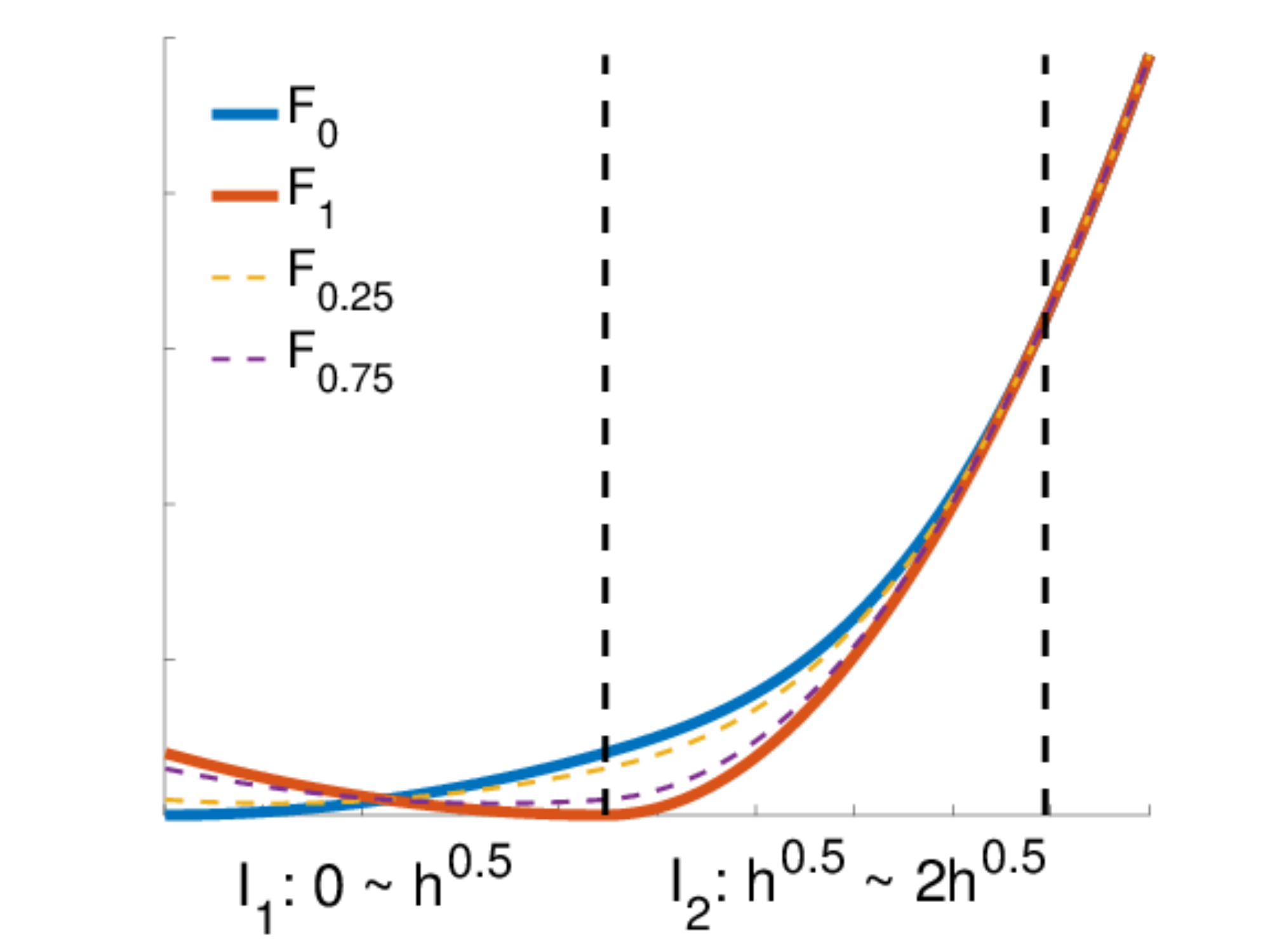}}
\subfigure[]{\label{fig:smooth} \includegraphics[width=0.47\textwidth]{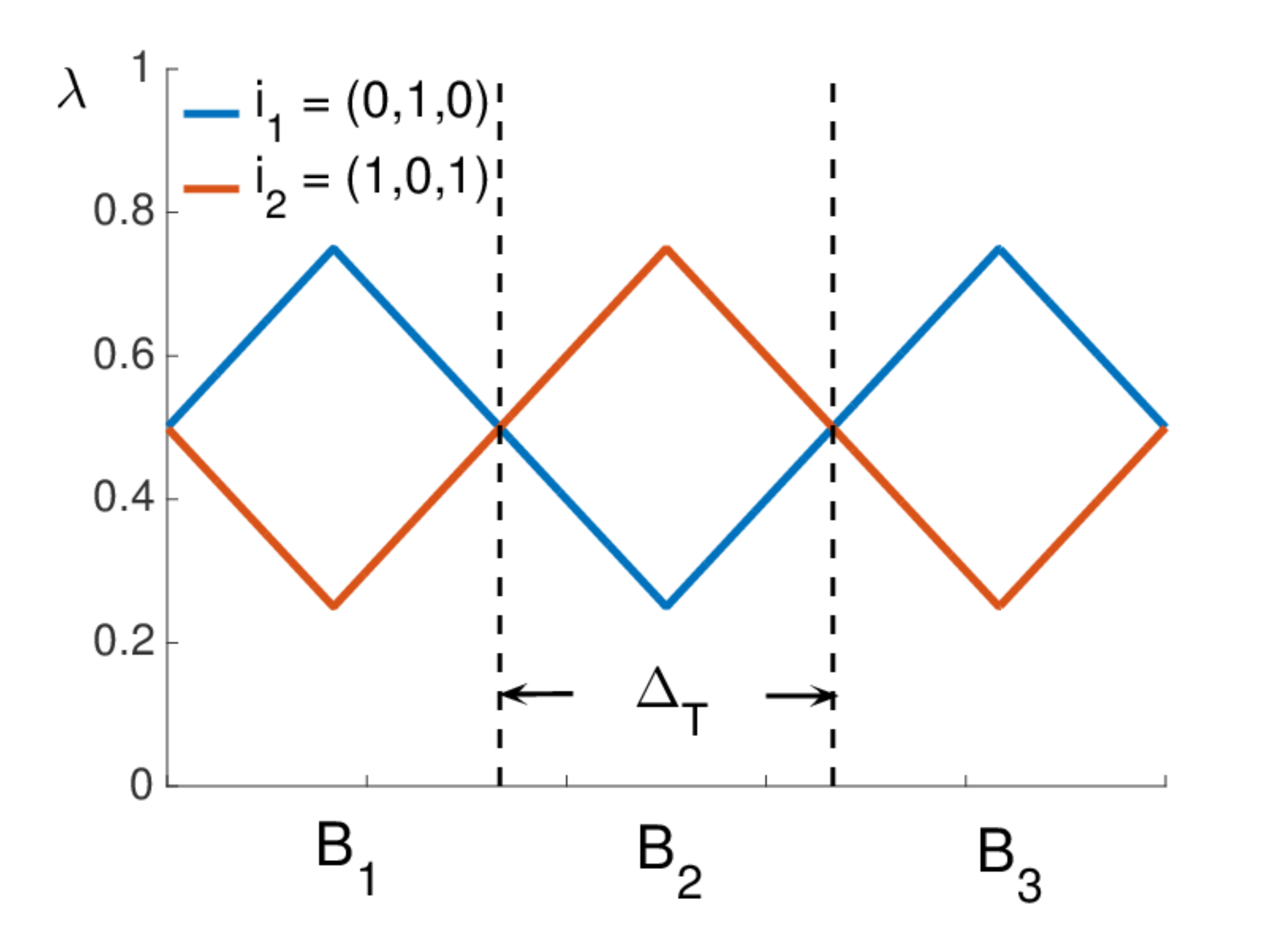}}
%\begin{subfigure}{0.49\textwidth}
%\centering
%\includegraphics[width=\linewidth]{fig3.pdf}
%\caption{}
%\label{fig:lowerbound}
%\end{subfigure}
%\begin{subfigure}{0.49\textwidth}
%\centering
%\includegraphics[width=\linewidth]{fig4.pdf}
%\caption{}
%\label{fig:smooth}
%\end{subfigure}
\caption{\small Figure \ref{fig:lowerbound} gives a graphical depiction of functions constructed in the lower bound, with thick solid lines corresponding to $F_0$ and $F_1$,
and thin dashed lines corresponding to $F_{\lambda}$ with intermediate values $\lambda=0.25$ and $\lambda=0.75$.
For the sake of better visualization, the regions $0\leq x\leq\sqrt{h}$ ($I_1$) and $\sqrt{h}\leq x\leq 2\sqrt{h}$ ($I_2$) are greatly exaggerated.
%{\bf [NOTE: Fix the x labels in 1(a)! should be $h^{0.5}$ instead of $h^{0.25}$]}
In the actual construction both regions are very small compared to the entire domain $\mathcal X=[0,1]$.
Figure \ref{fig:smooth} shows the two constructions of function sequences $\vct f$ on $J=3$ batches, according to Eq.~(\ref{eq:construction-1d}).
At the beginning and the end of each batch the function is always $F_{0.5}$, while within each batch the values of $\lambda$ first increase and then decrease, or
vice versa, depending on the coding $i_j\in\{0,1\}$ for the particular batch.
Also note that $\lambda$ will never be over $0.75$ nor under $0.25$ throughout the entire construction of the function sequence.}
%\xnote{Yellow color line is hard to observe.}
\end{figure}

Further define
\begin{equation}
F_{\lambda} := F_0 + \lambda (F_1-F_0), \;\;\;\;\;\;\lambda\in[0,1]
\label{eq:flambda}
\end{equation}
as a convex combination of $F_0$ and $F_1$.
Figure \ref{fig:lowerbound} gives a graphical sketch of $F_0$, $F_1$ and $F_\lambda$.
The key insight in the constructions of $F_0$ and $F_1$ is to use a cubic function to connect two quadratic functions of different curvatures,
and hence allow $F_\lambda$ to be the same on a wide region of $\mathcal X$ (in particular $[2\sqrt{h},1]$)
and produce small $L_p$ difference $\|F_0-F_1\|_p$. %\xnote{What do you mean by to be the same?}
In contrast, the lower bound construction in existing work \citep{besbes2015nonstationary} uses quadratic functions only,
which are not capable of producing smooth functions that differ locally and therefore only applies to the special case of $p=\infty$.

The following lemma lists some properties of $F_\lambda$. Their verification is left to Section \ref{sec:supp_lem_properties} in the online supplement.
\begin{lemma}
The following statements are true for all $\lambda,\mu\in[1/4,3/4]$.
\begin{enumerate}
\item $F_\lambda$ satisfies (A1) through (A5) with $D=2$, $\nu=1/64$, $H=16$, $L=26$ and $M=2$.
\item $\|F_\lambda-F_\mu\|_{\infty}\leq |\lambda-\mu|\cdot O(h)$ and $\|F_\lambda'-F_\mu'\|_{\infty}\leq |\lambda-\mu|\cdot O(\sqrt{h})$.
\item $\|F_\lambda-F_\mu\|_{p} \leq |\lambda-\mu|\cdot O(h^{(2p+1)/2p})$ for all $1\leq p<\infty$.
\item $\chi(F_\lambda, F_{1-\lambda}) = |1/2-\lambda|^2\cdot h/4$.
\end{enumerate}
\label{lem:properties}
\end{lemma}

We are now ready to describe our construction of a ``hard'' subset $\Theta\subseteq\mathcal F_{p,q}(V_T)$.
Note that $\mathcal F_{p,\infty}(V_T)\subseteq\mathcal F_{p,q}(V_T)$ for all $1\leq q\leq\infty$ due to the monotonicity of $\var_{p,q}(\vct f)$ (see Proposition \ref{prop:monotonicity}). %\xnote{correct this}
Therefore we shall focus solely on the $q=\infty$ case, whose construction is automatically valid for all $1\leq q\leq\infty$.

Let $1\leq J\leq T$ be a parameter to be determined later, and define $\Delta_T=\lfloor T/J\rfloor$.
Again partition the entire $T$ time epochs into $J$ disjoint batches $B_1,\cdots,B_J$, where each batch consists of either $\Delta_T$ or $\Delta_{T}+1$ consecutive epochs.
Let $\{0,1\}^J$ be the class of all binary vectors of length $J$ and let $\mathcal I\subseteq\{0,1\}^J$ be a certain subset of $\{0,1\}^J$ to be specified later.
The subset $\Theta\in\mathcal F_{p,\infty}(V_T)$ is constructed so that each function sequence $\vct f_{\vct i}\in\Theta$ is indexed by a unique $J$-dimensional
binary vector $\vct i\in\mathcal I$, with $\vct f_{\vct i}=(f_{\vct i,1},\cdots,f_{\vct i,T})$ defined as
\begin{equation}
f_{\vct i,(j-1)\Delta_T+\ell}=\left\{\begin{array}{ll}
 F_{0.5+0.5\ell/|B_j|},& \vct i_j=0, 1\leq \ell\leq\lfloor|B_j|/2\rfloor;\\
 F_{0.75-0.5\ell/|B_j|},& \vct i_j=0, \lfloor|B_j|/2\rfloor<\ell\leq|B_j|;\\
 F_{0.5-0.5\ell/|B_j|},& \vct i_j=1, 1\leq \ell\leq\lfloor|B_j|/2\rfloor;\\
 F_{0.25+0.5\ell/|B_j|},& \vct i_j=1, \lfloor|B_j|/2\rfloor<\ell\leq|B_j|	.
\end{array}\right.\;\;\;\;\;\;1\leq j\leq J.
\label{eq:construction-1d}
\end{equation}
%\xnote{The index has some problem: it should not be $(j-1)T +\ell$.}

Figure \ref{fig:smooth} gives a visual illustration of the change pattern of $\vct f_{\vct i}$ and $\vct f_{\vct i'}$ by plotting the values of $\lambda$ for each function
in the constructed sequences.
For a particular batch $B_j$, when $\vct i_j=\vct i'_j$ then $\vct f_{\vct i}$ and $\vct f_{\vct i'}$ are exactly the same within $B_j$;
on the other hand, if $\vct i_j=0$ then $\vct f_{\vct i}$ will drift towards the function $F_0$ and if $\vct i_j'=1$ the functions $\vct f_{\vct i'}$ will drift towards $F_1$,
creating gaps between $\vct f_{\vct i}$ and $\vct f_{\vct i'}$ within batch $B_j$.
For regularity reasons, we constrain the $\lambda$ value to be within the range of $(0.25, 0.75)$ regardless of $\vct i_j$ values.
We also note that $\vct f_{\vct i}$ and $\vct f_{\vct i'}$ always agree on the first and the last epochs within each batch.
This property makes repetition of constructions across all $J$ batches possible.
The following lemma lists some key quantities of interest between $\vct f_{\vct i}$ and $\vct f_{\vct i'}$:
\begin{lemma}
Suppose $\varepsilon_t\overset{i.i.d.}{\sim}\mathcal N(0,1)$ for $\phi=\phi_t^{\mathcal F}(x_t,f_t)$
and $\varepsilon_t\overset{i.i.d.}{\sim}\mathcal N_d(0,I)$ for $\phi=\phi_t^{\mathcal G}(x_t,f_t)$.
For any $\vct i,\vct i'\in\{0,1\}^J$ consider $\vct f_{\vct i}$ and $\vct f_{\vct i'}$ as defined in Eq.~(\ref{eq:construction-1d}).
Then the following statements are true:
\begin{enumerate}
\item (Variation). $\var_{p,q}(\vct f)\leq \var_{p,\infty}(\vct f)\leq O(h^{(2p+1)/2p}/\Delta_T)$, for all $1\leq p<\infty$ and $1\leq q\leq\infty$.
\item (Discrepancy). $\sum_{t=1}^T{\chi(\vct f_{\vct i,t},\vct f_{\vct i',t})} \geq  \Delta_H(\vct i,\vct i')\cdot \Omega(h\Delta_T)$,
where $\Delta_H(\vct i,\vct i')=\sum_{j=1}^J{\mathbb I[\vct i_j\neq \vct i'_j]}$ is the Hamming distance between $\vct i$ and $\vct i'$.
\item (KL divergence).
Let $P_{\vct f_{\vct i}}^{\phi,\pi}$ be the distribution of $\{\phi_t(x_t,f_{\vct i,t})\}_{t=1}^T$,
with $\{x_t\}_{t=1}^T\subseteq\mathcal X$ selected by an admissible policy $\pi$.
Then for any such policy $\pi$ we have that
$$
\kl(P_{\vct f_{\vct i}}^{\phi,\pi}\| P_{\vct f_{\vct i'}}^{\phi,\pi}) \leq \left\{\begin{array}{ll}
\Delta_H(\vct i,\vct i')\cdot O(h\Delta_T),& \phi=\phi_t^{\mathcal G}(x_t,f_t);\\
\Delta_H(\vct i,\vct i')\cdot O(h^2\Delta_T),& \phi=\phi_t^{\mathcal F}(x_t,f_t).\end{array}\right.
$$
\end{enumerate}
\label{lem:key-1d}
\end{lemma}
The proof of Lemma \ref{lem:key-1d} is deferred to Section \ref{sec:proof_lem_key_1d} in the online supplement.

Finally, we describe the construction of $\mathcal I\subseteq\{0,1\}^J$ and the choices of $J,\Delta_T$ and $h$ that give rise to matching lower bounds.
For simplicity we restrict ourselves to $J$ being an even number.
The construction of $\mathcal I$ is based on the concept of \emph{constant-weight codings}, where each code $\vct i\in\mathcal I$ has \emph{exactly} $J/2$ ones
and $J/2$ zeros, and each pair of codes $\vct i,\vct i'\in\mathcal I$ have large Hamming distance $\Delta_H(\vct i,\vct i')\geq J/16$.
The construction of constant-weight codings originates from \citep{graham1980lower}, and \cite{wang2016noiseadaptive} gave an explicit lower bound
on the size of $\mathcal I$, which we cite below:
\begin{lemma}[\cite{wang2016noiseadaptive}, Lemma 9]
Suppose $J\geq 2$ and $J$ is even.
There exists a subset $\mathcal I\subseteq\{0,1\}^J$ such that $\forall\vct i\in\mathcal I$, $\sum_{j=1}^J{\vct i_j}=J/2$,
and $\forall\vct i,\vct i'\in\mathcal I$, $\Delta_H(\vct i,\vct i')\geq J/16$.
Furthermore, $\log|\mathcal I|\geq 0.0625J$.
\label{lem:coding}
\end{lemma}

The univariate case of Theorem \ref{thm:main-lower} can then be proved by appropriately setting the scalings of $h$, $\Delta_T$
and invoking Lemmas \ref{lem:properties}, \ref{lem:key-1d} and \ref{lem:coding}.
{
Because $\Omega(\log T)$ and $\Omega(\sqrt{T})$ regret lower bounds for \emph{stationary} stochastic online optimization are known (see, for example, \citep{hazan2014beyond,jamieson2012query}),
we only need to prove the lower bound with the additional assumption that $V_T=\Omega(T^{-(4p+d)/2p})$ for noisy gradient feedback $\phi_t=\phi_t^{\mathcal G}(x_t,f_t)$
and/or $V_T=\Omega(T^{-(6p+d)/4p})$ for noisy function value feedback $\phi_t=\phi_t^{\mathcal F}(x_t,f_t)$.
More specifically, for noisy gradient feedback $\phi_t^{\mathcal G}(x_t,f_t)$ we set $h\asymp V_T^{2p/(4p+1)}$, $\Delta_T\asymp V_T^{-2p/(4p+1)}$,
and for noisy function feedback $\phi_t^{\mathcal F}(x_t,f_t)$ we set $h\asymp V_T^{2p/(6p+1)}$ and $\Delta_T\asymp V_T^{-4p/(6p+1)}$.
It is easy to verify that with the additional lower bound on $V_T$, $\Delta_T=o(T)$ and $h\gtrsim 1/T^2$, and therefore the constructions are valid.
A complete proof is given in Sec.~\ref{subsec:lower-highd} after we introduce our adversarial construction of $d>1$, which includes the univariate setting ($d=1$) as a special case.
%The complete proof of Theorem \ref{thm:main-lower} is given in the appendix.
}

\subsection{Extension to higher dimensions}\label{subsec:lower-highd}
The lower bound construction can be extended to higher dimensions $d>1$ to
obtain a matching lower bound of $V_T^{2p/(4p+d)}\cdot T$ for noisy gradient feedback and $V_T^{2p/(6p+d)}\cdot T$ for noisy function value feedback.
Let $\vct 1=(1,\cdots,1)\in\mathbb R^d$ be a $d$-dimensional vector with all components equal to $1$.
We consider $\mathcal X=\{x\in\mathbb R^d: x\geq 0,\vct 1^\top x\leq 1\}$.
%Clearly $\mathcal X$ is a convex, bounded subset of $\mathbb R^d$ because it is the unit $d$-dimensional ball in $\ell_1$ norm.
Define $\bar F_\lambda:\mathcal X\to\mathbb R$ as follows:
\begin{equation}
\bar F_\lambda(x) := F_\lambda(\vct 1^\top x) + \|x\|_2^2, \;\;\;\;\;\;\lambda\in[0,1], x\in\mathcal X.
\label{eq:barf}
\end{equation}
Here $F_\lambda$ is the univariate function defined in Eq.~(\ref{eq:flambda}).
%Note that $\vct 1^\top x$ is within the domain of $F_\lambda$ because $\|x\|_1\leq 1$ implies $-1\leq \vct 1^\top x\leq 1$.
Intuitively, the multi-variate function $\bar F$ is constructed by ``projecting'' a $d$-dimensional vector $x$ onto a 1-dimensional axis supported on $[0,1]$,
and subsequently invoking existing univariate construction of adversarial functions.
An additional quadratic term $\|x\|_2^2$ is appended to ensure the strong convexity of $\bar F_\lambda$ without interfering with the structure in $F_\lambda$.
The following lemma lists the properties of $\bar F$, which are rigorously verified in Section \ref{sec:lem_properties_highd} in the online supplement.
\begin{lemma}
Suppose $1/8T^2\leq h\leq 1/8$.
The following statements are true for any fixed $d\in\mathbb N$ and all $\lambda,\mu\in[1/4,3/4]$.
\begin{enumerate}
\item $\bar F_\lambda$ satisfies (A1) through (A5) with $D=2$, $\nu=1/16\sqrt{d+1}$, $H=16\sqrt{d}+2$, $L=26\sqrt{d}+2$ and $M=2$.
\item $\|\bar F_\lambda-\bar F_\mu\|_{\infty}\leq |\lambda-\mu|\cdot O(h)$ and $\sup_{x\in\mathcal X}\|\nabla\bar F_\lambda(x)-\nabla\bar F_\mu(x)\|_2\leq |\lambda-\mu|\cdot O(\sqrt{h})$.
\item $\|\bar F_\lambda-\bar F_\mu\|_{p}\leq |\lambda-\mu|\cdot O(h^{(2p+d)/2p})$ for all $1\leq p<\infty$.
\item $\chi(\bar F_\lambda,\bar F_{1-\lambda}) = \frac{d}{d+1}\left(\frac{1}{2}-\lambda\right)^2\cdot h$.
\end{enumerate}
\label{lem:properties-highd}
\end{lemma}

The third property in Lemma \ref{lem:properties-highd} deserves special attention, which is a key property that is significantly different from Lemma \ref{lem:properties}
for the univariate case,
because the dependency of $\|\bar F_\lambda-\bar F_\mu\|_p$ on $h$ has an extra term involving the domain dimension $d$ in the exponent.
At a higher level, the presence of the $O(h^{2p/(2p+d)})$ term comes from the \emph{concentration of measure} phenomenon in high dimensions.
%\begin{equation}
%\frac{\vol(\{x\in\mathbb R^d: x\geq 0, \vct 1^\top x\leq r\})}{\vol(\mathcal X)}
%= \frac{\vol(\mathbb B_d(r))}{\vol(\mathbb B_d(1))} = r^d, \;\;\;\;\;\forall r\in(0,1).
%\label{eq:concentration}
%\end{equation}
%Eq.~(\ref{eq:concentration}) shows that, as dimension $d$ increases, the volume ratio of the region on which $\bar F_\lambda$ remains unchanged for all $\lambda\in[1/4,3/4]$
%and the volume of the entire domain $\mathcal X$ decays exponentially with $d$.
%The $O(h^{2p/(2p+d)})$ term thus arises from this observation.
%\xnote{I do not completely follow this reasoning and why ``decays exponentially''?}

We then have the next corollary, by following the same construction of $\Theta\subseteq\mathcal F_{p,q}(V_T)$ in the univariate case and invoking Lemma \ref{lem:properties-highd}:
\begin{corollary}
Suppose $1\leq J\leq T$ is even, $\Delta_T=\lfloor T/J\rfloor$ and $1/8T^2\leq h\leq 1/8$.
Let $\mathcal I\subseteq\{0,1\}^J$ be constructed according to Lemma \ref{lem:coding},
and $\Theta=\{\vct f_{\vct i}:\vct i\in\mathcal I\}$, where $\vct f_{\vct i}$ is defined in Eq.~(\ref{eq:construction-1d})
except that $F_\lambda$ is replaced with its high-dimensional version $\bar F_\lambda$ defined in Eq.~(\ref{eq:barf}). Then
the following holds:
\begin{enumerate}
\item \emph{(Variation)}. $\sup_{\vct f\in\Theta}\var_{p,q}(\vct f)\leq O(h^{(2p+d)/2p}/\Delta_T)$ for all $1\leq p<\infty$, $1\leq q\leq \infty$.
\item \emph{(Discrepancy)}. $\inf_{\vct f,\tilde{\vct f}\in\Theta}\sum_{t=1}^T{\chi(f_t,\tilde f_t)} \geq \Omega(h T)$.
\item \emph{(KL-divergence)}. For all admissible policy $\pi$, $\sup_{\vct f,\tilde{\vct f}\in\Theta}\kl(P_{\vct f}^{\phi,\pi}\| P_{\tilde{\vct f}}^{\phi,\pi})\leq O(hT)$ for
$\phi=\phi_t^{\mathcal G}(x_t,f_t)$ and $\sup_{\vct f,\tilde{\vct f}\in\Theta}\kl(P_{\vct f}^{\phi,\pi}\| P_{\tilde{\vct f}}^{\phi,\pi})\leq O(h^2T)$
for $\phi=\phi_t^{\mathcal F}(x_t,f_t)$.
\end{enumerate}
\label{cor:lb-md}
\end{corollary}

{
%Finally, set $h\asymp V_T^{2p/(4p+d)}$, $\Delta_T\asymp  V_T^{-2p/(4p+d)}$ for noisy gradient feedbacks $\phi_t^{\mathcal G}(x_t,f_t)$,
%and $h\asymp V_T^{2p/(6p+d)}$ and $\Delta_T\asymp T\cdot V_T^{-4p/(6p+d)}$ for noisy function value feedbacks $\phi_t^{\mathcal F}(x_t,f_t)$.
%The complete proof is again deferred to the appendix.
}

We now prove the multi-dimensional case based on Corollary \ref{cor:lb-md}. %as the univariate case ($d=1$) is an easier special case.
First consider the noisy gradient feedback $\phi_t^{\mathcal G}(x_t,f_t)$.
Set $h\asymp V_T^{2p/(4p+d)}$ and $\Delta_T$ accordingly such that $\var_{p,\infty}(\vct f) \leq O(h^{(2p+d)/2p}/\Delta_T) = V_T$.
This yields $\Delta_T\asymp V_T^{-2p/(4p+d)}$ and $J=T/\Delta_T\asymp T V_T^{2p/(4p+d)}$.
The KL divergence is then upper bounded by $O(h T)=O(T V_T^{2p/(4p+d)})$ and $\log|\Theta| = \Omega(J) = \Omega(T V_T^{2p/(4p+d)})$.
By carefully selecting constants in the asymptotic notations, one can make the right-hand side of Eq.~(\ref{eq:fano}) to be lower bounded by 1/2.
Subsequently invoking Lemma \ref{lem:reduction}, we conclude that there does not exist an admissible policy $\pi$ such that
$\sup_{\vct f\in\mathcal F_{p,\infty}(V_T)}R_{\phi}^{\vct\pi}(\vct f) \leq 1/9\cdot \inf_{\vct f,\tilde{\vct f}\in\Theta}\sum_{t=1}^T{\chi(f_t,\tilde f_t)}$.
The lower bound proof is then completed by the discrepancy claim in Corollary \ref{cor:lb-md} that $\inf_{\vct f,\tilde{\vct f}\in\Theta}\sum_{t=1}^T{\chi(f_t,\tilde f_t)} \geq \Omega(hT) = \Omega(T V_T^{2p/(4p+d)})$.

The proof for the noisy function value feedback $\phi_t^{\mathcal F}(x_t,f_t)$ is similar.
The only difference is that the KL divergence is upper bounded by $O(h^2 T)$, for which we should set $h\asymp V_T^{2p/(6p+d)}$ that leads to $\Delta_T\asymp V_T^{-4p/(6p+d)}$
and $J\asymp T V_T^{4p/(6p+d)}$.
We then have that $\mathrm{KL}\leq O(TV_T^{4p/(6p+d)})$ and $\log|\Theta| \geq \Omega(J) =\Omega(TV_T^{4p/6p+d})$.
Using the same argument, the regret for any admissible policy is lower bounded by $\inf_{\vct f,\tilde{\vct f}\in\Theta}\sum_{t=1}^T{\chi(f_t,\tilde f_t)} \geq \Omega(hT) = \Omega(T V_T^{2p/(6p+d)})$.

\subsection{Proof of Lemma \ref{lem:reduction}}
Let $\pi$ be a policy that attains the minimax rate. By Markov's inequality, with probability $2/3$ it holds that
\begin{equation}
\frac{1}{T}\sum_{t=1}^T{f_t(x_t)-f_t(x_t^*)} \leq \frac{1}{3}\cdot \inf_{\vct f,\tilde{\vct f}\in\Theta}\sum_{t=1}^T{\chi(f_t,\tilde f_t)}, \;\;\;\;\;\forall \vct f,\tilde{\vct f}\in\Theta.
\label{eq:closeness}
\end{equation}
Define $\hat{\vct f} := \arg\min_{\hat{\vct f}\in\Theta}\sum_{t=1}^T{\hat f_t(x_t)-\hat f_t(\hat x_t^*)}$, where $\hat x_t^*$ is the (unique) minimizer of $\hat f_t$.
Let $\hat f_t^*=\inf_{x\in\mathcal X}\hat f_t(x)$ and $f_t^*=\inf_{x\in\mathcal X}f_t(x)$. %be the minimal solution of $\hat f_t$ and $f_t$ on $\mathcal X$.
Because $\hat{\vct f}$ minimizes the ``empirical'' regret on $\{x_t\}_{t=1}^T$, it holds that
$$
\sum_{t=1}^T{\hat f_t(x_t)-\hat f_t^*} \leq \sum_{t=1}^T{f_t(x_t) - f_t^*}.
$$
Subsequently,
\begin{align*}
\sum_{t=1}^T{\chi(\hat f_t,f_t)}&=\sum_{t=1}^T{\inf_{x\in\mathcal X}\max\left\{\hat f_t(x)-\hat f_t^*, f_t(x)-f_t^*\right\}} \\
&\leq \sum_{t=1}^T{\max\left\{\hat f_t(x_t)-\hat f_t^*, f_t(x_t)-f_t^*\right\}}\\
&\leq \sum_{t=1}^T{\hat f_t(x_t)-\hat f_t^* + f_t(x_t)-f_t^*}\\
&\leq 2\left(\sum_{t=1}^T{f_t(x_t)-f_t^*}\right) \leq \frac{2}{3}\cdot \inf_{\vct f,\tilde{\vct f}\in\Theta}\sum_{t=1}^T{\chi(f_t,\tilde f_t)}.
\end{align*}
Therefore, we must have $\hat{\vct f}=\vct f$ conditioned on Eq.~(\ref{eq:closeness}), which completes the proof.
%\endproof

\subsection{Proof of Lemma \ref{lem:properties}}
\label{sec:supp_lem_properties}
We verify the properties separately.

\noindent\textbf{Verification of property 1}:
(A1) is obvious because $\mathcal X=[0,1]$.
We next focus (A3), (A4) and (A5).
It is easy to check that if two functions $f$ and $g$ satisfy (A3) through (A5), then their convex combination
$f + \lambda(g-f)$ for $\lambda\in[0,1]$ also satisfies (A3) through (A5).
Therefore we only need to verify these conditions for $F_0$ and $F_1$, respectively.
We first prove that both $F_0$ and $F_1$ are differentiable.
Because both $F_0$ and $F_1$ are differentiable within each piece, to prove the global differentiablity we only need to show that
the left and right function values and derivatives of $F_0$ and $F_1$ at $x=\sqrt{h}$ and $x=2\sqrt{h}$ are equal.
Define $F(x^+) = \lim_{t\to 0^+}F(x+t)$, $F(x^-)=\lim_{t\to 0^-}F(x+t)$, $F'(x^+) = \lim_{t\to 0^+}\frac{F(x+t)-F(x)}{t}$ and $F'(x^-)=\lim_{t\to 0^-}\frac{F(x+t)-F(x)}{t}$.
We then have that
$F_1(\sqrt{h}^+)=F_0(\sqrt{h}^-)=h$, $F_0(2\sqrt{h}^+)=F_0(2\sqrt{h}^-)=8h$, $F_0'(\sqrt{h}^+)=F_0'(\sqrt{h}^-)=2\sqrt{h}$, $F_0'(2\sqrt{h}^+)=F_0'(2\sqrt{h}^-)=16\sqrt{h}$,
$F_2(\sqrt{h}^+)=F_1(\sqrt{h}^-)=0$, $F_1'(\sqrt{h}^+)=F_1'(\sqrt{h}^-)=0$.
Therefore, both $F_0$ and $F_1$ are differentiable on $[0,1]$.
It is then easy to check that $\sup_{0\leq x\leq 1}\max\{|F_0(x)|, |F_1(x)|\}\leq 8$ and $\sup_{0\leq x\leq 1}\max\{|F_0'(x)|,|F_1'(x)|\}\leq 16$.
Therefore (A3) is satisfied with $H=16$.

To verify (A4) and (A5) we need to compute the second-order derivatives of $F_0$ and $F_1$.
By construction, $F_0''(x)=F_1''(x)=2$ for $x\in[0,\sqrt{h}]$, $F_0''(x)=F_1''(x)=8$ for $x\in[\sqrt{h},1]$,
and $2\leq F_0''(x)\leq 26$ for $x\in[\sqrt{h},2\sqrt{h}]$.
Therefore, $F_0$ and $F_1$ satisfy (A4) and (A5) with $L=26$ and $M=2$.
Note that $F_0$ and $F_1$ are \emph{not} twice differentiable at $x=\sqrt{h}$ and $x=2\sqrt{h}$:
however, this does not affect the smoothness and strong convexity of both functions.

Finally we check (A2).
Let $x_{\lambda}^*$ be the unique minimizer of $F_\lambda = F_0 + \lambda(F_1-F_0)$.
Elementary algebra yields that $x_{\lambda}^* = \lambda \sqrt{h}$.
Because $h\geq 1/8T^2$, we know that $F_{\lambda}$ satisfies (A2) with $\nu=1/32$ for $\lambda\in[1/4, 3/4]$.

\noindent\textbf{Verification of property 2}: It is easy to see that $\|F_\lambda-F_\mu\|_p = |\lambda-\mu|\cdot \|F_0-F_1\|_p$ and $\|F_\lambda'-F_\mu'\|_p\leq |\lambda-\mu|\cdot \|F_0'-F_1'\|_p$ for all $1\leq p\leq\infty$.
Thus we only need to consider $\lambda=0$ and $\mu=1$.
It is easy to verify that $\|F_0-F_1\|_{\infty} = |F_0(0)-F_1(0)| = \sqrt{h}$ and $\|F_0'-F_1'\|_{\infty} = |F_0(0)'-F_1(1)'| = 2h$.

\noindent\textbf{Verification of property 3}: Similarly we only need to consider $\lambda=0$ and $\mu=1$.
Because $F_0$ and $F_1$ only differ on $[0,2\sqrt{h}]$, we have that
$$
\|F_0-F_1\|_p = \left(\int_0^{2\sqrt{h}}|F_0(x)-F_1(x)|^p\ud x\right)^{1/p} = O(h^{1/2p})\cdot \|F_0-F_1\|_{\infty} = O(h^{(2p+1)/2p}).
$$

\noindent\textbf{Verification of property 4}: We have that $x_{\lambda}^*=\lambda\sqrt{h}$ and $F_\lambda^* = F_\lambda(x_\lambda^*) = \lambda(1-\lambda)h$.
Subsequently, $\chi(F_\lambda,F_{1-\lambda}) = F_{\lambda}(\sqrt{h}/2) - F_\lambda^* = (1/2-\lambda)^2\cdot h/4$.

\subsection{Proof of Lemma \ref{lem:key-1d}}
\label{sec:proof_lem_key_1d}

Fix an arbitrary interval $I_j$ for some $j\in\{1,\cdots,J\}$.
Without loss of generality assume $|I_j|=\Delta_T$ (the extra one function in some intervals can be safely neglected as both $T$ and $\Delta_T$ are large).
Then
$$
\sup_{t\in I_j}\|f_{t+1}-f_t\|_p = \frac{1}{\Delta_T}\cdot O(h^{(2p+1)/2p}).
$$
Subsequently, $$\var_{p,\infty}(\vct f) = \sup_{1\leq t\leq T-1}\|f_{t+1}-f_t\|_p = O(h^{(2p+1)/2p}/\Delta_T).$$

For the discrepancy term, again fix $I_j$ for some $j\in\{1,\cdots,J\}$ such that $\vct i_j\neq\vct i_{j}'$.
We then have,
$$
\sum_{t\in I_j}{\chi(\vct f_{\vct i,t},\vct f_{\vct i',t})}
\geq 2\sum_{t=0}^{\lfloor \Delta_T/2\rfloor}{\chi(F_{0.5+t/\Delta_T}, F_{0.5-t/\Delta_T})}
\geq 2\sum_{t=0}^{\lfloor\Delta_T/2\rfloor}{\left(\frac{t}{\Delta_T}\right)^2\cdot \Omega(h)}
= \Omega(h\Delta_T).
$$
Subsequently, summing over all intervals with $\vct i_j\neq\vct i_j'$ we have that $\sum_{t=1}^T{\chi(\vct f_{\vct i,t},\vct f_{\vct i',t})} \geq \Delta_H(\vct i,\vct i')\cdot \Omega(h\Delta_T)$.

Finally we compute the KL divergence $\kl(P_{\vct f_{\vct i}}^{\phi,\pi}\|P_{\vct f_{\vct i'}}^{\phi,\pi})$.
We first consider the noisy function value feedback $\phi=\phi_t^{\mathcal F}(x_t,f_t)$.
Let $y_t=\phi_t^{\mathcal F}(x_t,f_t)$ be the random variables of the feedbacks
and denote $x^t=(x_1,\cdots,x_t)$ and $y^t=(y_1,\cdots,y_t)$.
For any admissible policy $\pi$, we have that
\begin{align*}
\kl(P_{\vct f_{\vct i}}^{\phi,\pi}\|P_{\vct f_{\vct i'}}^{\phi,\pi})
&= \mathbb E_{\vct f_{\vct i},\pi}\left[\log\frac{P_{\vct f_{\vct i}}^{\phi,\pi}(x^T,y^T)}{P_{\vct f_{\vct i'}}^{\phi,\pi}(x^T,y^T)}\right]\\
&= \mathbb E_{\vct f_{\vct i},\pi}\left[\log\frac{P_{\vct f_{\vct i}}^\phi(y^T|x^T)\cdot \prod_{t=1}^T{P_{\pi}(x_t|y^{t-1},x^{t-1})}}{P_{\vct f_{\vct i'}}^\phi(y^T|x^T)\cdot \prod_{t=1}^T{P_{\pi}(x_t|y^{t-1},x^{t-1})}}\right]\\
&= \sum_{t=1}^T{\mathbb E_{\vct f_{\vct i}, \pi}\left[\log\frac{P_{\vct f_{\vct i},t}^\phi(y_t|x_t)}{P_{\vct f_{\vct i'},t}^\phi(y_t|x_t)}\right]}\\
%&= \int P_{\vct f_{\vct i}}^\phi(y^T|x^T) \prod_{t=1}^T{P_{\pi}(x_t|y^{t-1},x^{t-1})}\cdot \log\frac{P_{\vct f_{\vct i}}^\phi(y^T|x^T)\cdot \prod_{t=1}^T{P_{\pi}(x_t|y^{t-1},x^{t-1})}}{P_{\vct f_{\vct i'}}^\phi(y^T|x^T)\cdot \prod_{t=1}^T{P_{\pi}(x_t|y^{t-1},x^{t-1})}}\ud x^T\ud y^T\\
%&= \int P_{\vct f_{\vct i}}^\phi(y^T|x^T) \prod_{t=1}^T{P_{\pi}(x_t|y^{t-1},x^{t-1})}\cdot \log\frac{P_{\vct f_{\vct i}}^\phi(y^T|x^T)}{P_{\vct f_{\vct i'}}^\phi(y^T|x^T)}\ud x^T\ud y^T\\
%&\leq \sup_{x^T}\int P_{\vct f_{\vct i}}^\phi(y^T|x^T)\log\frac{P_{\vct f_{\vct i}}^\phi(y^T|x^T)}{P_{\vct f_{\vct i'}}^\phi(y^T|x^T)}\ud y^T
%= \sup_{x^T}\kl(P_{\vct f_{\vct i}}^\phi(\cdot|x^T)\|P_{\vct f_{\vct i'}}^\phi(\cdot|x^T))\\
&\leq \sum_{t=1}^T{\sup_{x\in\mathcal X}\kl(P_{\vct f_{\vct i,t}}^\phi(\cdot|x)\|P_{\vct f_{\vct i',t}}^\phi(\cdot|x))}.
\end{align*}
Here the third identity holds because $\varepsilon_t$ are independent.
%Here the last equality holds because $\varepsilon_t$ are independent.
For $y_t=\phi_t(x_t,f_t)\sim\mathcal N(f_t(x_t), 1)$, it holds that
\begin{equation}
\sup_{x\in\mathcal X}\kl(P_{\vct f_{\vct i,t}}^\phi(\cdot|x)\|P_{\vct f_{\vct i',t}}^\phi(\cdot|x))
= \sup_{x\in\mathcal X}\big|\vct f_{\vct i,t}(x)-\vct f_{\vct i',t}(x)\big|^2 = \|\vct f_{\vct i,t}-\vct f_{\vct i',t}\|_{\infty}^2 = O(h^2).
\label{eq:kl-key}
\end{equation}
where in the last inequality we invoke Lemma \ref{lem:properties}.
Summing over $t=1$ to $T$ we have that $\kl(P_{\vct f_{\vct i}}^{\phi,\pi}\|P_{\vct f_{\vct i'}}^{\phi,\pi})= O(h^2T)$.

The noisy gradient feedback case $\phi=\phi_t^{\mathcal G}(x_t,f_t)$ can be handled by following the same argument, except that Eq.~(\ref{eq:kl-key}) should be replaced by
$$
\sup_{x\in\mathcal X}\kl(P_{\vct f_{\vct i,t}}^\phi(\cdot|x)\|P_{\vct f_{\vct i',t}}^\phi(\cdot|x))
= \sup_{x\in\mathcal X}\big|\vct f_{\vct i,t}'(x)-\vct f_{\vct i',t}'(x)\big|^2 = \|\vct f_{\vct i,t}'-\vct f_{\vct i',t}'\|_{\infty}^2 = O(h).
$$
Therefore, $\kl(P_{\vct f_{\vct i}}^{\phi,\pi}\|P_{\vct f_{\vct i'}}^{\phi,\pi})= O(hT)$.

\subsection{Proof of Lemma \ref{lem:properties-highd}}
\label{sec:lem_properties_highd}
We verify the properties separately.

\noindent\textbf{Verification of property 1}:
Because $\forall x\in\mathcal X$, $\|x\|_1\leq 1$, we have that $\|x-y\|_2\leq\|x-y\|_1\leq 2$ for all $x,y\in\mathcal X$ and therefore
$\mathcal X$ satisfies (A1) with $D=2$.

We next verify (A3). Because $F_\lambda$ is convex differentiable, it holds that $\bar F_{\lambda}(x)=F_{\lambda}(\vct 1^\top x)+\|x\|_2^2$ is also convex differentiable
because convexity is preserved with affine transform.
In particular, $\sup_{x\in\mathcal X}\bar F_{\lambda}(x) \leq \|F_{\lambda}\|_{\infty} + 1 \leq 9$ and
$\sup_{x\in\mathcal X}\|\bar F_{\lambda}(x)\|_2 \leq \sup_{x\in\mathcal X}|F_{\lambda}(\vct 1^\top x)|\cdot \|\vct 1\|_2 + 2\|x\|_2 \leq 16\sqrt{d} + 2$.
Therefore, (A3) is satisfied with $H=16\sqrt{d} + 2$.

To verify (A4) and (A5), note that $\bar F_\lambda$ is twice differentiable except at points $\{x: \vct 1^\top x=\sqrt{h}\}\cup\{x: \vct 1^\top x=2\sqrt{h}\}$.
Furthermore, $\nabla^2\bar F_{\lambda}(x) = F_{\lambda}''(\vct 1^\top x)\cdot \vct 1\vct 1^\top + 2I_d$.
Subsequently, on points $x\in\mathcal X$ where $\bar F_{\lambda}$ is twice differentiable, we have that $\nabla^2\bar F_\lambda(x) \preceq (\|F_\lambda''\|_{\infty}\sqrt{d}+2)I = (26\sqrt{d}+2)I$
and $\nabla^2\bar F_\lambda(x)\succeq 2I$.
Therefore, (A4) and (A5) are satisfied with $L=26\sqrt{d}+2$ and $M=2$.

Finally we check (A2).
Let $x_\lambda^*$ be the unique minimizer of $\bar F_\lambda$ on $\mathcal X$.
It is clear that $x_\lambda^*$ must take the form of $x_\lambda^* = (\bar x_\lambda^*,\cdots,\bar x_\lambda^*)$,
which gives the smallest $\|x\|_2^2$ without changing the value of $F_\lambda(\vct 1^\top x)$.
Completing the squares in $\bar F_\lambda$ we have that
\begin{equation}
\bar F_\lambda(\bar x_\lambda) = d(d+1)\left[\bar x-\frac{\lambda\sqrt{h}}{d+1}\right]^2 + \lambda h\left[1-\frac{\lambda d}{d+1}\right].\label{eq:complete-square}
\end{equation}
Subsequently, $\bar x_\lambda^* = \frac{\lambda\sqrt{h}}{d+1}$.
It is easy to verify that for $h\leq 1/8$, $\inf\{t\geq 0: \bar x_\lambda^*+tu\in\mathcal X \forall u\in\mathbb B_d(1)\}\geq \|\bar x_\lambda^*\|_2\geq \lambda\sqrt{h/(d+1)}$.
Therefore, for all $\lambda\in[1/4,3/4]$ and $1/8T^2\leq h1\leq 1/8$ the condition (A2) holds with $\nu=1/16\sqrt{d+1}$.

\noindent\textbf{Verification of property 2}:
$\|\bar F_\lambda-\bar F_{\mu}\|_{\infty} = \|F_{\lambda}-F_{\mu}\|_{\infty} = O(h)$.
In addition, $\sup_{x\in\mathcal X}\|\nabla\bar F_\lambda(x)-\nabla\bar F_{\mu}(x)\|_2 = \|F_{\lambda}'-F_\mu'\|_{\infty}\cdot \|\vct 1\|_2 = O(\sqrt{hd})$.
Omitting the dependency on $d$ we obtain property 2.

\noindent\textbf{Verification of property 3}:
Define $\tilde{\mathbb B}_d(r) := \{x\in\mathbb R^d: x\geq 0, \|x\|_1\leq r\}$.
It is easy to verify that $\vol(\tilde{\mathbb B}_d(r_1))/\vol(\tilde{\mathbb B}_d(r_2)) = (r_1/r_2)^d$.
Subsequently, for any $1\leq p<\infty$ we have that
$$
\|\bar F_0-\bar F_1\|_p \leq \left[\frac{\vol(\tilde{\mathbb B}_d(2\sqrt{h}))}{\vol(\tilde{\mathbb B}_d(1))}\cdot \|\bar F_0-\bar F_1\|_{\infty}^p\right]^{1/p} = O(h^{(2p+d)/2p}).
$$

\noindent\textbf{Verification of property 4}:
From previous derivations we know that $x_\lambda^*=(\bar x_{\lambda}^*,\cdots,\bar x_{\lambda}^*)$ with $\bar x_{\lambda}^*=\frac{\lambda\sqrt{h}}{d+1}$ and
$\bar F_\lambda^* = \inf_{x\in\mathcal X}\bar F_\lambda(x)=\lambda h(1-\frac{\lambda d}{d+1})$.
Subsequently,
$$
\chi(\bar F_\lambda, \bar F_{1-\lambda}) = \bar F_\lambda\left(\frac{1}{2}\frac{\sqrt{h}}{d+1}\right) = \frac{d}{d+1}\left|\frac{1}{2}-\lambda\right|^2\cdot h.
$$

{

\section{Proof of Theorem \ref{thm:general-convex}}\label{subsec:proof-sketch-convex}

We give the proof of Theorem \ref{thm:general-convex} that establishes an upper regret bound for functions that are merely convex;
i.e., do not satisfy smoothness (A4) or strong convexity (A5).
The meta-policy remains the same, and the sub-policy is also the OGD algorithm, but with a slightly different step size rule.
The following lemma then upper bounds the regret against \emph{stationary} benchmarks.
It is a standard result in online convex optimization, whose proof can be found in, e.g., \citep{bubeck2012regret}.
\begin{lemma}
Fix $1\leq T'\leq T$.
Let $\vct f=(f_1,\cdots,f_{T'})$ be an \emph{arbitrary} sequence of convex functions satisfying (A1) through (A3).
For noisy gradient feedback and the OGD policy, the following holds with $\eta_t=1/\sqrt{t}$:
$$
S_\phi^\pi(\vct f;x^*) = O(\sqrt{T'}), \;\;\;\;\;\;\text{for}\;\;\phi=\phi_t^{\mathcal G}(x_t,f_t),\;\; \pi=\pi_s^\mathcal G\;\;\text{and all}\;\; x^*\in\mathcal X.
$$
\label{lem:general-convex}
\end{lemma}

We next derive a decomposition of the strong regret and an affinity lemma, similar to Eq.~(\ref{eq:decomp}) and Lemma \ref{lem:key2}.
However, because strong convexity of $f_t$ is no longer assumed, it is technically challenging to upper bound $|f_t(x_t^*)-f_t(x_\tau^*)|$
as in Lemma \ref{lem:key2}.
To overcome such difficulties, we gave an alternative decomposition of the strong regret, which can then be upper bounded for general convex functions.

For any sequence of convex functions $\{f_t\}_{t=1}^{T'}$ let $x_t^*$ be the minimizer of $f_t$.
Then for any $\tau\in[T']$, the gap between strong and weak regret can be decomposed as
\begin{align}
R_\phi^\pi(\vct f) - S_\phi^\pi(\vct f,x_\tau^*)
&= \sum_{t=1}^{T'} f_t(x_\tau^*)-f_t(x_t^*)\leq \sum_{t=1}^{T'} f_\tau(x_t^*)-f_\tau(x_\tau^*) + f_t(x_\tau^*)-f_t(x_t^*)\nonumber\\
&\leq \sum_{t=1}^{T'} |f_\tau(x_t^*)-f_t(x_t^*)| + |f_\tau(x_\tau^*)-f_t(x_\tau^*)|
\leq 2T'\cdot\max_{1\leq t,\tau\leq T'} |f_t(x_t^*)-f_\tau(x_t^*)|.\label{eq:convex-decomposition}
\end{align}

Here the second inequality holds because $x_\tau^*$ is the minimizer of $f_\tau$, and therefore $f_\tau(x_t^*)\geq f_\tau(x_\tau^*)$ for all $x_t^*\in\mathcal X$.
We then have the following affinity lemma that upper bounds $|f_\tau(x_t^*)-f_t(x_t^*)|$ using $\|f_\tau-f_t\|_p$.
\begin{lemma}
Suppose $\mathcal X\subseteq\mathbb R^d$.
Fix $1\leq p<\infty$, $t\neq \tau$ and let $x_t^*,x_\tau^*$ be the minimizers of $f_t$ and $f_\tau$, respectively. Then under (A1) through (A3) we have that
$$
\max\left\{\big|f_t(x_t^*)-f_\tau(x_t^*)\big|, \big|f_t(x_\tau^*)-f_\tau(x_\tau^*)\big|\right\} = O\left(\|f_t-f_\tau\|_p^{r'}\right) \;\;\;\;\;\text{where}\;\;\;\;r'=\frac{p}{p+d}\in(0,1).
$$
\label{lem:key1}
\end{lemma}

While Lemmas \ref{lem:key1} and \ref{lem:key2} appear similar, their proofs are actually different.
Unlike the proof of Lemma \ref{lem:key2} that analyzes $\|x_t^*-x_\tau^*\|_2$ and uses the strong convexity of $f_t$ and $f_\tau$ in an essential way,
in the proof of Lemma \ref{lem:key1} we directly analyze the behavior of $f_t$ and $f_\tau$ in a small neighborhood of $x_t^*$ (or $x_\tau^*)$ without resorting to $\|x_t^*-x_\tau^*\|_2$.
Thus, Lemma \ref{lem:key1} works for non-smooth and non-strongly convex functions.
We give the complete proof of Lemma \ref{lem:key1} in Sec.~\ref{subsec:proof-key1}.

Combining Lemmas \ref{lem:general-convex}, \ref{lem:key1} and Eq.~(\ref{eq:convex-decomposition})
we can prove the desired upper bound in Theorem \ref{thm:general-convex} using the same analysis in the proof of Theorem \ref{thm:main-upper}.
More specifically, we have
\begin{align*}
R_\phi^\pi(\vct f)
&\leq O(T/\sqrt{\Delta_T}) + O(\Delta_T)\cdot \max_{t,\tau\in\{\underline b_\ell,\cdots,\overline b_\ell-1\}} \|f_t-f_\tau\|_p^{r'}\\
&\leq O(T/\sqrt{\Delta_T}) + O(\Delta_T)\cdot \sum_{\ell=1}^J\left(\sum_{t=\underline b_\ell}^{\overline b_\ell-1}\|f_{t+1}-f_t\|_p\right)^{r'}\\
&\leq O(T/\sqrt{\Delta_T}) + O(\Delta_T)\cdot \Delta_T^{r'-r'/q}J^{1-r'/q}T^{r'/q}V_T^{r'} = O(T/\sqrt{\Delta_T} + \Delta_T^{r'} TV_T^{r'}).
\end{align*}
The critical parameter $\Delta_T$ should then be set as $\Delta_T=T$ if $V_T=O(T^{-(3p+d)/2p})$ and $\Delta_T\asymp V_T^{-2r'/(2r'+1)}=V_T^{-p/(3p+d)}$
otherwise.

\subsection{Proof of Lemma \ref{lem:key1}}\label{subsec:proof-key1}

Because of symmetry we only need to prove $|f_t(x_t^*)-f_\tau(x_t^*)|=O(\|f_t-f_\tau\|_p^{r'})$.
Recall the definition $\mathcal X_\alpha(x) := \{x + \rho(y-x): 0\leq \rho\leq\alpha, y\in\partial\mathcal X\}$ for $x\in\mathcal X^o$
and the properties that $\mathcal X_\alpha(x)\subseteq\mathcal X$, $\sup_{x'\in\mathcal X_\alpha(x)}\|x'-x\|_2\leq \alpha D$
and $\vol(\mathcal X_\alpha(x))\geq \alpha^d\cdot \vol(\mathcal X)$.

Now set $\alpha=\delta/D$ for some $\delta\in(0,D)$ to be specified later.
Because $f_\tau$ is Lipschitz continuous, we have $f_\tau(x)\leq f_\tau(x_t^*) + O(\delta)$ for all $x\in\mathcal X_\alpha(x_t^*)$.
On the other hand, $f_t(x)\geq f_t(x_t^*)$ for all $x\in\mathcal X$ because $x_t^*$ is the minimizer of $f_t$.
Combining both inequalities we have
$$
\big|f_t(x_t^*)-f_\tau(x_t^*)\big|^p \leq C\cdot\left[\big|f_t(x)-f_\tau(x)\big|^p + O(\delta^p)\right] \;\;\;\;\;\;\forall x\in\mathcal X_\alpha(x_t^*).
$$
Here $C>0$ is a constant that only depends on $H, D$ and $p$.
Integrating both sides of the above inequality on $\mathcal X_\alpha(x_t^*)$ and noting that $\vol(\mathcal X_\alpha(x))\geq \alpha^d\cdot \vol(\mathcal X) = \Omega(\delta^d)$, we have
$$
\delta^d\big|f_t(x_t^*)-f_\tau(x_t^*)\big|^p  \leq C \|f_t-f_\tau\|_p^p + O(\delta^{pd}).
$$
Subsequently,
$
\big|f_t(x_t^*)-f_\tau(x_t^*)\big|  \leq O(\delta^{-d/p} \|f_t-f_\tau\|_p + \delta^p).
$
Taking $\delta = \min\{\|f_t-f_\tau\|_p^{r'}, D/2\}$ where $r'=p/(p+d)$ we
have $|f_t(x_t^*)-f_\tau(x_t^*)| \leq \max\{O(\|f_t-f_\tau\|_p^{r'}), O(\|f_t-f_\tau\|_p)\}$.
Because both $f_t$ and $f_\tau$ are uniformly bounded (thanks to assumption (A2)) on a compact domain $\mathcal X$,
we have that $\|f_t-f_\tau\|_p=O(1)$ and therefore the $\|f_t-f_\tau\|_p$ term is dominated by $\|f_t-f_\tau\|_p^{r'}$,
because $r'<1$.

}

\section{Proofs of other technical propositions}

\subsection{Proof of Proposition \ref{prop:monotonicity}}
\label{sec:proof_mono}
By monotonicity of $L_p$-space, we know that for any measurable function $f:\mathcal X\to\mathbb R$,
$$
\left(\int_{\mathcal X}|f(x)|^p\ud x\right)^{1/p} \leq \vol(\mathcal X)^{1/p-1/p'}\cdot \left(\int_{\mathcal X}|f(x)|^q\ud x\right)^{1/q}, \;\;\;\;\;\;\forall 0<p\leq p'\leq\infty,
$$
provided that integration on both sides of the inequality (or there limits) exist.
Hence, $\|f_{t+1}-f_t\|_p\leq \|f_{t+1}-f_t\|_{p'}$ for all $1\leq p\leq p'\leq\infty$ and therefore $\var_{p,q}(\vct f)\leq \var_{p',q}(\vct f)$.

By H\"{o}lder's inequality, we know that for any $d$-dimensional vector $x$ it holds that
\begin{equation}
\|x\|_{q'}\leq \|x\|_q\leq d^{1/q-1/q'}\|x\|_{q'}, \;\;\;\;\;\;\forall 0<q\leq q'\leq\infty,
\label{eq:holder}
\end{equation}
where $\|x\|_q=(\sum_{i=1}^d{|x_i|^q})^{1/q}$ for $0<q<\infty$ and $\|x\|_\infty = \max_{1\leq i\leq d}|x_i|$ for $q=\infty$
is the $\ell_q$ norm of vector $x$.
Applying Eq.~(\ref{eq:holder}) on the $T$-dimensional vector $(\|f_2-f_1\|_p, \cdots, \|f_T-f_{T-1}\|_p, 0)$ we have that
$$
\left(\sum_{t=1}^{T-1}{\|f_{t+1}-f_t\|_p^q}\right)^{1/q} \leq T^{1/q-1/q'}\cdot \left(\sum_{t=1}^{T-1}{\|f_{t+1}-f_t\|_p^{q'}}\right)^{1/q'}.
$$
Multiplying both sides of the above inequality by $T^{-1/q}$ we have that $\var_{p,q}(\vct f)\leq \var_{p,q'}(\vct f)$.

\subsection{Proof of Proposition \ref{prop:re-sound}}
For any $x\in\mathcal X^o$ and $z\in\mathbb R^d$ define $\|z\|_x := \sqrt{z^\top\nabla^2\varphi(x) z}$.
The \emph{Dikin ellipsoid} $W_1(x)$ is defined as
$
W_1(x) := \left\{z\in\mathbb R^d: \|z-x\|_x\leq 1\right\}$
for all $x\in\mathcal X^o$. It is a well-known fact that $W_1(x)\subseteq\mathcal X$ for all $x\in\mathcal X^o$ \citep{abernethy2008competing,saha2011improved,hazan2014bandit}.
It remains to verify that $z=x+(\nabla^2\varphi(x)+\delta I_d)^{-1/2}u$ is in $W_1(x)$.
To see this, note that
\begin{align*}
\|z-x\|_x^2
&= u^\top(\nabla^2\varphi(x)+\delta I_d)^{-1/2}\nabla^2\varphi(x)(\nabla\varphi(x)+\delta I_d)^{-1/2}u\\
&= \|u\|_2^2 - \delta \|(\nabla^2\varphi(x)+\delta I_d)^{-1/2}u\|_2^2 \leq \|u\|_2^2 = 1.
\end{align*}
Hence, $z\in W_1(x)\subseteq\mathcal X$.
% Appendix here
% Options are (1) APPENDIX (with or without general title) or
%             (2) APPENDICES (if it has more than one unrelated sections)
% Outcomment the appropriate case if necessary
%
% \begin{APPENDIX}{<Title of the Appendix>}
% \end{APPENDIX}
%
%   or
%
% \begin{APPENDICES}
% \section{<Title of Section A>}
% \section{<Title of Section B>}
% etc
% \end{APPENDICES}

% References here (outcomment the appropriate case)

% CASE 1: BiBTeX used to constantly update the references
%   (while the paper is being written).
%\bibliographystyle{informs2014} % outcomment this and next line in Case 1
%\bibliography{<your bib file(s)>} % if more than one, comma separated

% CASE 2: BiBTeX used to generate mypaper.bbl (to be further fine tuned)
%\input{mypaper.bbl} % outcomment this line in Case 2

%If you don't use BiBTex, you can manually itemize references as shown below.

\bibliographystyle{apa-good}
\bibliography{refs}

\end{document}